\documentclass[final]{cvpr}

\usepackage{times}
\usepackage{epsfig}
\usepackage{graphicx}
\usepackage{amsmath}
\usepackage{amssymb}
\usepackage{bm}
\usepackage{booktabs}
\usepackage{breakcites}
\usepackage{capt-of}
\usepackage{colortbl}
\usepackage{multirow}
\usepackage{tabularx}
\newcolumntype{C}{>{\centering\arraybackslash}X}
\newcolumntype{L}{>{\raggedright\arraybackslash}X}
\newcolumntype{R}{>{\raggedleft\arraybackslash}X}
\makeatletter
\newcommand\footnoteref[1]{\protected@xdef\@thefnmark{\ref{#1}}\@footnotemark}
\makeatother

\newcommand\best[1]{\textbf{#1}}
\newcommand\second[1]{\textbf{\textit{#1}}}

\usepackage[pagebackref=true,breaklinks=true,colorlinks,bookmarks=false]{hyperref}

\def\cvprPaperID{8688}
\def\confYear{CVPR 2021}

\begin{document}

\title{\vspace{-0.5em}Blur, Noise, and Compression Robust Generative Adversarial Networks\vspace{-1em}}

\author{Takuhiro Kaneko$^1$
  \quad Tatsuya Harada$^{1,2}$
  \vspace{2mm}\\
  $^1$The University of Tokyo
  \quad $^2$RIKEN
}

\twocolumn[{
  \renewcommand\twocolumn[1][]{#1}
  \maketitle
  \vspace{-10mm}
  \begin{center}
    \includegraphics[width=\textwidth]{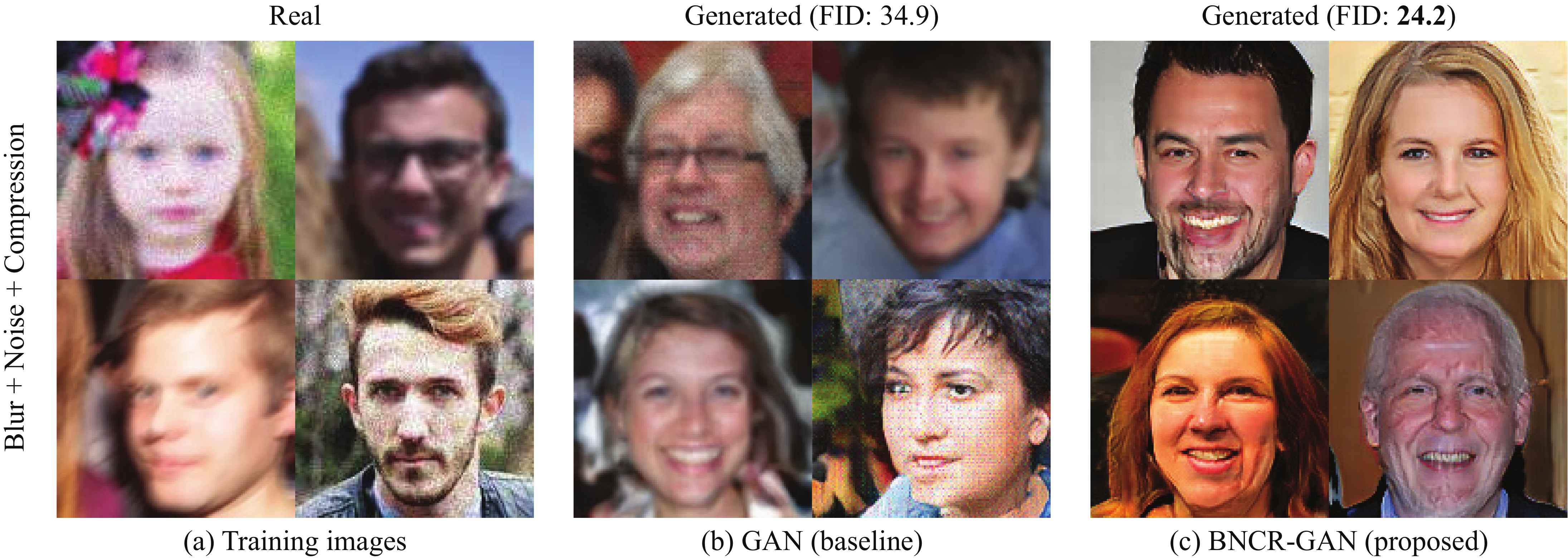}
  \end{center}
  \vspace{-4mm}
  \captionof{figure}{\textbf{Examples of blur, noise, and compression robust image generation.}
    Although recent GANs have shown remarkable results in image reproduction, they can recreate training images faithfully (b), despite degradation by blur, noise, and compression (a).
    To address this limitation, we propose \textit{blur, noise, and compression robust GAN (BNCR-GAN)}, which can learn to generate clean images (c) even when trained with degraded images (a) and without knowledge of degradation parameters (e.g., blur kernel types, noise amounts, or quality factor values).
    The project page is available at \url{https://takuhirok.github.io/BNCR-GAN/}.}
  \label{fig:examples_ffhq_all05_select}
  \vspace{3mm}
}]

\begin{abstract}
  \vspace{-3mm}
  Generative adversarial networks (GANs) have gained considerable attention owing to their ability to reproduce images. However, they can recreate training images faithfully despite image degradation in the form of blur, noise, and compression, generating similarly degraded images. To solve this problem, the recently proposed noise robust GAN (NR-GAN) provides a partial solution by demonstrating the ability to learn a clean image generator directly from noisy images using a two-generator model comprising image and noise generators. However, its application is limited to noise, which is relatively easy to decompose owing to its additive and reversible characteristics, and its application to irreversible image degradation, in the form of blur, compression, and combination of all, remains a challenge. To address these problems, we propose blur, noise, and compression robust GAN (BNCR-GAN) that can learn a clean image generator directly from degraded images without knowledge of degradation parameters (e.g., blur kernel types, noise amounts, or quality factor values). Inspired by NR-GAN, BNCR-GAN uses a multiple-generator model composed of image, blur-kernel, noise, and quality-factor generators. However, in contrast to NR-GAN, to address irreversible characteristics, we introduce masking architectures adjusting degradation strength values in a data-driven manner using bypasses before and after degradation. Furthermore, to suppress uncertainty caused by the combination of blur, noise, and compression, we introduce adaptive consistency losses imposing consistency between irreversible degradation processes according to the degradation strengths. We demonstrate the effectiveness of BNCR-GAN through large-scale comparative studies on CIFAR-10 and a generality analysis on FFHQ. In addition, we demonstrate the applicability of BNCR-GAN in image restoration.
\end{abstract}

\vspace{-8.5mm}
\section{Introduction}
\label{sec:introduction}

Constructing generative models to generate images that are indistinguishable from real images is a fundamental problem in computer vision and machine learning.
Recently, however significant advancements have been made in this regard, enabled by the emergence of deep generative models.
Among them, generative adversarial networks (GANs)~\cite{IGoodfellowNIPS2014}, which learn data distributions through adversarial training, have attracted considerable attention owing to their high image reproduction ability.

However, a persistent problem is that high-capacity GANs can replicate training images with high fidelity, even when the images are degraded, and they thus tend to replicate various forms of image degradation in their generated images.
As shown in Figure~\ref{fig:examples_ffhq_all05_select}, when standard GAN is trained with images degraded by blur, noise, and compression (i.e., JPEG) (Figure~\ref{fig:examples_ffhq_all05_select}(a)), it produces similarly degraded images (Figure~\ref{fig:examples_ffhq_all05_select}(b)) because standard GAN architectures do not consider such image degradation.
This is problematic when training images are collected in real-world scenarios (e.g., web crawling) because they potentially include degraded images.
To address this problem, painstaking manual prescreening is often conducted.

One well-explored solution involves restoring images using an image restoration model, such as model-based image restoration methods~\cite{KDabovTIP2007,SGuCVPR2014,MEladTIP2006,JMairalTIP2007,ABuadesCVPR2005,JMairalICCV2009,WDongTIP2012,FLuisierTIP2010,MMakitaloTIP2012,RFergusTOG2006,QShanTOG2008,DKrishnanCVPR2011,LXuCVPR2013,JPanCVPR2016,AFoiTIP2007}, prior to the training of GANs.
However, images restored by these methods tend to be either over- or under-restored owing to the gap between predefined and real priors.\footnote{Deep image prior-based approaches~\cite{DUlyanovCVPR2018,XPanECCV2020} can be used alternatively; however, they require optimization for each individual image.
  Pre-trained model-based approaches~\cite{XPanECCV2020} provide another solution; however, they require the collection of clean images for training the pre-trained model.}
To overcome this drawback, various learning-based methods have been developed.
However, most of these methods require additional supervision for training, such as paired supervision (e.g., pairs of clean and degraded images)~\cite{LXuNIPS2014,JSunCVPR2015,CSchulerPAMI2015,AChakrabartiECCV2016,SNahCVPR2017,DGongCVPR2017,OKupynCVPR2018,OKupynICCV2019,VJainNIPS2009,XMaoNIPS2016,YTaiICCV2017,KZhangTIP2017,KZhangTIP2018,JChenCVPR2018,SGuoCVPR2019,CDongICCV2015,PSvobodaWSCG2016,LGalteriICCV2017,JLehtinenICML2018,ZXiaNeurIPS2019}
or set-level supervision (i.e., labels indicating whether images are degraded)~\cite{TMadamECCV2018,BLuCVPR2019}.\footnote{Self-supervised learning methods~\cite{AKrullCVPR2019,JBatsonICML2019,SLaineNIPS2019} have been also proposed; however, their application has been limited to denoising.}

AmbientGAN~\cite{ABoraICLR2018} was recently proposed as a different approach.
This provides a promising solution by simulating image degradation on generated images and learning a discriminator that distinguishes a real \textit{degraded} image from a \textit{degraded} generated image.
This formulation allows the learning of a clean image generator directly from degraded images without any pre-processing or paired/set-level supervision.
However, it relies on a strong assumption that degradation parameters, such as blur kernel types, noise amounts, and quality factor values, are known in advance.

Motivated by these previous studies, we address the problem of developing a model to \textit{learn a clean image generator directly from degraded images without knowledge of degradation parameters}.
In particular, to apply the solution to real-world images, we aim to handle images degraded by a representative image degradation model~\cite{LXuNIPS2014}, which addresses blur, noise, and compression in the same order as in a real image acquisition process (detailed in Equation~\ref{eqn:degradation}).
Based on this objective, we focus on blur, noise, and compression, and refer to the abovementioned problem of \textit{blur, noise, and compression robust image generation}.
We exemplify a solution using our proposed model, as shown in Figure~\ref{fig:examples_ffhq_all05_select}(c).
We aim to devise a model that can learn to generate clean images (Figure~\ref{fig:examples_ffhq_all05_select}(c)), even when trained with blurred, noisy, and compressed images (Figure~\ref{fig:examples_ffhq_all05_select}(a)).

The recently proposed noise robust GAN (NR-GAN)~\cite{TKanekoCVPR2020}, which uses a two-generator model comprising noise and image generators, has provided a partial solution to this problem by demonstrating the ability to learn to generate clean images directly from noisy images.
However, NR-GAN assumes that image information is lossless before and after degradation and utilizes this characteristic to decompose a degraded image into clean image and degradation components.
Hence, its application is limited to noise, which has additive and reversible characteristics, and its application to irreversible degradation, in the form of blur, compression, and combination of all, remains a challenge.

To address these problems, we propose \textit{blur, noise, and compression robust GAN (BNCR-GAN)}, that can learn a clean image generator directly from blurred, noisy, and compressed images.
To solve the sub-problems, we first propose two variants: \textit{blur robust GAN (BR-GAN)} and \textit{compression robust GAN (CR-GAN)}, which are specific to blur and compression, respectively.
Along the lines of NR-GAN, BR-GAN and CR-GAN learn a blur-kernel generator and a quality-factor generator, respectively, along with clean image generators, to learn a blur-kernel/quality-factor distribution jointly with an image distribution.
However, in contrast to NR-GAN, to address the irreversible blur/compression characteristics, \textit{masking architectures} adapting degradation strengths in a data-driven are introduced, using bypasses before and after image degradation.
This architectural constraint is useful for conducting only the necessary changes through blur or compression while suppressing unnecessary changes.

The unique problem of BNCR-GAN, which is a unified model integrating BR-GAN, NR-GAN, and CR-GAN, is that it needs to handle the uncertainty caused by the combination of blur, noise, and compression.
Thus, we incorporate novel losses called \textit{adaptive consistency losses} that impose consistency between irreversible degradation processes according to the degradation strengths.
This loss helps prevent the generated image from yielding unexpected artifacts, which can disappear and become unrecognizable after irreversible processes.

As the effects of blur, noise, and compression on GANs have not been sufficiently examined in previous studies, we first conducted large-scale comparative studies on \textsc{CIFAR-10}~\cite{AKrizhevskyTech2009}, in which we compared diverse models under various degradation settings, in which we tested 134 total conditions.
Moreover, following recent large-scale studies on GANs~\cite{KKurachICML2019} and NR-GANs~\cite{TKanekoCVPR2020}, we analyze a generality on a more complex dataset, that is, \textsc{FFHQ}~\cite{TKarrasCVPR2019}.\footnote{\label{foot:lsun}We excluded \textsc{LSUN Bedroom}~\cite{FYuArXiv2015}, which was used in~\cite{KKurachICML2019,TKanekoCVPR2020}, because its images were compressed with JPEG and ground-truth non-degraded images were not available.}
Finally, we examined the applicability of BNCR-GAN in image restoration and demonstrated that, although BNCR-GAN is designed to be trained in an unsupervised manner, it is nonetheless competitive with two supervised models (i.e., CycleGAN with set-level supervision~\cite{JYZhuICCV2017} and unsupervised adversarial image reconstruction (UNIR) with a predefined image degradation model~\cite{APajotICLR2019}).

Our contributions are summarized as follows:
\begin{itemize}
  \vspace{-1mm}
  \setlength{\parskip}{1pt}
  \setlength{\itemsep}{1pt}
\item We propose \textit{blur, noise, and compression robust image generation}, wherein generation of clean images is learned directly from degraded images without knowledge of degradation parameters.
\item To address the sub-problems, we propose \textit{BR-GAN} and \textit{CR-GAN}, which train a blur-kernel generator and a quality-factor generator, respectively, along with clean image generators.
  In particular, we devise \textit{masking architectures} to adjust the degradation strengths using bypasses before and after degradation.
\item To handle all types of image degradation, we further propose \textit{BNCR-GAN}, which unifies BR-GAN, NR-GAN, and CR-GAN as a single model.
  In particular, to address the uncertainty caused by the combination, we introduce \textit{adaptive consistency losses}.
\item We provide benchmark scores for these new problems through large-scale comparative studies on \textsc{CIFAR-10} (in which we tested 134 conditions) and a generality analysis on \textsc{FFHQ}.
  We also demonstrate the applicability of BNCR-GAN in image restoration.
  The project page is available at \url{https://takuhirok.github.io/BNCR-GAN/}.
\end{itemize}

\section{Related work}
\label{sec:related_work}

\noindent\textbf{Deep generative models.}
Image generation is a fundamental problem in computer vision and machine learning.
Recently, deep generative models, such as GANs~\cite{IGoodfellowNIPS2014}, variational autoencoders~\cite{DKingmaICLR2014,DRezendeICML2014}, autoregressive models~\cite{AOordICML2016}, and flow-based models~\cite{LDinhICLRW2015,LDinhICLR2017}, have garnered attention with promising results.
All models have benefits and limitations.
A common drawback of GANs is their training instability; however, this has been mitigated by recent progress~\cite{MArjovskyICML2017,XMaoICCV2017,JHLimArXiv2017,MBellemarearXiv2017,TSalimansICLR2018,IGulrajaniNIPS2017,NKodaliArXiv2017,LMeschederICML2018,TMiyatoICLR2018b,TKarrasICLR2018,HZhangICML2019,ABrockICLR2019,TChenICLR2019,TKarrasCVPR2019,TKarrasCVPR2020,SZhaoNeurIPS2020,TKarrasNeurIPS2020}.
Here, we target GANs because their design flexibility allows the incorporation of our core ideas, that is, a multiple-generator model.
With regard to other models~\cite{AOordNIPS2017,ARazaviNeurIPS2019,JMenickICLR2019,DKingmaNeurIPS2018}, image reproducibility has improved; therefore, susceptibility to image degradation could be problematic.
Applying our ideas to them remains as a potential direction for future work.

\smallskip\noindent\textbf{Image restoration.}
Image restoration, such as deblurring, denoising, and deblocking (or compression artifact removal), is also a fundamental problem and a large body of work exists.
Typical methods are categorized as model-based methods~\cite{KDabovTIP2007,SGuCVPR2014,MEladTIP2006,JMairalTIP2007,ABuadesCVPR2005,JMairalICCV2009,WDongTIP2012,FLuisierTIP2010,MMakitaloTIP2012,RFergusTOG2006,QShanTOG2008,DKrishnanCVPR2011,LXuCVPR2013,JPanCVPR2016,AFoiTIP2007} and learning-based methods \cite{LXuNIPS2014,JSunCVPR2015,CSchulerPAMI2015,AChakrabartiECCV2016,SNahCVPR2017,DGongCVPR2017,OKupynCVPR2018,OKupynICCV2019,VJainNIPS2009,XMaoNIPS2016,YTaiICCV2017,KZhangTIP2017,KZhangTIP2018,JChenCVPR2018,SGuoCVPR2019,CDongICCV2015,PSvobodaWSCG2016,LGalteriICCV2017,JLehtinenICML2018,ZXiaNeurIPS2019,TMadamECCV2018,BLuCVPR2019,AKrullCVPR2019,JBatsonICML2019,SLaineNIPS2019}.
Recently, learning-based methods have achieved better performance; however, as discussed in Section~\ref{sec:introduction}, most require additional supervision, such as paired or set-level supervision.
By contrast, model-based methods (e.g., deblurring \cite{RFergusTOG2006,QShanTOG2008,DKrishnanCVPR2011,LXuCVPR2013,JPanCVPR2016} and deblocking~\cite{AFoiTIP2007}) can be used without such supervision.
However, the gap between predefined and real priors causes either over- or under-restoration and damages image fidelity.
We confirm this through experiments in Section~\ref{sec:experiments}.

\smallskip\noindent\textbf{Degradation robust image generation.}
Two categories of studies on degradation robust image generation have recently become evident, including those addressing label degradation~\cite{TKanekoCVPR2019,TKanekoBMVC2019,TKanekoArXiv2019,KKThekumparampilNeurIPS2018} and those addressing image degradation~\cite{ABoraICLR2018,APajotICLR2019,JYoonICML2018,SLiICLR2019,TKanekoCVPR2020}.
This study belongs to the latter category.
As mentioned in Section~\ref{sec:introduction}, AmbientGAN~\cite{ABoraICLR2018} was a pioneering model in this category; however, it is restricted by the assumption that the degradation parameters are predefined.
UNIR~\cite{APajotICLR2019} extends AmbientGAN to a conditional setting but suffers from the same restriction.
Generative adversarial imputation nets~\cite{JYoonICML2018} and MisGAN~\cite{SLiICLR2019} addressed a similar problem in the context of data imputation. However, they relied on another strong assumption that both degraded images and the corresponding masks (i.e., blur kernel or quality factor in our case) could be obtained during training.
NR-GAN~\cite{TKanekoCVPR2020} remedied these drawbacks by introducing a trainable noise generator; however, its application is limited to noise, which is relatively easy to decompose owing to its additive and reversible characteristics.
To address these drawbacks and widen the field of degradation robust image generation, we introduce BNCR-GAN, which is applicable to irreversible image degradation without knowledge of the degradation parameters.

\section{Notation and problem statement}
\label{sec:notation}

We begin by defining some notation and a problem statement.
Hereafter, we use superscripts $r$ and $g$ to denote the real and generated data, respectively.
Let $\bm{y}$ be a degraded image and $\bm{x}$ be the corresponding clean image.
Based on~\cite{LXuNIPS2014}, we consider an image degradation model simulating an image acquisition process and including typical image degradation (i.e., blur, noise, and compression):\footnote{In this study, we focus on Equation~\ref{eqn:degradation}; however, our approach can be easily extended to more general settings by incorporating differentiable image processing (e.g.,~\cite{TBrooksCVPR2019}).
  Even in this case, blur, noise, and compression remain the dominant degradation factors.
  Hence, we believe that our findings are not limited to a specific problem.}
\begin{flalign}
  \label{eqn:degradation}
  \bm{y} = \psi ( (\bm{x} * \bm{k} + \bm{n}), q ),
\end{flalign}
where $\bm{k}$ is a blur kernel (e.g., an out-of-focus or motion blur kernel), $*$ is a convolution operation, $\bm{n}$ is an additive camera noise (e.g., read and shot noise), and $\psi$ is a JPEG compression operator with a quality factor $q \in [0, 100]$.
We aim to learn a clean image generator able to produce clean images such that $p^g(\bm{x}) = p^r(\bm{x})$, from partially or completely degraded training images.\footnote{For simplicity, in Sections~\ref{sec:brgan}--\ref{sec:bncrgan}, we only explain the case where all images are degraded (i.e., all images are sampled from $p^r(\bm{y})$).
  However, as verified in Section~\ref{sec:experiments}, our models can be applied to images of which only parts are degraded (i.e., partial images are sampled from $p^r(\bm{y})$, while the remaining images are sampled from $p^r(\bm{x})$), without modifying the model.}

\begin{figure*}[tb]
  \centering
  \includegraphics[width=\textwidth]{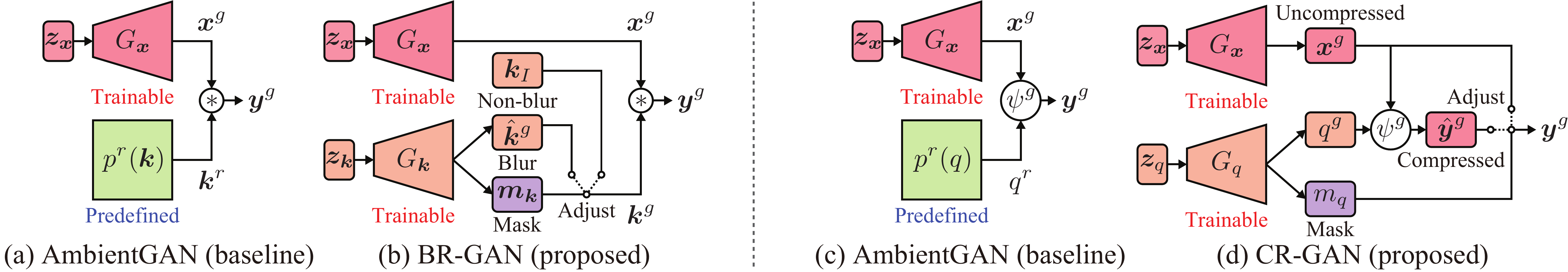}
  \caption{\textbf{Comparison of AmbientGANs (baseline), BR-GAN (proposed), and CR-GAN (proposed).}
    Because the discriminators are the same for all models, we only depict the generators.
    (a)(c) AmbientGANs assume that blur or compression simulation models are predefined.
    (b)(d) In order to eliminate this assumption, we introduce a blur-kernel generator $G_{\bm{k}}$ in BR-GAN (b) and a quality-factor generator $G_{q}$ in CR-GAN (d), and train them along with clean image generators $G_{\bm{x}}$.
    In BR-GAN (b), we introduce a \textit{masking architecture} adjusting the balance between a generated kernel $\hat{\bm{k}}^g$ (i.e., blur) and an identity kernel $\bm{k}_I$ (i.e., non-blur) based on a mask $\bm{m}_{\bm{k}}$.
    Similarly, in CR-GAN (d), we incorporate a \textit{masking architecture} adjusting the balance between a compressed image $\hat{\bm{y}}^g$ and an uncompressed image $\bm{x}^g$ based on a mask ${m}_{q}$.
    Note that every parameter (i.e., $\hat{\bm{k}}^g$, $\bm{m}_{\bm{k}}$, $q^g$, and $m_q$) is trainable, and is determined in a data-driven manner.}
  \label{fig:brgan_crgan}
  \vspace{-5mm}
\end{figure*}

As discussed in Section~\ref{sec:introduction}, AmbientGAN~\cite{ABoraICLR2018} can solve this problem by simulating image degradation on generated images before passing them to a discriminator;\footnote{Note that this is different from differentiable augmentation (DiffAugment)~\cite{SZhaoNeurIPS2020,TKarrasNeurIPS2020}.
  DiffAugment applies augmentation to \textit{both real and generated images} to learn \textit{observable} $p^r(\bm{x})$ from a few images $\bm{x}^r \sim p^r(\bm{x})$, whereas AmbientGAN applies image degradation \textit{only to generated images} to learn \textit{unobservable} $p^r(\bm{x})$ from degraded images $\bm{y}^r \sim p^r(\bm{y})$.} however, it is restricted by the need to predefine degradation simulation models (i.e., $\bm{k}^r \sim p^r(\bm{k})$, $\bm{n}^r \sim p^r(\bm{n})$, and $q^r \sim p^r(q)$ must be predefined).
NR-GAN~\cite{TKanekoCVPR2020} eliminates this restriction by making $p(\bm{n})$ learnable.
However, its application is limited to noise, where Equation~\ref{eqn:degradation} is simplified as $\bm{y} = \bm{x} + \bm{n}$ (i.e., $\bm{x}$ and $\bm{n}$ need to be decomposed additively).
Considering this, first, we develop solutions to address the remaining two types of irreversible degradation, namely, blur (in which Equation~\ref{eqn:degradation} is rewritten as $\bm{y} = \bm{x} * \bm{k}$) and compression (in which Equation~\ref{eqn:degradation} is replaced with $\bm{y} = \psi(\bm{x}, q)$).
Subsequently, we provide a solution for all types of degradation defined in Equation~\ref{eqn:degradation}.
Each solution is provided in Sections~\ref{sec:brgan},~\ref{sec:crgan}, and~\ref{sec:bncrgan}.

\section{Blur robust GAN: BR-GAN}
\label{sec:brgan}

First, we provide a solution to \textit{blur robust image generation}, which learns a clean image generator $G_{\bm{x}}$ from blurred images produced by $\bm{y}^r = \bm{x}^r * \bm{k}^r$.
As discussed in Section~\ref{sec:introduction}, AmbientGAN~\cite{ABoraICLR2018} can solve this problem via incorporation of the blur simulation model $\bm{k}^r \sim p^r(\bm{k})$ (Figure~\ref{fig:brgan_crgan}(a)).
However, it needs $p^r(\bm{k})$ to be predefined.
To alleviate this issue and for $p^r(\bm{k})$ to be learned from the data, in BR-GAN, we replace the blur simulation model with a blur-kernel generator $\bm{k}^g = G_{\bm{k}}(\bm{z}_{\bm{k}})$ (Figure~\ref{fig:brgan_crgan}(b)) and train it along with $G_{\bm{x}}$ using the following objective function.
\begin{flalign}
  \label{eqn:brgan}
  & \mathcal{L}_{\text{BR-GAN}} = \mathbb{E}_{\bm{y}^r \sim p^r(\bm{y})} [\log D_{\bm{y}}(\bm{y}^r)]
  \nonumber \\
  & + \mathbb{E}_{\bm{z}_{\bm{x}} \sim p(\bm{z}_{\bm{x}}), \bm{z}_{\bm{k}} \sim p(\bm{z}_{\bm{k}})} [\log ( 1 - D_{\bm{y}} (G_{\bm{x}}(\bm{z}_{\bm{x}}) * G_{\bm{k}}(\bm{z}_{\bm{k}})) )],
\end{flalign}
where $D_{\bm{y}}$ is a discriminator distinguishing a real \textit{blurred} image $\bm{y}^r$ from a \textit{blurred} generated image $\bm{y}^g = \bm{x}^g * \bm{k}^g$, where $\bm{x}^g = G_{\bm{x}}(\bm{z}_{\bm{x}})$ and  $\bm{k}^g = G_{\bm{k}}(\bm{z}_{\bm{k}})$.
$G_{\bm{x}}$ and $G_{\bm{n}}$ are optimized by minimizing $\mathcal{L}_{\text{BR-GAN}}$, whereas $D_{\bm{y}}$ is optimized by maximizing $\mathcal{L}_{\text{BR-GAN}}$.

In this formulation, it is challenging to adjust the strength of the blur kernel $\bm{k}^g$ because $\mathcal{L}_{\text{BR-GAN}}$ only regularizes a degraded image $\bm{y}^g$ and an over- or under-blurred $\bm{x}^g$ satisfies the solution conditions.
Hence, we introduce a \textit{masking architecture} (Figure~\ref{fig:brgan_crgan}(b)) adjusting $\bm{k}^g$'s strength using a bypass between blur and non-blur as follows:
\begin{flalign}
  \label{eqn:brgan_mask}
  \bm{k}^g = \bm{m}_{\bm{k}} \cdot \hat{\bm{k}}^g + (\bm{1} - \bm{m}_{\bm{k}}) \cdot \bm{k}_I,
\end{flalign}
where $G_{\bm{k}}$ is decomposed into a blur-kernel generator $\hat{\bm{k}}^g = G_{\bm{k}}^{\hat{\bm{k}}}(\bm{z}_{\bm{k}})$ and a mask generator $\bm{m}_{\bm{k}} = G_{\bm{k}}^{\bm{m}_{\bm{k}}}(\bm{z}_{\bm{k}})$, and $\bm{k}_I$ is the identity kernel that does not affect any changes.
$\bm{m}_{\bm{k}}$ is a matrix of the same size as $\bm{k}^g$ and values within the range of $[0, 1]$, controlling the balance between $\hat{\bm{k}}^g$ (i.e., blur) and $\bm{k}_I$ (i.e., non-blur).
Both $\hat{\bm{k}}^g$ and $\bm{m}_{\bm{k}}$ are trainable, and their distributions are optimized in a data-driven manner.

This architecture design was inspired by the recent success of masks for adjusting generative components (e.g.,~\cite{CVondrickNIPS2016} varies the foreground while preserving the background, and~\cite{APumarolaECCV2018} varies facial expressions while preserving identity).
In our case, the mask helps suppressing unnecessary changes (e.g., artifacts caused by over-/under-blurring) and allows only the necessary changes to be performed.

\begin{figure*}[tb]
  \centering
  \includegraphics[width=\textwidth]{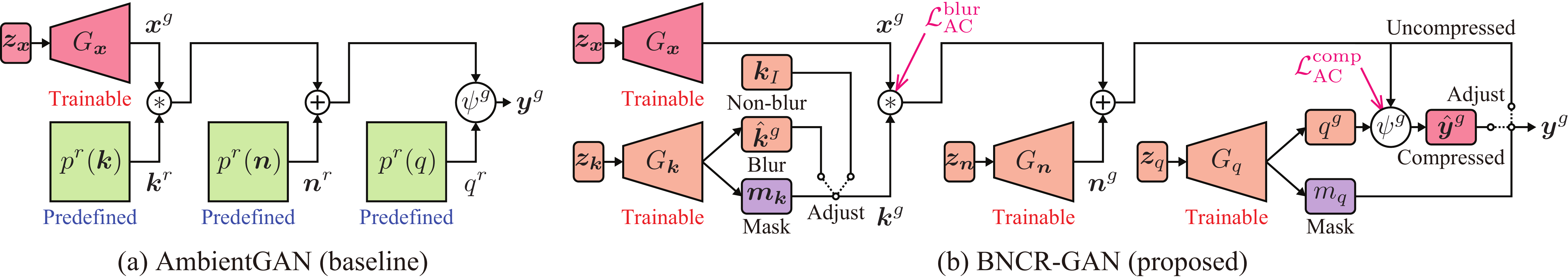}
  \caption{\textbf{Comparison of AmbientGAN (baseline) and BNCR-GAN (proposed).}
    We depict only the generators because the discriminators are the same for both models.
    (a) AmbientGAN assumes that blur, noise, and compression simulation models are predefined.
    (b) To reduce this assumption, we introduce a blur-kernel generator $G_{\bm{k}}$, a noise generator $G_{\bm{n}}$, and a quality-factor generator $G_{q}$ and train them along with a clean image generator $G_{\bm{x}}$ with \textit{adaptive consistency losses} (i.e., $\mathcal{L}_{\text{AC}}^{\text{blur}}$ and $\mathcal{L}_{\text{AC}}^{\text{comp}}$), which impose consistency between before and after blur and compression according to the blur and compression strengths (i.e., $\bm{k}^g$ and $q^g$), respectively.}
  \label{fig:bncrgan}
  \vspace{-4mm}
\end{figure*}

\section{Compression robust GAN: CR-GAN}
\label{sec:crgan}

Regarding \textit{compression robust image generation}, which learns a clean image generator $G_{\bm{x}}$ from compressed images yielded by $\bm{y}^r = \psi(\bm{x}^r, q^r)$, we use roughly the same strategy as that described in Section~\ref{sec:brgan}.
In CR-GAN, we replace the compression simulation model $q^r \sim p^r(q)$ of AmbientGAN (Figure~\ref{fig:brgan_crgan}(c)) with a quality-factor generator $q^g = G_{q}(\bm{z}_{q})$ (Figure~\ref{fig:brgan_crgan}(d)) and train it along with $G_{\bm{x}}$ using the following objective function.
\begin{flalign}
  \label{eqn:crgan}
  & \mathcal{L}_{\text{CR-GAN}} = \mathbb{E}_{\bm{y}^r \sim p^r(\bm{y})} [\log D_{\bm{y}}(\bm{y}^r)]
  \nonumber \\
  & + \mathbb{E}_{\bm{z}_{\bm{x}} \sim p(\bm{z}_{\bm{x}}), \bm{z}_{q} \sim p(\bm{z}_{q})} [\log ( 1 - D_{\bm{y}} (\psi^g (G_{\bm{x}}(\bm{z}_{\bm{x}}), G_{q}(\bm{z}_{q})) ) )],
\end{flalign}
where we use a differentiable JPEG $\psi^g$~\cite{RShinNIPSW2017,PKorusCVPR2019}, which approximates the non-differentiable rounding operation in JPEG using a differentiable continuous function.
The optimization is performed similar to that in Equation~\ref{eqn:brgan}.

In a typical setting, JPEG is lossy even at the best quality of 100 owing to chroma subsampling in the Y'CbCr domain and rounding in the discrete cosine transform domain.
This can yield unexpected artifacts, which can disappear and become unrecognizable after compression.
To alleviate this problem, we introduce a \textit{masking architecture} (Figure~\ref{fig:brgan_crgan}(d)) that provides a bypass for producing a lossless image and adjusts the balance between compression and non-compression using the bypass as follows:
\begin{flalign}
  \label{eqn:crgan_mask}
  \bm{y}^g = {m}_{q} \hat{\bm{y}}^g + (1 - {m}_{q}) \bm{x}^g,
\end{flalign}
where $G_{q}$ is decomposed into a quality-factor generator ${q}^g = G_{q}^{q}(\bm{z}_{q})$ and a mask generator ${m}_{q} = G_{q}^{{m}_{q}}(\bm{z}_{q})$, $\bm{x}^g = G_{\bm{x}}(\bm{z}_{\bm{x}})$ is an uncompressed image, and $\hat{\bm{y}}^g$ is a compressed image produced by $\hat{\bm{y}}^g = \psi^g(\bm{x}^g, q^g)$.
${m}_{q}$ is a scalar within the range of $[0, 1]$, adjusting the balance between $\hat{\bm{y}}^g$ (i.e., compression) and $\bm{x}^g$ (i.e., non-compression).
In addition to $q^g$, ${m}_{q}$ is generated from $G_{q}$, and their distributions are optimized through training.

Similar to the masking architecture in BR-GAN (Section~\ref{sec:brgan}), this masking architecture is useful for suppressing unexpected artifacts, which are unrecognizable after compression, and allows only the necessary changes to be performed.

\section{Blur, noise, and compression robust GAN: BNCR-GAN}
\label{sec:bncrgan}

Based on BR-GAN (Section~\ref{sec:brgan}), NR-GAN~\cite{TKanekoCVPR2020}, and CR-GAN (Section~\ref{sec:crgan}), we consider \textit{blur, noise, and compression robust image generation}, which learns a clean image generator directly from images exhibiting all types of degradation (Equation~\ref{eqn:degradation}).
To achieve this, we replace the predefined degradation simulation models of AmbientGAN (Figure~\ref{fig:bncrgan}(a)) with trainable generators (i.e., a blur-kernel generator $G_{\bm{k}}$, a noise generator $G_{\bm{n}}$, and a quality-factor generator $G_{q}$) in BNCR-GAN (Figure~\ref{fig:bncrgan}(b)) and train them with a clean image generator $G_{\bm{x}}$ using the following objective function.
\begin{flalign}
  \label{eqn:bncrgan}
  & \mathcal{L}_{\text{BNCR-GAN}} = \mathbb{E}_{\bm{y}^r \sim p^r(\bm{y})} [\log D_{\bm{y}}(\bm{y}^r)]
  \nonumber \\
  & + \mathbb{E}_{\bm{z}_{\bm{x}} \sim p(\bm{z}_{\bm{x}}), \bm{z}_{\bm{k}} \sim p(\bm{z}_{\bm{k}}), \bm{z}_{\bm{n}} \sim p(\bm{z}_{\bm{n}}), \bm{z}_{q} \sim p(\bm{z}_{q})}
  \nonumber \\
  & [\log ( 1 - D_{\bm{y}} (\psi^g (G_{\bm{x}}(\bm{z}_{\bm{x}}) * G_{\bm{k}}(\bm{z}_{\bm{k}}) + G_{\bm{n}}(\bm{z}_{\bm{n}}), G_{q}(\bm{z}_{q})) ) )],
\end{flalign}
The optimization is performed similar to that represented in Equation~\ref{eqn:brgan}.

In this unified model, dealing with the uncertainty caused by combining multiple irreversible processes poses a challenge because $\mathcal{L}_{\text{BNCR-GAN}}$ only regularizes the final output $\bm{y}^g$ and imposes no regularization on each process.
To address this, we introduce \textit{adaptive consistency (AC) losses} $\mathcal{L}_{\text{AC}} = \mathcal{L}_{\text{AC}}^{\text{blur}} + \mathcal{L}_{\text{AC}}^{\text{comp}}$, suppressing the changes between irreversible blur and compression processes according to their strengths.
\begin{flalign}
  \label{eqn:blur_consistency}
  \mathcal{L}_{\text{AC}}^{\text{blur}} =
  & \: \mathbb{E}_{\bm{z}_{\bm{x}} \sim p(\bm{z}_{\bm{x}}), \bm{z}_{\bm{k}} \sim p(\bm{z}_{\bm{k}})} [e^{- \mu_{\bm{k}} H(G_{\bm{k}}(\bm{z}_{\bm{k}}))}
  \nonumber \\
  & \: \| G_{\bm{x}}(\bm{z}_{\bm{x}}) - G_{\bm{x}}(\bm{z}_{\bm{x}}) * G_{\bm{k}}(\bm{z}_{\bm{k}}) \|^2],
  \\
  \label{eqn:compression_consistency}
  \mathcal{L}_{\text{AC}}^{\text{comp}} =
  & \: \mathbb{E}_{\bm{z}_{\bm{x}} \sim p(\bm{z}_{\bm{x}}), \bm{z}_{q} \sim p(\bm{z}_{q})} [e^{- \mu_{q} \frac{100 - G_{q}(\bm{z}_{q})}{100}}
  \nonumber \\
  & \: \| G_{\bm{x}}(\bm{z}_{\bm{x}}) - \psi^g (G_{\bm{x}}(\bm{z}_{\bm{x}}), G_{q}(\bm{z}_{q})) \|^2],
\end{flalign}
where $H(G_{\bm{k}}(\bm{z}_{\bm{k}}))$ is the entropy of the blur kernel, and $\mu_{\bm{k}}$ and $\mu_{q}$ are scale parameters.
The weight term of $\mathcal{L}_{\text{AC}}^{\text{blur}}$ (i.e., $e^{- \mu_{\bm{k}} H(G_{\bm{k}}(\bm{z}_{\bm{k}}))}$) increases as the generated kernel $G_{\bm{k}}(\bm{z}_{\bm{k}})$ becomes closer to the identity (or non-blur) kernel.
The weight term of $\mathcal{L}_{\text{AC}}^{\text{comp}}$ (i.e., $e^{- \mu_{q} \frac{100 - G_{q}(\bm{z}_{q})}{100}}$) increases as the generated quality factor $G_{q}(\bm{z}_{q})$ approaches 100.
Namely, $\mathcal{L}_{\text{AC}}$ weighs the consistency when blur and compression are weak.
In our implementation, the gradients were propagated only for the left term $G_{\bm{x}}$ (an image before blur/compression) in $\mathcal{L}_{\text{AC}}^{\text{blur}}$ and $\mathcal{L}_{\text{AC}}^{\text{comp}}$, and not for the right term (an image after blur/compression) because the right term could be regularized by the adversarial loss, whereas the left term could not be regularized in our training settings, in which clean images were not available for training.

\section{Experiments}
\label{sec:experiments}

\subsection{Experimental settings in comparative studies}
\label{subsec:experimental_settings}

To advance research on blur, noise, and compression robust image generation, we first conducted large-scale comparative studies on blur robust (Section~\ref{subsec:eval_cifar10_blur}), compression robust (Section~\ref{subsec:eval_cifar10_compression}), and blur, noise, and compression robust (Section~\ref{subsec:eval_cifar10_all}) image generation.
In this section, we describe the common experimental settings and provide the individual settings and results in Sections~\ref{subsec:eval_cifar10_blur}--\ref{subsec:eval_cifar10_all}.

\begin{table*}[tb]
  \centering
  \scriptsize{
    \begin{tabularx}{\textwidth}{ccCCCCCCCccCCCCCCC}
      \addlinespace[-\aboverulesep]
      \cmidrule[\heavyrulewidth](r){1-9}
      \cmidrule[\heavyrulewidth]{10-18}
      \multirow{2}{*}{\textsc{No.}} & \multirow{2}{*}{\textsc{Model}}
      & (A) & (B) & (C) & (D) & (E) & (F) & (G)
      & \multirow{2}{*}{\textsc{No.}} & \multirow{2}{*}{\textsc{Model}}
      & (A) & (H) & (I) & (J) & (K) & (L) & (M)
      \\ \cmidrule(rl){3-3} \cmidrule(rl){4-6} \cmidrule(rl){7-9} \cmidrule(rl){12-12} \cmidrule(rl){13-15} \cmidrule(l){16-18}
      & & \tiny{\textsc{Clean}} & \tiny{\textsc{Focus}} & \tiny{\textsc{Motion}} & (B)/(C) & $\frac{1}{4}$(D) & $\frac{1}{2}$(D) & $\frac{3}{4}$(D)
      & & & \tiny{\textsc{Clean}} & 60--80 & 80--100 & 60--100 & $\frac{1}{4}$(J) & $\frac{1}{2}$(J) & $\frac{3}{4}$(J)
      \\
      \cmidrule[\lightrulewidth](r){1-9}
      \cmidrule[\lightrulewidth]{10-18}
      1 & GAN
      & \second{18.5} & 47.9 & 36.1 & 43.1 & 22.4 & 27.4 & 34.1 &
      9 & GAN
      & \best{18.5} & 41.7 & 27.8 & 33.3 & 21.0 & 24.7 & 29.0
      \\
      2 & AmbientGAN$^\dag$
      &  --  & \second{25.5} & \second{23.0} & \second{24.2} & \second{18.9} & \second{20.7} & \second{21.8} &
      10 & AmbientGAN$^\dag$
      &  --  & 43.9 & 36.6 & 40.5 & 20.6 & 22.8 & \second{24.5}
      \\
      3 & P-AmbientGAN
      & \second{18.5} & 48.9 & 35.7 & 41.5 & 22.5 & 27.5 & 34.1 &
      11 & P-AmbientGAN
      & 35.7 & 42.8 & 36.5 & 42.0 & 34.1 & 30.2 & 36.9
      \\
      \cmidrule[\lightrulewidth](r){1-9}
      \cmidrule[\lightrulewidth]{10-18}
      4 & BR-GAN
      & \best{17.6} & \best{24.8} & \best{22.9} & \best{23.4} & \best{18.4} & \best{20.2} & \best{21.7} &
      12 & CR-GAN
      & 18.8 & \second{34.0} & \second{25.4} & \best{26.3} & \best{20.1} & \second{22.7} & \second{24.5}
      \\
      5 & BR-GAN w/o mask
      & 21.9 & 27.6 & 25.4 & 26.6 & 22.5 & 22.9 & 24.0 &
      13 & CR-GAN w/o mask
      & 34.6 & 41.0 & 35.9 & 38.6 & 33.9 & 31.1 & 34.2
      \\
      6 & BNCR-GAN
      & 18.6 & 28.0 & 26.4 & 26.8 & 20.2 & 21.4 & 24.1 &
      14 & BNCR-GAN
      & \second{18.6} & \best{29.9} & \best{24.6} & \second{26.6} & \second{20.3} & \best{22.4} & \best{24.4}
      \\
      7 & BNCR-GAN w/o $\mathcal{L}_{\text{AC}}$
      & 20.5 & 28.6 & 27.2 & 27.3 & 23.4 & 22.1 & 24.2 &
      15 & BNCR-GAN w/o $\mathcal{L}_{\text{AC}}$
      & 20.5 & 36.1 & 30.6 & 33.4 & 22.4 & 24.1 & 30.6
      \\
      \cmidrule[\lightrulewidth](r){1-9}
      \cmidrule[\lightrulewidth]{10-18}
      8 & Deblur+GAN$^\dag$
      &  --  & 31.4 & 24.8 & 28.0 & 19.8 & 22.8 & 25.2 &
      16 & Deblock+GAN
      &  --  & 39.9 & 28.1 & 33.3 & 21.4 & 25.4 & 28.5
      \\
      \cmidrule[\heavyrulewidth](r){1-9}
      \cmidrule[\heavyrulewidth]{10-18}
      \addlinespace[-\belowrulesep] 
    \end{tabularx}
  }
  \caption{\textbf{Comparison of FID$\downarrow$ on \textsc{CIFAR-10} under blur settings (left) and compression settings (right).}
    Smaller values are preferable.
    We report the median score across three random seeds.
    Bold and bold italic fonts indicate the best and second-best scores, respectively.
    The symbol $^\dag$ indicates that the models were trained under advantageous conditions (i.e., trained using the ground-truth image degradation\protect\footnotemark).}
  \label{table:eval_cifar10_blur_compression}
  \vspace{-4mm}
\end{table*}

\smallskip\noindent\textbf{Dataset.}
In these studies, we used \textsc{CIFAR-10}~\cite{AKrizhevskyTech2009}, which includes $60k$ $32 \times 32$ natural images, split into $50k$ training and $10k$ test images, which are commonly used to examine the benchmark performance of generative models (it is also in the studies of AmbientGAN~\cite{ABoraICLR2018} and NR-GAN~\cite{TKanekoCVPR2020}). In addition, the image size was reasonable for a large-scale comparative study.
We demonstrate the versatility on a more complex dataset in Section~\ref{subsec:eval_ffhq}.

\smallskip\noindent\textbf{Metrics.}
Following the NR-GAN study~\cite{TKanekoCVPR2020}, we used the Fr\'{e}chet inception distance (FID)~\cite{MHeuselNIPS2017}, which assesses the distance between real and generative distributions.
The validity of this measure has been shown in large-scale studies on GANs~\cite{MLucicNeurIPS2018,KKurachICML2019}.
Its sensitivity to image degradation has been also demonstrated~\cite{MHeuselNIPS2017,TKanekoCVPR2020}.
On this metric, a smaller value is preferred.

\smallskip\noindent\textbf{Implementation.}
We implemented GANs following the NR-GAN study~\cite{TKanekoCVPR2020}.
Specifically, we used ResNet architectures~\cite{KHeCVPR2016} for $G_{\bm{x}}$ and $D_{\bm{y}}$ and optimized them using a non-saturating GAN loss~\cite{IGoodfellowNIPS2014} with a real gradient penalty regularization~\cite{LMeschederICML2018}.
We used similar architectures for $G_{\bm{x}}$ and $G_{\bm{n}}$.
With regard to $G_{\bm{k}}$ and $G_{q}$, we used two-hidden-layer multilayer perceptrons.
Inspired by the findings in~\cite{TKanekoCVPR2020}, we imposed diversity-sensitive regularization~\cite{DYangICLR2019,QMaoCVPR2019} on $G_{\bm{k}}$, $G_{\bm{n}}$, and $G_{q}$.
As we aim to construct a generative model applicable to diverse degradation settings without specific tuning, we examined the performance for fixed hyperparameters values across all experiments.

\subsection{Evaluation on blur robust image generation}
\label{subsec:eval_cifar10_blur}

\smallskip\noindent\textbf{Degradation settings.}
We tested three full-blur settings; here, all images were degraded by (B) out-of-focus blur with a disk kernel of radius $r \in [ 0.5, 2 ]$, (C) motion blur~\cite{GBoracchiTIP2012,OKupynCVPR2018} with a trajectory length of 5 and exposure time $T \in \{ 0.1, 0.5, 1\}$, and (D) either (B) or (C) with a selection rate of 0.5.
Additionally, we tested three partial blur settings, in which one-quarter (E), one-half (F), and three-quarters (G) of images were blurred by setting (D), while the remaining were clean images.
For reference, we also examined the performance of clean images (A).

\smallskip\noindent\textbf{Compared models.}
To examine the performance of our \textit{BR-GAN} and \textit{BNCR-GAN}, we compared them with three baseline GANs: standard \textit{GAN}, which is agnostic to image degradation, \textit{AmbientGAN}, with a ground-truth degradation simulation model, and \textit{P-AmbientGAN}, which learns a degradation parameter (e.g., $\bm{k}$ and $q$) directly without using a generator.
Furthermore, we conducted ablation studies on two models, including \textit{BR-GAN w/o mask} and \textit{BNCR-GAN w/o $\mathcal{L}_{\text{AC}}$}.
We ablated the key components of our proposals (i.e., masking architectures and adaptive consistency losses).
We also examined the performance of GAN trained with deblurred images (\textit{Deblur+GAN}).
We used a model-based deblurring method because typical learning-based methods cannot be trained in our setting, in which clean images were not available.
To examine the upper-bound performance, we used a non-blind deblurring method~\cite{DKrishnanNIPS2009}, which deblurred using the ground-truth kernel.

\smallskip\noindent\textbf{Results.}
The results are summarized in Table~\ref{table:eval_cifar10_blur_compression} (left).
Our main findings are as follows:
\textit{(i) Comparison with baseline GANs (Nos. 1--4, 6).}
As expected, AmbientGAN showed reasonable performance owing to ground-truth blur information.
We found that BR-GAN, which must estimate such information through training, achieved competitive performance.
BNCR-GAN, which should learn \textit{no noise} and \textit{non-compression} in this task, was inferior to these two models but outperformed standard GAN under all blur settings (B--G).
P-AmbientGAN showed poor performance because it could not represent blur diversity, which is included in all blur settings (B--G).
\textit{(ii) Comparison with Deblur+GAN (Nos. 4, 6, 8).}
We found that BR-GAN outperformed Deblur+GAN under all blur settings (B--G), and BNCR-GAN outperformed Deblur+GAN in the majority of cases (4/6).
\textit{(iii) Ablation study on BR-GAN (Nos. 4, 5).}
We confirmed that the masking architecture was useful for boosting the performance under all settings (A--G).
\textit{(iv) Ablation study on BNCR-GAN (Nos. 6, 7).}
We found that $\mathcal{L}_{\text{AC}}$ was effective when parts or all of images were clean (A and E), and the negative effect did not exist across all settings (A--G).
\footnotetext{Specifically, the ground-truth information is linked to the individual image in Deblur+GAN, while this link is absent in AmbientGAN.}

\subsection{Evaluation on compression robust image generation}
\label{subsec:eval_cifar10_compression}

\smallskip\noindent\textbf{Degradation settings.}
We tested three full compression settings; here, all images were compressed with a quality factor within the ranges of $[60, 80]$ (H), $[80, 100]$ (I), and $[60, 100]$ (J).
We also tested three partial compression settings, in which one-quarter (K), one-half (L), and three-quarters (M) of images were compressed by setting (J), while the remaining images were clean.
For reference, we also examined the performance of clean images (A).

\smallskip\noindent\textbf{Compared models.}
In addition to the abovementioned models, we examined the performance of \textit{CR-GAN}, \textit{CR-GAN w/o mask}, and \textit{Deblock+GAN} (GAN trained with deblocked images).
In Deblock+GAN, we used a model-based deblocking method~\cite{AFoiTIP2007} for the same reason mentioned in Section~\ref{subsec:eval_cifar10_blur}.

\smallskip\noindent\textbf{Results.}
The results are presented in Table~\ref{table:eval_cifar10_blur_compression} (right).
Our key findings are as follows:
\textit{(i) Comparison with baseline GANs (Nos. 9--12, 14).}
CR-GAN and BNCR-GAN achieved the best or second-best scores under all compression settings (H--M).
AmbientGAN was comparable to them when compression was applied to parts of images (K--M), while it underperformed them by a large margin (over 9.9) when compression was applied to all images (H--J).
We consider that this occurred because of the lossy characteristics of JPEG (it remains lossy even at the best quality (Section~\ref{sec:crgan})), allowing $G_{\bm{x}}$ to create unexpected artifacts, which were unrecognizable after the application of JPEG.
By contrast, in CR-GAN, the masking architecture provides a bypass to a lossless image.
This was useful for suppressing such artifacts.
P-AmbientGAN showed poor performance in all compression settings (H--M) because it could not handle compression variety.
\textit{(ii) Comparison with Deblock+GAN (Nos. 12, 14, 16).}
We confirmed that CR-GAN and BNCR-GAN outperformed Deblock+GAN under all compression settings (H--M).
\textit{(iii) Ablation study on CR-GAN (Nos. 12, 13).}
In all settings (A and H--M), the masking architecture resulted in a positive effect.
\textit{(iv) Ablation study on BNCR-GAN (Nos. 14, 15).}
We found that $\mathcal{L}_{\text{AC}}$ improved performance under all settings (A and H--M).\footnote{We found that BNCR-GAN was better than CR-GAN in some cases (e.g., setting (H)).
  We consider that $\mathcal{L}_{\text{AC}}^{\text{comp}}$, which was used in BNCR-GAN but not in CR-GAN, resulted in this behavior.
  To verify this statement, we examined the performance of CR-GAN with $\mathcal{L}_{\text{AC}}^{\text{comp}}$ and found that in setting (H), CR-GAN with $\mathcal{L}_{\text{AC}}^{\text{comp}}$ was comparable to BNCR-GAN and outperformed CR-GAN without $\mathcal{L}_{\text{AC}}^{\text{comp}}$.
  See Appendix~\ref{subsubsec:eval_crgan_acloss} for details of this experiment.}

\subsection{Evaluation on blur, noise, and compression robust image generation}
\label{subsec:eval_cifar10_all}

\smallskip\noindent\textbf{Degradation settings.}
We tested a full degradation setting (N) where all images were blurred by setting (D), noised by the setting of~\cite{TBrooksCVPR2019} (which consists of read and shot noise that simulated noise from a real noise dataset~\cite{TPlotzCVPR2017}), and compressed by setting (J).
We also analyzed three partial degradation settings, in which one-quarter (O), one-half (P), and three-quarters (Q) of images were independently blurred, noised, and compressed by setting (N).
Here, ``independently'' implies that the target image of each degradation (i.e., blur, noise, or compression) was randomly selected independently of the other degradation.
For reference, we also tested clean images (A).

\smallskip\noindent\textbf{Compared models.}
In addition to the abovementioned models, we examined the performance of NR-GAN~\cite{TKanekoCVPR2020} to clarify its limitations.

\begin{table}[tb]
  \scriptsize{  
    \begin{tabularx}{\columnwidth}{ccCCCCC}
      \toprule
      \multirow{2}{*}{\textsc{No.}} & \multirow{2}{*}{\textsc{Model}}
      & (A) & (N) & (O) & (P) & (Q)
      \\ \cmidrule(r){3-3} \cmidrule(rl){4-4} \cmidrule(l){5-7}
      & & \textsc{Clean} & \textsc{All} & $\frac{1}{4}$(N) & $\frac{1}{2}$(N) & $\frac{3}{4}$(N)
      \\ \midrule
      17 & GAN
      & \second{18.5} & 58.0 & 24.6 & 34.3 & 45.5
      \\
      18 & AmbientGAN$^\dag$
      &  --  & 46.8 & \second{22.1} & \best{25.6} & 37.3
      \\ \midrule
      19 & BR-GAN
      & \best{17.6} & 52.3 & 22.4 & 31.6 & 41.5
      \\
      20 & NR-GAN
      & 20.0 & 58.1 & 26.1 & 34.6 & 44.3
      \\
      21 & CR-GAN
      & 18.8 & 51.1 & 26.5 & 36.8 & 48.2
      \\ \midrule
      22 & BNCR-GAN
      & 18.6 & \best{34.1} & \best{22.0} & \second{25.7} & \best{29.4}
      \\
      23 & BNCR-GAN w/o $\mathcal{L}_{\text{AC}}$
      & 20.5 & \second{41.0} & 25.3 & 28.6 & \second{34.6}
      \\ \bottomrule
    \end{tabularx}
  }
  \caption{\textbf{Comparison of FID$\downarrow$ on \textsc{CIFAR-10} under blur, noise, and compression settings.}
    The score calculation method and notation are the same as those in Table~\ref{table:eval_cifar10_blur_compression}.}
  \label{table:eval_cifar10_all}
  \vspace{-4mm}
\end{table}

\smallskip\noindent\textbf{Results.}
Table~\ref{table:eval_cifar10_all} presents the results.
Our main findings are as follows:
\textit{(i) Comparison with baseline GANs (Nos. 17, 18, 22).}
BNCR-GAN achieved the best or competitive scores in comparison to GAN and AmbientGAN.
In particular, we found that BNCR-GAN outperformed AmbientGAN by a large margin (over 7.9) when the rate of degraded images was relatively high (N and Q).
We consider that the lossy characteristics of JPEG affected the results, similar to those observed in Section~\ref{subsec:eval_cifar10_compression}.
\textit{(ii) Comparison with single-degradation robust GANs (Nos. 19--22).}
In all degradation settings (N--Q), BNCR-GAN outperformed BR-GAN, NR-GAN, and CR-GAN.
These results demonstrate the limitations of the single-degradation robust GANs on the datasets that contain several types of degradation.
\textit{(iii) Ablation study on BNCR-GAN (Nos. 22, 23).}
We confirmed that $\mathcal{L}_{\text{AC}}$ contributed to performance enhancement in all settings (A and N--Q).

\smallskip\noindent\textbf{Summary.}
Through the three comparative studies (Sections~\ref{subsec:eval_cifar10_blur}--\ref{subsec:eval_cifar10_all}, in which we tested 134 total conditions), we found that BR-GAN, CR-GAN, and BNCR-GAN achieved better or competitive performance than AmbientGAN despite the disadvantageous training conditions.
In particular, we confirmed that the key components of our proposals (i.e., masking architectures and adaptive consistency losses) contributed to achieving such high performance.
In Section~\ref{subsec:eval_cifar10_all}, we also demonstrate the limitations of single-degradation robust GANs and the importance of BNCR-GAN in the datasets exhibiting several types of degradation.

\smallskip\noindent\textbf{Further analysis.}
As a further analysis, we evaluated BNCR-GAN on noise robust image generation in the setting of~\cite{TBrooksCVPR2019}.
We found that FID obtained by BNCR-GAN (20.5) was comparable to those obtained by AmbientGAN (20.0) and NR-GAN (20.3).
We provide details of this experiment and other further analyses in Appendix~\ref{subsec:analyses_cifar10}.

\subsection{Evaluation on complex dataset}
\label{subsec:eval_ffhq}

\smallskip\noindent\textbf{Experimental settings.}
Inspired by the findings of recent large-scale studies on GANs~\cite{KKurachICML2019} and NR-GAN~\cite{TKanekoCVPR2020}, we also investigated the performance of the proposed models on a more complex dataset, i.e., the $128 \times 128$ version of \textsc{FFHQ}~\cite{TKarrasCVPR2019}, which includes $70k$ face images, comprising $60k$ training and $10k$ test images.\footnoteref{foot:lsun}
To focus on the discussion, we compared the representative models in three representative settings in Sections~\ref{subsec:eval_cifar10_blur}--\ref{subsec:eval_cifar10_all} (D, J, and P).\footnote{According to the change in image size, we enlarged the blur size.
  We used out-of-focus blur with a disk kernel of radius $r \in [0.5, 4]$ and motion blur with a trajectory length of 10 and exposure time $T \in \{ 0.1, 0.5, 1 \}$.}

\begin{table}[tb]
  \scriptsize{    
    \begin{tabularx}{\columnwidth}{cCcCcC}
      \addlinespace[-\aboverulesep]
      \cmidrule[\heavyrulewidth](r){1-2}
      \cmidrule[\heavyrulewidth](r){3-4}
      \cmidrule[\heavyrulewidth]{5-6}
      \textsc{Model} & (D) \scriptsize{\textsc{Blur}}
      & \textsc{Model} & (J) \scriptsize{\textsc{Comp.}}
      & \textsc{Model} & (P) \scriptsize{\textsc{All}}
      \\
      \cmidrule[\lightrulewidth](r){1-2}
      \cmidrule[\lightrulewidth](r){3-4}
      \cmidrule[\lightrulewidth]{5-6}
      GAN & 34.1
      & GAN & 45.5
      & GAN & 34.9
      \\
      AmbientGAN$^\dag$ & \second{27.7}
      & AmbientGAN$^\dag$ & \second{27.0}
      & AmbientGAN$^\dag$ & \second{24.5}
      \\
      \cmidrule[\lightrulewidth](r){1-2}
      \cmidrule[\lightrulewidth](r){3-4}
      \cmidrule[\lightrulewidth]{5-6}
      BR-GAN & \best{25.3}
      & CR-GAN & \best{25.7}
      & BNCR-GAN & \best{24.2}
      \\
      \cmidrule[\lightrulewidth](r){1-2}
      \cmidrule[\lightrulewidth](r){3-4}
      \cmidrule[\heavyrulewidth]{5-6}
      Deblur+GAN$^\dag$ & 32.9
      & Deblock+GAN & 50.0
      \\
      \cmidrule[\heavyrulewidth](r){1-2}
      \cmidrule[\heavyrulewidth](r){3-4}
      \addlinespace[-\belowrulesep] 
    \end{tabularx}
  }
  \caption{\textbf{Comparison of FID$\downarrow$ on \textsc{FFHQ}.}
    Smaller values are better.
    Experiments were conducted once owing to the time-consuming training.
    The notation is the same as that in Table~\ref{table:eval_cifar10_blur_compression}.}
  \label{table:eval_ffhq}
  \vspace{-2mm}
\end{table}

\begin{figure}[tb]
  \centering
  \includegraphics[width=\columnwidth]{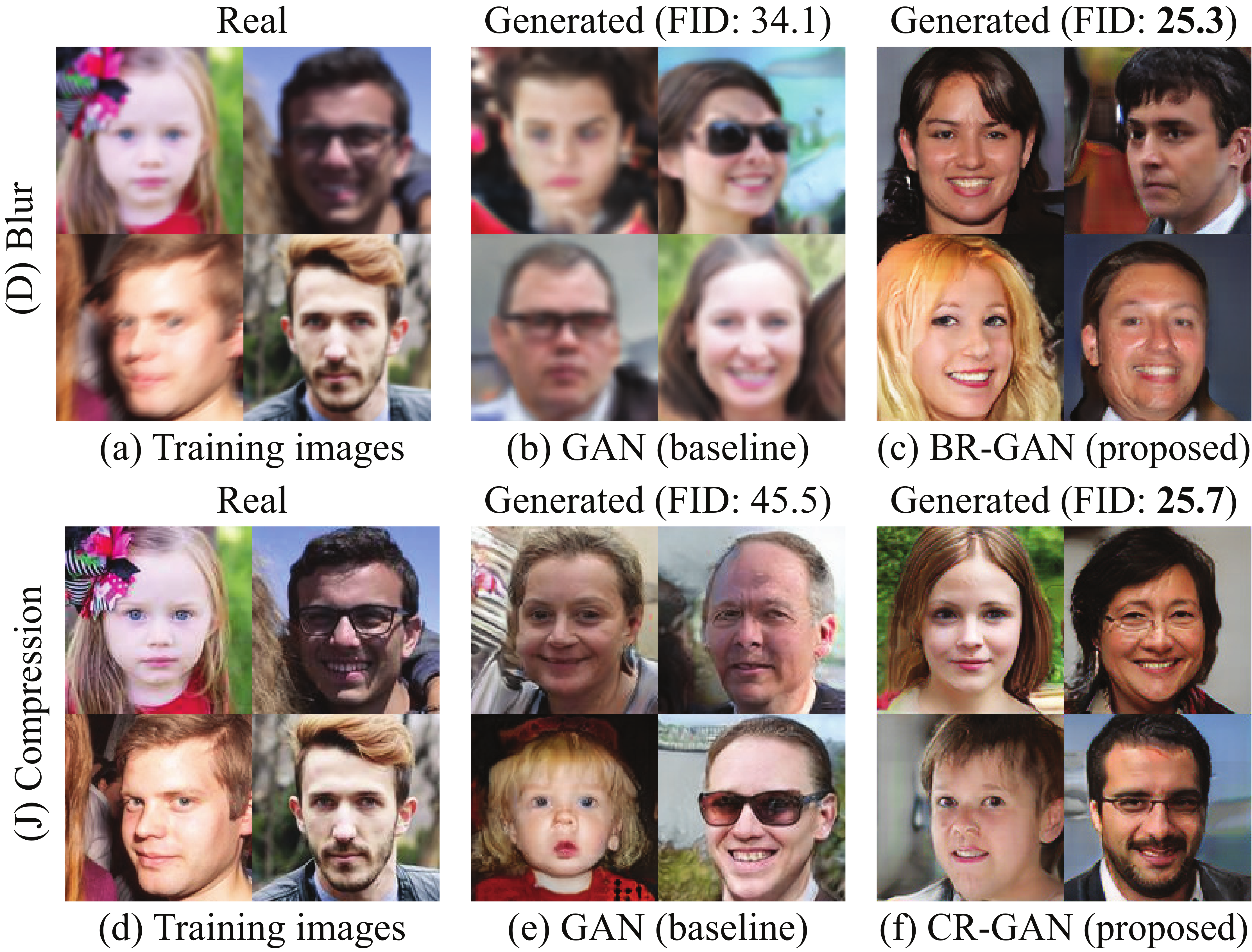}
  \caption{\textbf{Examples of blur robust image generation (upper) and compression robust image generation (bottom).}
    Best viewed at high resolution.
    We provide examples of blur, noise, and compression robust image generation
    (setting (P)) in Figure~\ref{fig:examples_ffhq_all05_select}.
    We provide additional results in Appendix~\ref{subsec:examples_ffhq}.}
    \label{fig:examples_ffhq_blur_compression_select}
    \vspace{-4mm}
\end{figure}

\smallskip\noindent\textbf{Results.}
Table~\ref{table:eval_ffhq} lists the results.
We found that there was a similar tendency in this dataset: BR-GAN, CR-GAN, and BNCR-GAN achieved better or competitive performance compared with AmbientGAN, which was trained under advantageous conditions, and outperformed the other baselines (standard GAN, Deblur+GAN, and Deblock+GAN).\footnote{We found that Deblock+GAN underperformed standard GAN in this case.
  A drawback of model-based deblocking methods is over-smoothing, and we suppose that this degraded FID more than compression artifacts.}
We show examples of generated images for settings (D and J) in Figure~\ref{fig:examples_ffhq_blur_compression_select} and those for setting (P) in Figure~\ref{fig:examples_ffhq_all05_select}.

\subsection{Application to image restoration}
\label{subsec:restoration}

The recently proposed UNIR~\cite{APajotICLR2019}, a conditional extension of AmbientGAN~\cite{ABoraICLR2018}, can learn an image restoration model directly from degraded images without paired or set-level supervision.
However, it requires that degradation simulation models (i.e., $p^r(\bm{k})$, $p^r(\bm{n})$, and $p^r(q)$) be predefined, similar to AmbientGAN.
A reasonable solution was to utilize the blur-kernel, noise, and quality-factor generators obtained by BNCR-GAN, instead of the predefined degradation simulation models.
In this section, we evaluated this approach (\textit{BNCR-GAN+UNIR}).

\smallskip\noindent\textbf{Experimental settings.}
We trained the models under the same degradation settings used in Section~\ref{subsec:eval_cifar10_all} (settings (N--Q)).
In the evaluation, we considered that test images were degraded in the most severe setting (setting (N)).
With regard to BNCR-GAN+UNIR, we used the degradation generators that were obtained in the experiments described in Section~\ref{subsec:eval_cifar10_all}.
We compared BNCR-GAN+UNIR with \textit{UNIR} with predefined degradation models~\cite{APajotICLR2019} and \textit{CycleGAN} with set-level supervision \cite{JYZhuICCV2017}.\footnote{We did not examine the performance of CycleGAN in setting (N) because it cannot be used when clean images were not available.}

\begin{table}
  \centering
  \scriptsize{
    \begin{tabularx}{\columnwidth}{cCCCC}
      \toprule
      \multirow{2}{*}{\textsc{Model}}
      & (O) & (P) & (Q) & (N)
      \\ \cmidrule{2-5}
      & $\frac{1}{4}$(N) & $\frac{1}{2}$(N) & $\frac{3}{4}$(N) & \textsc{All}
      \\ \midrule
      UNIR$^\dag$
      & \second{28.4} / \second{0.0204}
      & \second{22.4} / \best{0.0155}
      & \second{21.0} / \best{0.0137}
      & \best{24.8} / \best{0.0146}
      \\ \midrule
      BNCR-GAN+UNIR
      & \best{27.7} / \best{0.0184}
      & \best{21.5} / \second{0.0172}
      & \best{20.4} / \second{0.0152}
      & \best{24.8} / \second{0.0206}
      \\ \midrule
      CycleGAN$^\ddag$
      & 30.2 / 0.0226
      & 27.1 / 0.0177
      & 33.7 / 0.0259
      & N/A / N/A
      \\ \bottomrule
    \end{tabularx}
  }
  \caption{\textbf{Comparison of FID$\downarrow$ and LPIPS$\downarrow$ on \textsc{CIFAR-10} in image restoration.}
    The results are listed as FID$\downarrow$ / LPIPS$\downarrow$.
    We report the scores for the model that achieved the median FID over three training runs.
    Bold and bold italic fonts indicate the best and second-best scores, respectively.
    The symbols $^\dag$ and $^\ddag$ indicate that the models require additional supervision (i.e., predefined degradation models and set-level supervision, respectively).}
  \label{table:eval_restoration}
  \vspace{-3mm}
\end{table}

\smallskip\noindent\textbf{Results.}
We report FID and learned perceptual image patch similarity (LPIPS)~\cite{RZhangCVPR2018}\footnote{It is known that typical metrics, e.g., PSNR and SSIM~\cite{ZWangTIP2004}, are inconsistent with human visual perception~\cite{RZhangCVPR2018}.
  Hence, we used LPIPS, which was proposed as an alternative, along with FID.
  As a reference, we report PSNR and SSIM in Appendix~\ref{subsec:analyses_image_restoration}} in Table~\ref{table:eval_restoration}.
We found that the set-level supervision-based model (CycleGAN) was susceptible to data distribution, while degradation model-based models (i.e., UNIR and BNCR-GAN+UNIR) constantly performed better in all settings.
Between UNIR and BNCR-GAN+UNIR, superiority or inferiority depended on the settings and metrics.
These results indicate the potential use of BNCR-GAN+UNIR in image restoration, specifically when only degraded images are available for training.

\section{Conclusion}
\label{sec:conclusion}

We have presented a new variant of GANs called BNCR-GAN to achieve blur, noise, and compression robust image generation without knowledge of degradation parameters.
The two sub-variants, BR-GAN and CR-GAN, used a two-generator model composed of an image and blur-kernel/quality-factor generators.
In particular, we devised masking architectures to adjust the blur and compression strengths using bypasses before and after degradation.
In BNCR-GAN, which is a unified model integrating BR-GAN, NR-GAN, and CR-GAN, we developed adaptive consistency losses to mitigate the uncertainty caused by the combination.
We examined the effectiveness of the proposed models through comparative studies and a generality analysis on two benchmark datasets.
We also demonstrated the applicability of BNCR-GAN in image restoration.
We hope that our findings will facilitate the creation of generative models on real-world datasets that may contain several types of image degradation.

\smallskip\noindent\textbf{Acknowledgment.}
This work was partially supported by JST AIP Acceleration Research Grant Number JPMJCR20U3, JST CREST Grant Number JPMJCR2015, and JSPS KAKENHI Grant Number JP19H01115.

\clearpage

{\small
\bibliographystyle{ieee_fullname}
\bibliography{refs}
}

\clearpage
\appendix

\section*{Appendix}

In this appendix, we provide further analyses, examples of generated and restored images, and implementation details in Appendices~\ref{sec:further_analysis}, \ref{sec:examples}, and \ref{sec:details}, respectively.

\section{Further analyses}
\label{sec:further_analysis}

\subsection{Further analyses on CIFAR-10}
\label{subsec:analyses_cifar10}

\subsubsection{Evaluation on noise robust image generation}
\label{subsubsec:eval_cifar10_noise}

In Section~\ref{subsec:eval_cifar10_all}, we briefly explain the evaluation of noise robust image generation because of space limitations.
For clarification, we provide details of the experiment and an additional analysis in this section.

\smallskip\noindent\textbf{Degradation settings.}
We tested the noise setting of~\cite{TBrooksCVPR2019} (R), which consists of read and shot noise that simulated noise from a real noise dataset~\cite{TPlotzCVPR2017}.
This noise setting is the same as that used in the evaluation of blur, noise, and compression robust image generation (Section~\ref{subsec:eval_cifar10_all}).
We also show the results with a clean image setting (A) as a reference.

\smallskip\noindent\textbf{Compared models.}
In addition to the models mentioned in Sections~\ref{subsec:eval_cifar10_blur}--\ref{subsec:eval_cifar10_all}, we examined the performance of \textit{Denoise+GAN} (GAN trained with denoised images).
As a denoiser, we used a model-based denoising method for the same reason mentioned in Section~\ref{subsec:eval_cifar10_blur}.
In particular, we used the CBM3D~\cite{KDabovICIP2007}, which is a commonly used model-based denoiser.
To investigate its upper-bound performance, we used ground-truth noise information when applying CBM3D to noisy images.

\smallskip\noindent\textbf{Results.}
The results are summarized in Table~\ref{table:eval_cifar10_noise}.
Our key findings are as follows:
\textit{(i) Comparison with baseline GANs (Nos. 1--4).}
Similar to the observations described in a previous study~\cite{TKanekoCVPR2020}, we found that AmbientGAN achieved the best performance owing to the advantageous training conditions under which the ground-truth noise information was provided, whereas NR-GAN, which must estimate such information through training, showed a competitive performance (with a difference of 0.3) between them.
We also found that BNCR-GAN, which further should estimate \textit{no blur} and \textit{no compression} simultaneously, was comparable to AmbientGAN (with a difference of 0.5) and NR-GAN (with a difference of 0.2), while outperforming GAN by a large margin (with a difference of 26.9).
\textit{(ii) Comparison with Denoise+GAN (Nos. 3, 4, 6).}
Similar to the findings of a previous study~\cite{TKanekoCVPR2020}, NR-GAN outperformed Denoise+GAN.
BNCR-GAN took over this feature and outperformed Denoise+GAN.
\textit{(iii) Ablation study on BNCR-GAN (Nos. 4, 5).}
We confirmed that $\mathcal{L}_{\text{AC}}$ was effective at improving the performance for this task, as well as for other tasks (i.e., blur robust image generation (Section~\ref{subsec:eval_cifar10_blur}), compression robust image generation (Section~\ref{subsec:eval_cifar10_compression}), and blur, noise, and compression robust image generation (Section~\ref{subsec:eval_cifar10_all})).

\begin{table}[tb]
  \centering
  \scriptsize{
    \begin{tabularx}{0.66\columnwidth}{ccCC}
      \toprule
      \multirow{2}{*}{\textsc{No.}} & \multirow{2}{*}{\textsc{Model}}
      & (A) & (R)
      \\ \cmidrule(r){3-3} \cmidrule(rl){4-4}
      & & \textsc{Clean} & \textsc{Noise}
      \\ \midrule
      1 & GAN
      & \best{18.5} & 47.4
      \\
      2 & AmbientGAN$^\dag$
      &  --  & \best{20.0}
      \\ \midrule
      3 & NR-GAN
      & 20.0 & \second{20.3}
      \\
      4 & BNCR-GAN
      & \second{18.6} & 20.5
      \\
      5 & BNCR-GAN w/o $\mathcal{L}_{\text{AC}}$
      & 20.5 & 23.0
      \\ \midrule
      6 & Denoise+GAN$^\dag$
      &  --  & 21.4
      \\ \bottomrule
    \end{tabularx}
  }
  \vspace{1mm}
  \caption{\textbf{Comparison of FID$\downarrow$ on \textsc{CIFAR-10} under noise settings.}
    Smaller values are preferable.
    The score calculation method and notation are the same as those in Table~\ref{table:eval_cifar10_blur_compression}.}
  \label{table:eval_cifar10_noise}
  \vspace{-4mm}
\end{table}

\begin{table*}[tb]
  \centering
  \scriptsize{
    \begin{tabularx}{\textwidth}{ccccCCCCCC}
      \toprule
      \multirow{2}{*}{\textsc{No.}} & \multirow{2}{*}{\textsc{Model}}
      & \multirow{2}{*}{$\mathcal{L}_{\text{AC}}^{\text{blur}}$}
      & \multirow{2}{*}{$\mathcal{L}_{\text{AC}}^{\text{comp}}$}
      & (A) & (D) & (R) & (J) & (N) & (P)
      \\ \cmidrule(r){5-5} \cmidrule(rl){6-6} \cmidrule(rl){7-7} \cmidrule(rl){8-8} \cmidrule(l){9-9} \cmidrule(l){10-10}
      & & & & \textsc{Clean} & \textsc{Blur} & \textsc{Noise} & \textsc{Comp.} & \textsc{All} & $\frac{1}{2}$(N)
      \\ \midrule
      1 & BNCR-GAN & \checkmark & \checkmark
      & \best{18.6} & \second{26.8} & \best{20.5} & \best{26.6} & \best{34.1} & \best{25.7}
      \\
      2 & BNCR-GAN w/o $\mathcal{L}_{\text{AC}}$ & &
      & 20.5 & 27.3 & 23.0 & 33.4 & 41.0 & 28.6
      \\
      3 & BNCR-GAN w/o $\mathcal{L}_{\text{AC}}^{\text{comp}}$ & \checkmark &
      & \best{18.6} & 27.3 & 21.5 & 29.7 & \second{38.1} & \best{25.7}
      \\
      4 & BNCR-GAN w/o $\mathcal{L}_{\text{AC}}^{\text{blur}}$ & & \checkmark
      & 20.0 & \best{25.7} & 21.3 & 30.7 & \second{38.1} & 26.4
      \\ \midrule
      5 & BNCR-GAN w/ $\mathcal{L}_{\text{C}}$
      & $\mathcal{L}_{\text{C}}^{\text{blur}}$
      & $\mathcal{L}_{\text{C}}^{\text{comp}}$
      & 19.3 & 31.1 & \best{20.5} & \second{27.9} & 41.2 & 27.7
      \\ \bottomrule
    \end{tabularx}
  }
  \caption{\textbf{Detailed ablation study on BNCR-GAN.}
    Smaller values are preferable.
    The score calculation method and notation are the same as those in Table~\ref{table:eval_cifar10_blur_compression}.
    The check mark \checkmark denotes that the corresponding loss was used.}
  \label{table:eval_ablation}
  \vspace{-4mm}
\end{table*}

\subsubsection{Detailed ablation study on BNCR-GAN}
\label{subsubsec:eval_ablation}

In ablation studies on BNCR-GAN described in the main text (Sections~\ref{subsec:eval_cifar10_blur}--\ref{subsec:eval_cifar10_all}), we ablated the \textit{sum} of the adaptive consistency losses, i.e., $\mathcal{L}_{\text{AC}} = \mathcal{L}_{\text{AC}}^{\text{blur}} + \mathcal{L}_{\text{AC}}^{\text{comp}}$.
For further clarification, we examined the performance when ablating the \textit{individual} loss, that is, either $\mathcal{L}_{\text{AC}}^{\text{blur}}$ or $\mathcal{L}_{\text{AC}}^{\text{comp}}$.
Furthermore, to investigate the effect of the \textit{adaptiveness} of $\mathcal{L}_{\text{AC}}$, we also examined the performance when disabling the adaptiveness.

\smallskip\noindent\textbf{Degradation settings.}
We analyzed the performance under diverse settings, including blur robust image generation (D), noise robust image generation (R), compression robust image generation (J), and blur, noise, and compression robust image generation (N and P).
As a reference, we also examined the performance under the clean image setting (A).

\smallskip\noindent\textbf{Compared models.}
In addition to the models mentioned in the main text (i.e., \textit{BNCR-GAN} (w/ $\mathcal{L}_{\text{AC}}$) and \textit{BNCR-GAN w/o $\mathcal{L}_{\text{AC}}$}), we examined the performance of \textit{BNCR-GAN w/o $\mathcal{L}_{\text{AC}}^{\text{blur}}$} and \textit{BNCR-GAN w/o $\mathcal{L}_{\text{AC}}^{\text{comp}}$}, in which $\mathcal{L}_{\text{AC}}^{\text{blur}}$ and $\mathcal{L}_{\text{AC}}^{\text{comp}}$ were ablated, respectively.
Furthermore, to investigate the effect of \textit{adaptiveness}, we also examined the performance of \textit{BNCR-GAN w/ $\mathcal{L}_{\text{C}}$}, which was trained with \textit{non-adaptive} consistency losses $\mathcal{L}_{\text{C}} = \mathcal{L}_{\text{C}}^{\text{blur}} + \mathcal{L}_{\text{C}}^{\text{comp}}$.
In this model, we used $\mathcal{L}_{\text{C}}^{\text{blur}}$ and $\mathcal{L}_{\text{C}}^{\text{comp}}$ as alternatives to $\mathcal{L}_{\text{AC}}^{\text{blur}}$ and $\mathcal{L}_{\text{AC}}^{\text{comp}}$, respectively.
In particular, in $\mathcal{L}_{\text{C}}^{\text{blur}}$ and $\mathcal{L}_{\text{C}}^{\text{comp}}$, the weight term in $\mathcal{L}_{\text{AC}}^{\text{blur}}$ (i.e., $e^{- \mu_{\bm{k}} H(G_{\bm{k}}(\bm{z}_{\bm{k}}))}$) and that in $\mathcal{L}_{\text{AC}}^{\text{comp}}$ (i.e., $e^{- \mu_{q} \frac{100 - G_{q}(\bm{z}_{q})}{100}}$) were replaced with a constant of one.

\smallskip\noindent\textbf{Results.}
The results are presented in Table~\ref{table:eval_ablation}.
The main findings were two-fold:
\textit{(i) Detailed ablation study on $\mathcal{L}_{\text{AC}}$ (Nos. 1--4).}
We found that BNCR-GAN performed reasonably well across all settings and achieved the best or second-best scores.
We also found that in the blur setting (D), BNCR-GAN w/o $\mathcal{L}_{\text{AC}}^{\text{blur}}$ achieved the best performance.
We consider this result because the suppression of blur (i.e., $\mathcal{L}_{\text{AC}}^{\text{blur}}$) was redundant in this case and that the suppression of compression (i.e., $\mathcal{L}_{\text{AC}}^{\text{comp}}$) was necessary and sufficient.
\textit{(ii) Effect of adaptiveness (Nos. 1, 5).}
We confirmed that BNCR-GAN (w/ $\mathcal{L}_{\text{AC}}$) outperformed BNCR-GAN w/ $\mathcal{L}_{\text{C}}$ in most cases (5/6).
One exception is setting (R), in which BNCR-GAN and BNCR-GAN w/ $\mathcal{L}_{\text{C}}$ showed the same performance.
In this particular case, we consider this to be due to $\mathcal{L}_{\text{AC}}$ becoming close to $\mathcal{L}_{\text{C}}$.
More precisely, in the noise-only case, BNCR-GAN attempted to learn \textit{no blur} or \textit{compression}.
When \textit{no blur} and \textit{no compression} was learned, the weight terms $e^{- \mu_{\bm{k}} H(G_{\bm{k}}(\bm{z}_{\bm{k}}))}$ and $e^{- \mu_{q} \frac{100 - G_{q}(\bm{z}_{q})}{100}}$ became one.
Consequently, $\mathcal{L}_{\text{AC}}$ becomes equal to $\mathcal{L}_{\text{C}}$.

\begin{table}[tb]
  \vspace{2mm}
  \centering
  \scriptsize{
    \begin{tabularx}{\columnwidth}{ccCCCCC}
      \toprule
      \multirow{2}{*}{\textsc{No.}} & \multirow{2}{*}{\textsc{Model}}
      & (S) & (T) & (H) & (I) & (A)
      \\ \cmidrule(r){3-6} \cmidrule(l){7-7}
      & & 20--40 & 40--60 & 60--80 & 80--100 & \textsc{Clean}
      \\ \midrule
      1 & GAN
      & 68.1 & 53.7 & 41.7 & 27.8 & \best{18.5}
      \\
      2 & AmbientGAN$^\dag$
      & 40.9 & 35.2 & 43.9 & 36.6 &  --
      \\
      3 & P-AmbientGAN
      & 41.7 & 39.0 & 42.8 & 36.5 & 35.7
      \\ \midrule
      4 & CR-GAN
      & \second{40.5} & \best{34.3} & \second{34.0} & \second{25.4} & 18.8 
      \\
      5 & CR-GAN w/o mask
      & 41.4 & 35.1 & 41.0 & 35.9 & 34.6 
      \\
      6 & BNCR-GAN
      & \best{38.6} & \second{34.7} & \best{29.9} & \best{24.6} & \second{18.6} 
      \\
      7 & BNCR-GAN w/o $\mathcal{L}_{\text{AC}}$
      & 41.5 & 36.8 & 36.1 & 30.6 & 20.5 
      \\ \midrule
      8 & Deblock+GAN
      & 59.9 & 49.2 & 39.9 & 28.1 &  -- 
      \\ \bottomrule
    \end{tabularx}
  }
  \caption{\textbf{Comparison of FID$\downarrow$ on \textsc{CIFAR-10} under severe compression settings.}
    Smaller values are preferable.
    The score calculation method and notation are the same as those in Table~\ref{table:eval_cifar10_blur_compression}.}
  \label{table:eval_severe_compression}
  \vspace{-5mm}
\end{table}

\begin{table*}[tb]
  \centering
  \scriptsize{
    \begin{tabularx}{\textwidth}{ccCCCCCCCCC}
      \toprule
      \multirow{2}{*}{\textsc{No.}} & \multirow{2}{*}{\textsc{Model}}
      & (S) & (T) & (H) & (I) & (J) & (K) & (L) & (M) & (A)
      \\ \cmidrule(r){3-7} \cmidrule(rl){8-10} \cmidrule(l){11-11}
      & & 20--40 & 40--60 & 60--80 & 80--100 & 60--100 & $\frac{1}{4}$(J) & $\frac{1}{2}$(J) & $\frac{3}{4}$(J)  & \textsc{Clean}
      \\ \midrule
      1 & CR-GAN
      & 40.5 & \second{34.3} & 34.0 & 25.4 & \best{26.3} & \best{20.1} & 22.7 & 24.5 & 18.8 
      \\
      2 & CR-GAN w/ $\mathcal{L}_{\text{AC}}^{\text{comp}}$ 
      & \second{39.6} & \best{33.4} & \best{29.0} & \best{24.5} & \second{26.6} & \second{20.2} & \second{22.5} & \best{24.4} & \best{18.5}
      \\ \midrule
      3 & BNCR-GAN
      & \best{38.6} & 34.7 & \second{29.9} & \second{24.6} & \second{26.6} & 20.3 & \best{22.4} & \best{24.4} & \second{18.6}
      \\
      4 & BNCR-GAN w/o $\mathcal{L}_{\text{AC}}$
      & 41.5 & 36.8 & 36.1 & 30.6 & 33.4 & 22.4 & 24.1 & 30.6 & 20.5 
      \\ \bottomrule
    \end{tabularx}
  }
  \caption{\textbf{Comparison of FID$\downarrow$ on \textsc{CIFAR-10} using CR-GANs with and without adaptive consistency loss.}
    Smaller values are preferred.
    The score calculation method and notation are the same as those in Table~\ref{table:eval_cifar10_blur_compression}.
    Refer to Tables~\ref{table:eval_cifar10_blur_compression} and \ref{table:eval_severe_compression} for the scores of the other models.}
  \label{table:eval_crgan_acloss}
  \vspace{-4mm}
\end{table*}

\subsubsection{Evaluation on compression robust image generation under severe compression settings}
\label{subsubsec:eval_severe_compression}

In the evaluation of compression robust image generation (Section~\ref{subsec:eval_cifar10_compression}), we set the quality factor to within the ranges of $[60, 80]$ (I), $[80, 100]$ (H), and $[60, 100]$ (J), based on the fact that typical software uses a near value as a default.
The remaining question is how the performance is affected under more severe compression settings.
To answer this question, we provide an additional analysis in this section.

\smallskip\noindent\textbf{Degradation settings.}
We examined the performance when training images were compressed with a relatively small quality factor, within the ranges of $[20, 40]$ (S) and $[40, 60]$ (T).
As references, we also provide the results under the compression settings (I and H), as well as under the clean image setting (A).

\smallskip\noindent\textbf{Compared models.}
We compared the same models as in Section~\ref{subsec:eval_cifar10_compression}.

\smallskip\noindent\textbf{Results.}
The results are summarized in Table~\ref{table:eval_severe_compression}.
Our major findings are as follows:
\textit{(i) Comparison with baseline GANs (Nos. 1--4, 6).}
Similar to the findings in Section~\ref{subsec:eval_cifar10_compression}, CR-GAN and BNCR-GAN achieved the best or second-best scores under severe compression settings (S and T).
AmbientGAN and CR-GAN w/o mask, which applied JPEG compression to all generated images, were continuously outperformed by these two models.
As discussed in Section~\ref{subsec:eval_cifar10_compression}, we consider that the lossy characteristics of JPEG caused this degradation.
P-AmbientGAN, which also applied JPEG compression to all generated images and learned a degradation parameter directly without using a generator, showed poor performance in all settings because it could not handle compression variety, which is included in all compression settings, as well as could not address the lossy characteristics of JPEG, similar to AmbientGAN and CR-GAN w/o mask.
The results of AmbientGAN, CR-GAN w/o mask, and P-AmbientGAN demonstrated a non-monotonic tendency according to the quality factor.
We suppose that a conflict may occur between the original compression artifacts and unexpected artifacts, which was yielded in the generated image and can disappear after compression.
A detailed analysis remains a potential direction for future studies.
\textit{(ii) Comparison with Deblock+GAN (Nos. 1, 4, 6, 8).}
We found that the superiority of Deblock+GAN over standard GAN tends to be achieved as the quality factor decreases.
In particular, Deblock+GAN outperformed standard GAN by a large margin (with a difference of over 4.5) under severe compression settings (S and T).
However, Deblock+GAN underperformed CR-GAN and BNCR-GAN by a large margin (with a difference of over 10) even under these settings.
\textit{(iii) Ablation study on CR-GAN (Nos. 4, 5).}
We confirmed that the masking architecture contributed to improved performance under all settings, including severe compression settings (S and T).
\textit{(iv) Ablation study on BNCR-GAN (Nos. 6, 7).}
Here, $\mathcal{L}_{\text{AC}}$ also brought about a positive effect under all settings, including severe compression settings (S and T).

\smallskip\noindent\textbf{Summary.}
Through these analyses, we confirmed that the statements in Section~\ref{subsec:eval_cifar10_compression} also hold under severe compression settings (S and T).

\subsubsection{Effectiveness of adaptive consistency loss in CR-GAN}
\label{subsubsec:eval_crgan_acloss}

From the results in Tables~\ref{table:eval_cifar10_blur_compression} and \ref{table:eval_severe_compression}, we found that in some cases, BNCR-GAN, which must learn \textit{no} noise and \textit{no} blur as well as quality-factor and image distributions, outperformed CR-GAN, which focuses on the learning of quality-factor and image distributions.
Particularly, in setting (H), BNCR-GAN outperformed CR-GAN by a large margin (with a difference of 4.1).
We consider that the adaptive consistency loss $\mathcal{L}_{\text{AC}}^{\text{comp}}$, which was used in BNCR-GAN but not in CR-GAN, resulted in this behavior.
To confirm this hypothesis, we evaluated the performance of CR-GAN with $\mathcal{L}_{\text{AC}}^{\text{comp}}$.

\smallskip\noindent\textbf{Degradation settings.}
We investigated the performance under all compression settings, including full compression settings (H--J, S--T) and partial compression settings (K--M).
As a reference, we also examined the performance under the clean image setting (A).

\smallskip\noindent\textbf{Compared models.}
To analyze the effect of the adaptive consistency loss, we compared \textit{CR-GAN} (w/o $\mathcal{L}_{\text{AC}}^{\text{comp}}$) with \textit{CR-GAN w/ $\mathcal{L}_{\text{AC}}^{\text{comp}}$}.
As a reference, we also report the scores for \textit{BNCR-GAN} and \textit{BNCR-GAN w/o $\mathcal{L}_{\text{AC}}$}.

\smallskip\noindent\textbf{Results.}
The results are summarized in Table~\ref{table:eval_crgan_acloss}.
As discussed above, CR-GAN (w/o $\mathcal{L}_{\text{AC}}^{\text{comp}}$) was inferior to BNCR-GAN in setting (H), whereas CR-GAN w/ $\mathcal{L}_{\text{AC}}^{\text{comp}}$ was comparable to BNCR-GAN.
These results indicate that the adaptive consistency loss is also useful for CR-GAN, and its absence is attributed to CR-GAN underperforming BNCR-GAN in some cases.
By comparing CR-GAN with CR-GAN w/ $\mathcal{L}_{\text{AC}}^{\text{comp}}$ in all settings, we found that the negative effect caused by $\mathcal{L}_{\text{AC}}^{\text{comp}}$ was relatively small.
In the main text, we omitted $\mathcal{L}_{\text{AC}}^{\text{comp}}$ in CR-GAN for simplicity; however, incorporating it is one option.

\subsection{Further analyses on FFHQ}
\label{subsec:analyses_ffhq}

In Section~\ref{subsec:eval_ffhq}, we selected three representative settings (D, J, and P) and compared the representative models to focus on the discussion.
As references, we provide other case results in this section.

\smallskip\noindent\textbf{Degradation settings.}
We examined the performance under diverse settings, including the blur setting (D), noise setting (R), compression setting (J), blur, noise, and compression settings (N and P), and clean image setting (A).

\smallskip\noindent\textbf{Compared models.}
In the blur setting (D), we compared the proposed \textit{BR-GAN} and \textit{BNCR-GAN} with three baseline GANs: standard \textit{GAN}, \textit{AmbientGAN}, and \textit{Deblur+GAN}.
In the noise setting (R), we compared the proposed \textit{BNCR-GAN} with four baseline GANs: standard \textit{GAN}, \textit{AmbientGAN}, \textit{NR-GAN}, and \textit{Denoise+GAN}.
In the compression setting (D), we compared the proposed \textit{CR-GAN} and \textit{BNCR-GAN} with three baseline GANs: standard \textit{GAN}, \textit{AmbientGAN}, and \textit{Deblock+GAN}.
In the blur, noise, and compression settings (N and P), we compared the proposed \textit{BNCR-GAN} with standard \textit{GAN} and \textit{AmbientGAN}.
Finally, in the clean image setting (A), we compared the proposed \textit{BR-GAN}, \textit{CR-GAN}, and \textit{BNCR-GAN} with standard \textit{GAN} and \textit{NR-GAN}.

\smallskip\noindent\textbf{Results.}
Table~\ref{table:eval_ffhq_ex} presents the results.
We discuss the results based on particular tasks.

\begin{table}[tb]
  \vspace{2mm}
  \centering
  \scriptsize{    
    \begin{tabularx}{\columnwidth}{cCcCcC}
      \addlinespace[-\aboverulesep]
      \cmidrule[\heavyrulewidth](r){1-2}
      \cmidrule[\heavyrulewidth](r){3-4}
      \cmidrule[\heavyrulewidth]{5-6}
      \multirow{2}{*}{\textsc{Model}} & (D)
      & \multirow{2}{*}{\textsc{Model}} & (R)
      & \multirow{2}{*}{\textsc{Model}} & (J)
      \\
      \cmidrule(rl){2-2} \cmidrule(rl){4-4} \cmidrule(l){6-6}
      & \textsc{Blur}
      & & \textsc{Noise}
      & & \textsc{Comp.}
      \\
      \cmidrule[\lightrulewidth](r){1-2}
      \cmidrule[\lightrulewidth](r){3-4}
      \cmidrule[\lightrulewidth]{5-6}
      GAN & 34.1
      & GAN & 33.4
      & GAN & 45.5
      \\
      AmbientGAN$^\dag$ & 27.7
      & AmbientGAN$^\dag$ & \best{18.4}
      & AmbientGAN$^\dag$ & 27.0
      \\
      \cmidrule[\lightrulewidth](r){1-2}
      \cmidrule[\lightrulewidth](r){3-4}
      \cmidrule[\lightrulewidth]{5-6}
      BR-GAN & \best{25.3}
      & NR-GAN & \second{19.8}
      & CR-GAN & \best{25.7}
      \\
      BNCR-GAN & \second{26.8}
      & BNCR-GAN & 20.0
      & BNCR-GAN & \second{26.2}
      \\
      \cmidrule[\lightrulewidth](r){1-2}
      \cmidrule[\lightrulewidth](r){3-4}
      \cmidrule[\lightrulewidth]{5-6}
      Deblur+GAN$^\dag$ & 32.9
      & Denoise+GAN$^\dag$ & 22.8
      & Deblock+GAN & 50.0
      \\
      \cmidrule[\heavyrulewidth](r){1-2}
      \cmidrule[\heavyrulewidth](r){3-4}
      \cmidrule[\heavyrulewidth]{5-6}      
    \end{tabularx}
    \begin{tabularx}{0.66\columnwidth}{cCC}
      \cmidrule[\heavyrulewidth](r){1-3}
      \multirow{2}{*}{\textsc{Model}} & (N) & (P)
      \\
      \cmidrule(rl){2-3}
      & \textsc{All}
      & $\frac{1}{2}$(N)
      \\
      \cmidrule[\lightrulewidth](r){1-3}
      GAN & 70.2 & 34.9
      \\
      AmbientGAN$^\dag$ & \second{57.7} & \second{24.5}
      \\
      \cmidrule[\lightrulewidth](r){1-3}
      BNCR-GAN & \best{51.6} & \best{24.2}
      \\
      \cmidrule[\heavyrulewidth](r){1-3}
    \end{tabularx}
    \\
    \begin{tabularx}{0.33\columnwidth}{cC}
      \addlinespace[\aboverulesep]
      \toprule
      \multirow{2}{*}{\textsc{Model}} & (A)
      \\
      \cmidrule{2-2}
      & \textsc{Clean}
      \\
      \cmidrule[\lightrulewidth]{1-2}
      GAN & \best{15.2}
      \\
      \cmidrule[\lightrulewidth]{1-2}
      BR-GAN & 16.3
      \\
      NR-GAN & 17.7
      \\
      CR-GAN & \second{15.4}
      \\
      BNCR-GAN & 18.9
      \\
      \bottomrule
    \end{tabularx}
  }
  \vspace{1mm}
  \caption{\textbf{Detailed comparison of FID$\downarrow$ on \textsc{FFHQ}.} This is an extended version of Table~\ref{table:eval_ffhq}.
    Smaller values are better.
    The score calculation method and notation are the same as those in Table~\ref{table:eval_cifar10_blur_compression}.}
  \label{table:eval_ffhq_ex}
  \vspace{-4mm}
\end{table}

\begin{table*}[tb]
  \centering
  \scriptsize{
    \begin{tabularx}{\textwidth}{cCCCC}
      \toprule
      \multirow{2}{*}{\textsc{Model}}
      & (O) & (P) & (Q) & (N)
      \\ \cmidrule{2-5} 
      & $\frac{1}{4}$(N) & $\frac{1}{2}$(N) & $\frac{3}{4}$(N) & \textsc{All}
      \\ \midrule
      UNIR$^\dag$
      & \second{28.4} / \second{0.0204} / \best{26.8} / \best{0.882}
      & \second{22.4} / \best{0.0155} / \best{27.0} / \best{0.888}
      & \second{21.0} / \best{0.0137} / \best{27.0} / \best{0.890}
      & \best{24.8} / \best{0.0146} / \best{26.6} / \second{0.883}
      \\ \midrule
      BNCR-GAN+UNIR
      & \best{27.7} / \best{0.0184} / 26.3 / \second{0.877}
      & \best{21.5} / \second{0.0172} / 25.7 / 0.870
      & \best{20.4} / \second{0.0152} / \second{26.5} / \second{0.888}
      & \best{24.8} / \second{0.0206} / \second{26.5} / \best{0.885}
      \\ \midrule
      CycleGAN$^\ddag$
      & 30.2 / 0.0226 / \second{26.7} / \second{0.877}
      & 27.1 / 0.0177 / \second{26.5} / \second{0.879}
      & 33.7 / 0.0259 / 24.0 / 0.835
      & N/A / N/A / N/A / N/A
      \\ \bottomrule
    \end{tabularx}
  }
  \caption{\textbf{Comparison of FID$\downarrow$, LPIPS$\downarrow$, PSNR$\uparrow$, and SSIM$\uparrow$ on \textsc{CIFAR-10} in image restoration.}
    This is an extended version of Table~\ref{table:eval_restoration}.
    The results are listed as FID$\downarrow$ / LPIPS$\downarrow$ / PSNR$\uparrow$ / SSIM$\uparrow$.
    The score calculation method and notation are the same as those in Table~\ref{table:eval_restoration}.}
  \label{table:eval_restoration_ex}
  \vspace{-4mm}
\end{table*}

\subsubsection{Evaluation on blur robust image generation (D)}
\label{subsubsec:eval_ffhq_blur}

Similar to the results on \textsc{CIFAR-10} (Table~\ref{table:eval_cifar10_blur_compression} (left)), we found that BR-GAN, which must estimate the blur parameters through training, achieved the best performance and outperformed AmbientGAN, which was trained under advantageous conditions in which the ground-truth blur information was provided.
We also found that, for this dataset, BNCR-GAN, which further should learn \textit{no noise} and \textit{no compression}, outperformed AmbientGAN and obtained the second-best score.
Deblur+GAN underperformed BR-GAN and BNCR-GAN, similar to the observation on CIFAR-10.
As a reference, we provide examples of the generated images in Figure~\ref{fig:examples_ffhq_blur}.

\subsubsection{Evaluation on noise robust image generation (R)}
\label{subsubsec:eval_ffhq_noise}

Similar to the tendency on \textsc{CIFAR-10} (Table~\ref{table:eval_cifar10_noise}), AmbientGAN, which was trained with the ground-truth noise information, achieved the best performance.
NR-GAN, which must estimate such information through training, showed a competitive performance (with a difference of 1.4), and the performance of BNCR-GAN, which further should learn \textit{no blur} and \textit{no compression}, was also comparable to that of AmbientGAN (with a difference of 1.6) and that of NR-GAN (with a difference of 0.2).
Denoise+GAN underperformed these three models.
This is also a tendency similar to that found in \textsc{CIFAR-10}.
For further clarification, we present examples of the generated images in Figure~\ref{fig:examples_ffhq_noise}.

\subsubsection{Evaluation on compression robust image generation (J)}
\label{subsubsec:eval_ffhq_compression}

Similar to the results on \textsc{CIFAR-10} (Table~\ref{table:eval_cifar10_blur_compression} (right)), CR-GAN and BNCR-GAN also achieved the best and second-best scores for this dataset.
Although AmbientGAN was trained under advantageous conditions, it was inferior to the two models.
Deblock+GAN achieved the worst score and underperformed standard GAN.
As discussed in Section~\ref{subsec:eval_ffhq}, we suppose that over-smoothing, which is a common drawback of model-based deblocking methods, degraded FID more than compression artifacts.
We present examples of the generated images in Figure~\ref{fig:examples_ffhq_compression}.

\subsubsection{Evaluation on blur, noise, and compression robust image generation (N and P)}
\label{subsubsec:eval_ffhq_all}

Similar to the findings on \textsc{CIFAR-10} (Table~\ref{table:eval_cifar10_all}), BNCR-GAN, which must estimate the blur, noise, and compression simultaneously in this case, outperformed standard GAN in both settings (N and P).
BNCR-GAN was comparable to AmbientGAN when the rate of degraded images was relatively low (P), and outperformed AmbientGAN by a large margin (with a difference of 6.1) when the rate of degraded images was relatively high (N).
We provide examples of the generated images in Figures~\ref{fig:examples_ffhq_all05} and \ref{fig:examples_ffhq_all}.

\subsubsection{Evaluation on standard image generation (A)}
\label{subsubsec:eval_ffhq_clean}

As expected, standard GAN achieved the best performance because $G_{\bm{k}}$, $G_{\bm{n}}$, or $G_{q}$ was redundant in this case.
The other models underperformed standard GAN, but the deterioration was relatively small for CR-GAN and BR-GAN (with differences of 0.2 and 1.1, respectively).
BNCR-GAN, which must learn \textit{no blur}, \textit{no noise}, and \textit{no compression} simultaneously in this case, achieved the worst score.
Although further improvement remains a possible direction for future studies, we would like to note that the score was still better than those of BNCR-GAN in the other cases, that is, settings (D, R, J, N, and P).
Examples of the generated images are shown in Figure~\ref{fig:examples_ffhq_clean}.

\smallskip\noindent\textbf{Summary.}
Through this detailed comparative study, we did not find any results that significantly contradict the results for \textsc{CIFAR-10}, as mentioned in Section~\ref{subsec:eval_ffhq}.

\subsection{Further analyses on image restoration}
\label{subsec:analyses_image_restoration}

In Section~\ref{subsec:restoration}, we used LPIPS~\cite{RZhangCVPR2018} instead of traditional metrics such as PSNR and SSIM~\cite{ZWangTIP2004} because a previous study~\cite{RZhangCVPR2018} demonstrated that when images are distorted by low-level image editing operations (including blur, noise, and compression, which were addressed in this study), LPIPS has a better correlation with human perceptual judgments than PSNR and SSIM.
As mentioned in \cite{RZhangCVPR2018}, this is because PSNR and SSIM are too simple and shallow functions to explain the nuances of human perception.
These characteristics are particularly problematic when images include various types of image degradation (e.g., blur, noise, and compression).
To validate this statement, we examined PSNR and SSIM and analyzed them along with FID and LPIPS.

\smallskip\noindent\textbf{Results.}
Table~\ref{table:eval_restoration_ex} lists FID, LPIPS, PSNR, and SSIM for the models used in Section~\ref{subsec:restoration}.
The performance improved when the FID and LPIPS values were small, and when the PSNR and SSIM values were large.
We found that the differences in PSNR and SSIM were relatively small in most cases (the scores were within $26.0 \pm 1.0$ and $0.880 \pm 0.010$, respectively), except for the scores of CycleGAN in setting (Q), where PSNR and SSIM are 24.0 and 0.835, respectively.
As a reference, we also investigated the scores of real test images that were degraded in setting (N).
The FID, LPIPS, PSNR, and SSIM were 43.3, 0.0383, 26.2, and 0.862, respectively.
These results imply that in most cases, PSNR is almost unchanged before and after restoration, although the restoration effect is obvious, as shown in Figure~\ref{fig:examples_restoration}.
By contrast, FID and LPIPS reflect these differences.
We consider that this is because PSNR is sensitive to the pixel shift, which often occurs when a motion blur is applied.
When it occurs, perfect restoration (e.g., pixel-wise position correction) is not trivial without prior knowledge.
However, PSNR cannot ignore the difference resulting from the pixel shift; consequently, it fails to capture the differences resulting from other factors, such as blur, noise, and compression.
SSIM can absorb this weakness to some extent; however, FID and LPIPS are more effective in explaining the differences, as presented in Table~\ref{table:eval_restoration_ex} and Figure~\ref{fig:examples_restoration}.
From these results, we conclude that FID and LPIPS are more appropriate than PSNR and SSIM in this experiment.

\clearpage
\section{Examples of generated and restored images}
\label{sec:examples}

\subsection{Examples of generated images on CIFAR-10}
\label{subsec:examples_cifar10}

\begin{figure*}[hbtp]
  \centering
  \includegraphics[width=0.7\textwidth]{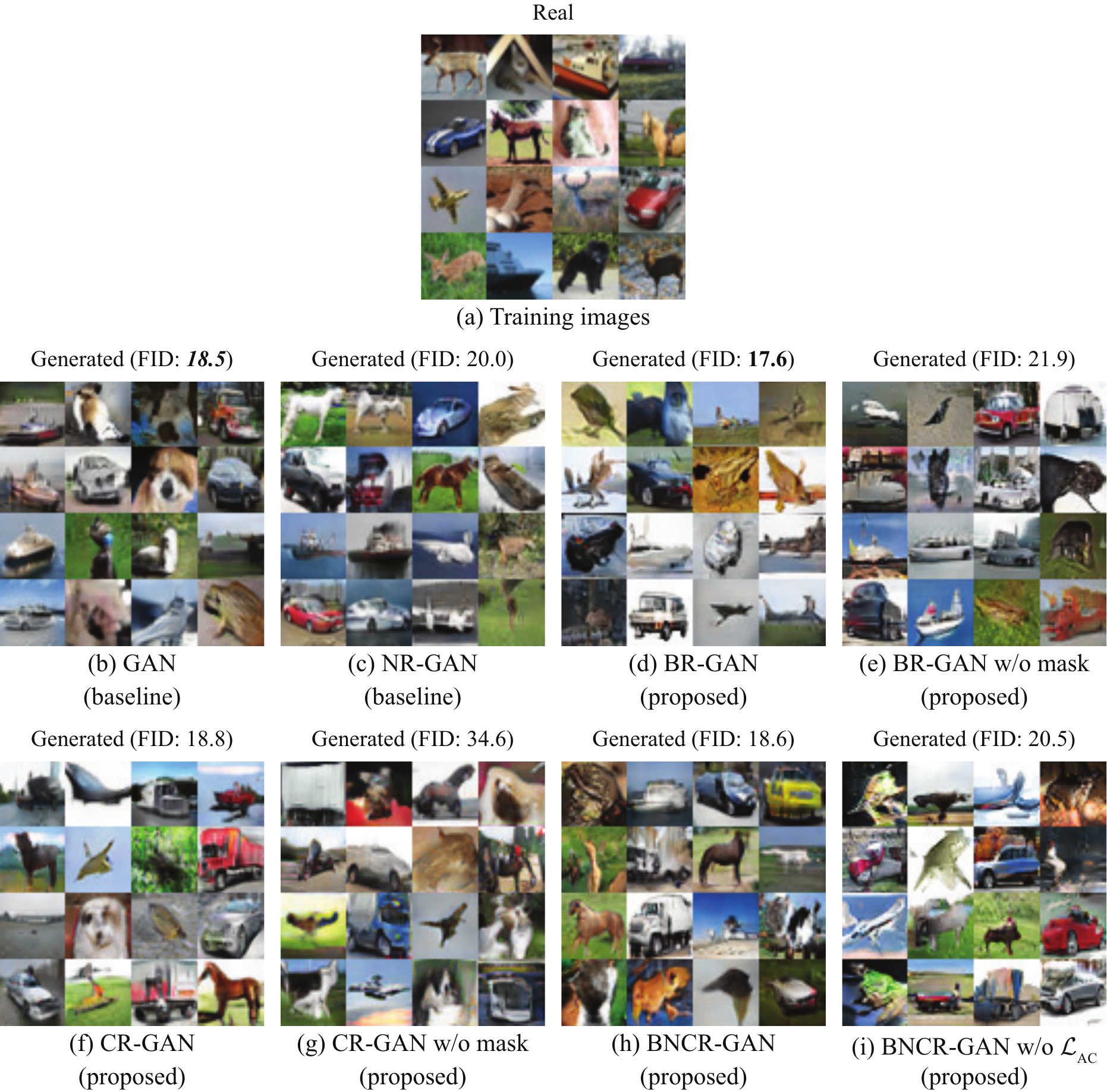}
  \caption{\textbf{Examples of standard image generation on \textsc{CIFAR-10} with clean image setting (A).}}
  \label{fig:examples_cifar10_clean}
\end{figure*}

\begin{figure*}[tb]
  \centering
  \includegraphics[width=0.7\textwidth]{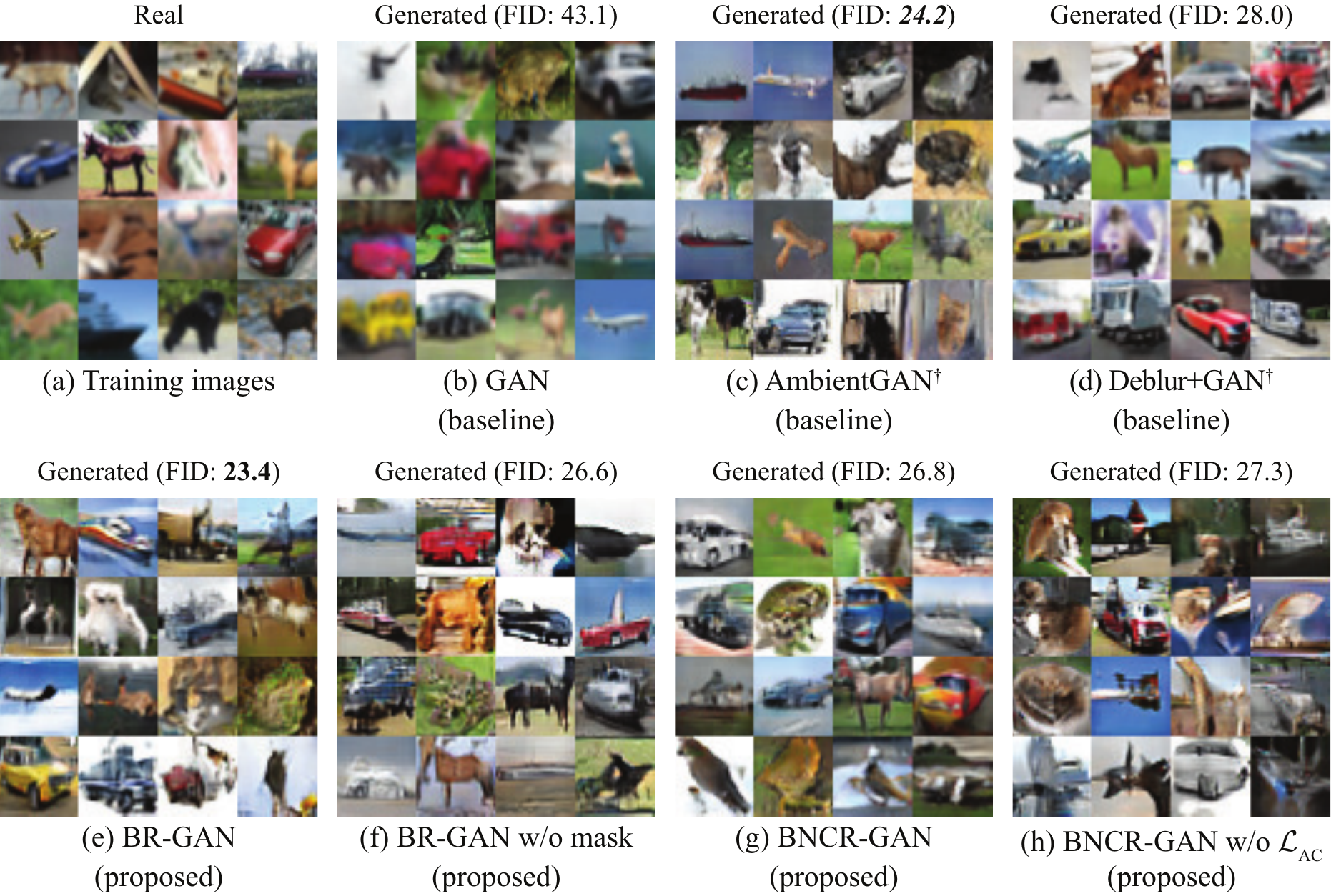}
  \caption{\textbf{Examples of blur robust image generation on \textsc{CIFAR-10} with blur setting (D).}
    AmbientGAN$^\dag$ and Deblur+GAN$^\dag$ were trained using ground-truth blur information.
    The other models were trained without using this information.}
  \label{fig:examples_cifar10_blur}
\end{figure*}

\begin{figure*}[tb]
  \centering
  \includegraphics[width=0.7\textwidth]{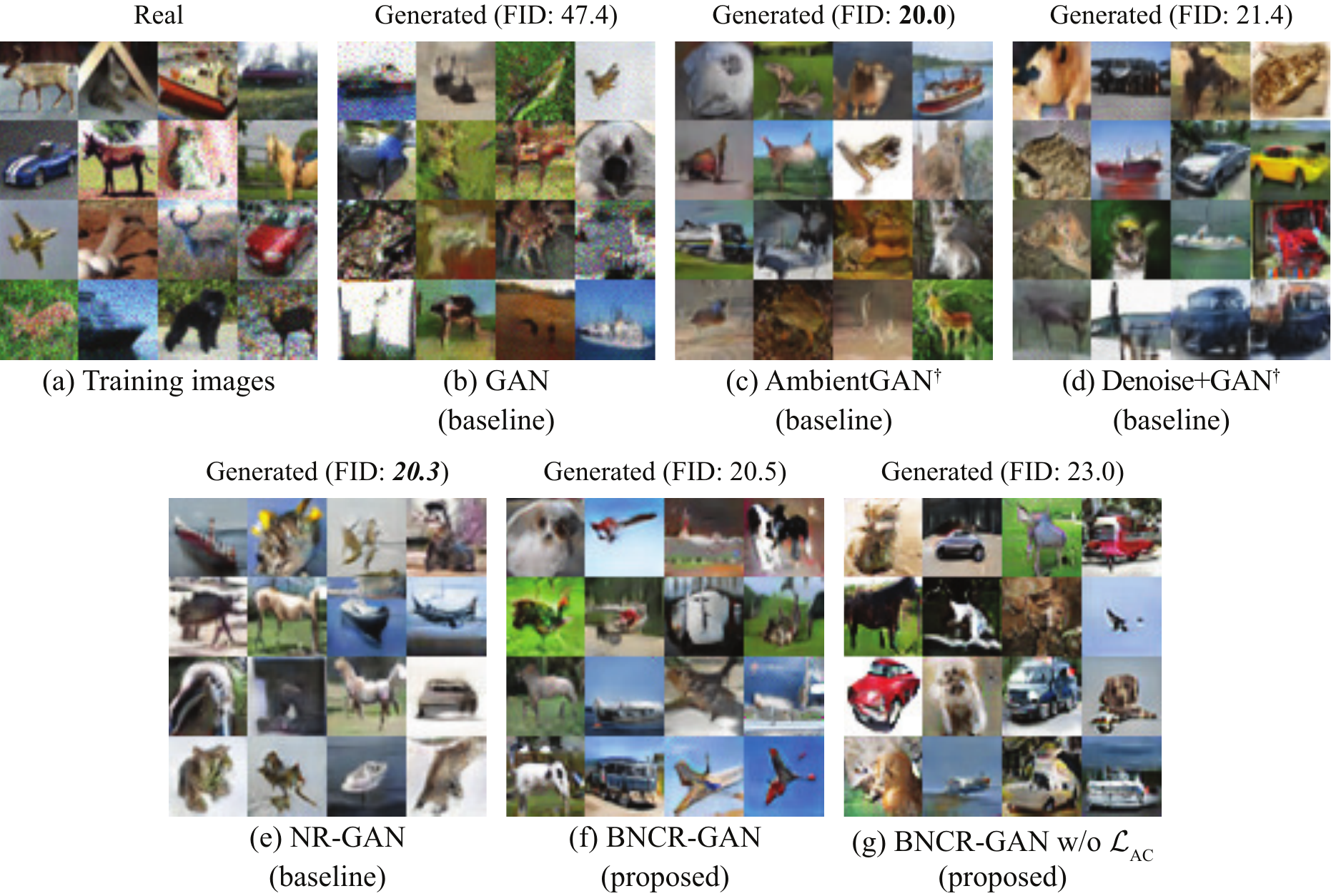}
  \caption{\textbf{Examples of noise robust image generation on \textsc{CIFAR-10} with noise setting (R).}
    AmbientGAN$^\dag$ and Denoise+GAN$^\dag$ were trained using ground-truth noise information.
    The other models were trained without using this information.}
  \label{fig:examples_cifar10_noise}
\end{figure*}

\begin{figure*}[tb]
  \centering
  \includegraphics[width=0.7\textwidth]{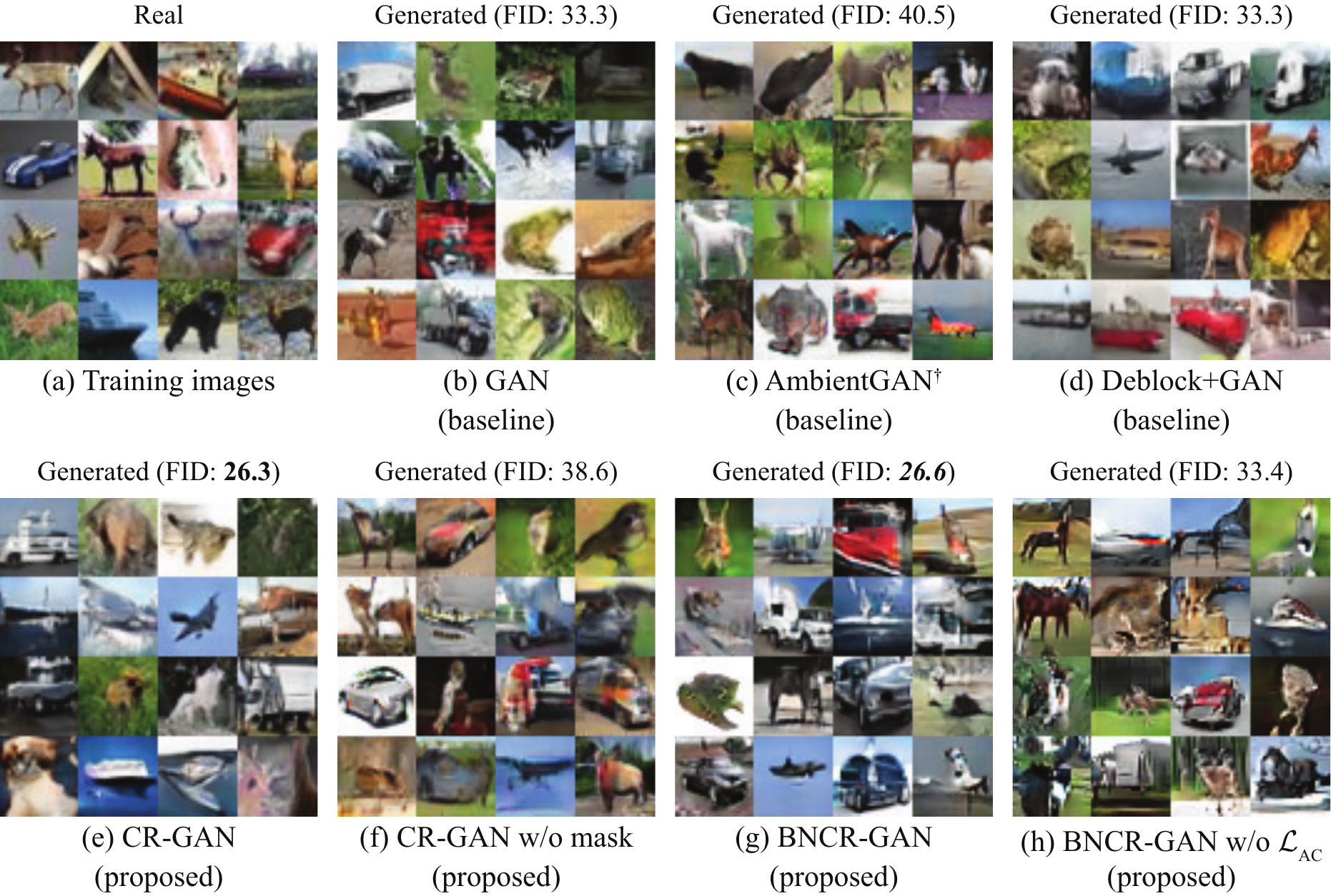}
  \caption{\textbf{Examples of compression robust image generation on \textsc{CIFAR-10} with compression setting (J).}
    AmbientGAN$^\dag$ was trained using ground-truth compression information.
    The other models were trained without using this information.}
  \label{fig:examples_cifar10_compression}
\end{figure*}

\begin{figure*}[tb]
  \centering  
  \includegraphics[width=0.7\textwidth]{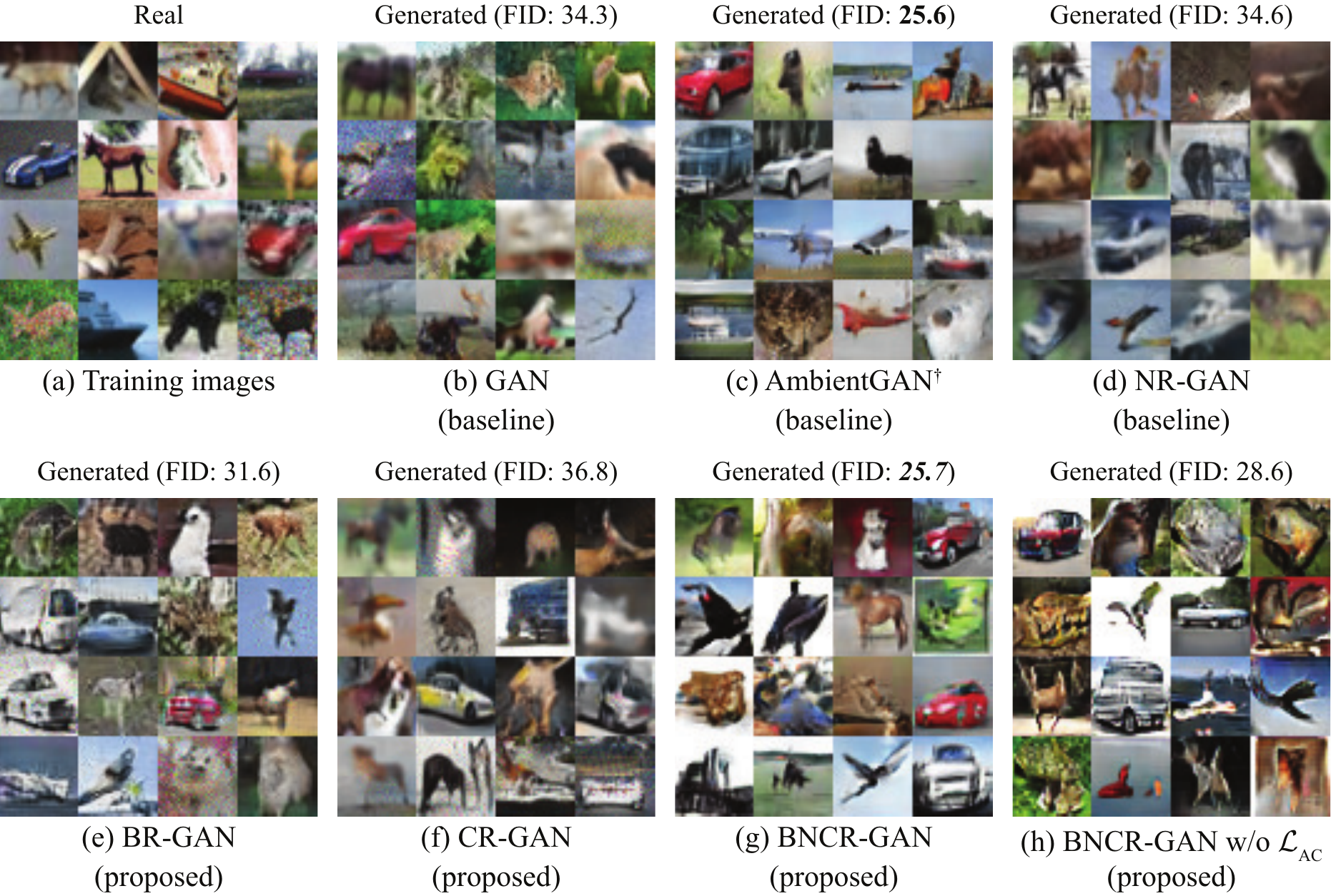}
  \caption{\textbf{Examples of blur, noise, and compression robust image generation on \textsc{CIFAR-10} with degradation setting (P).}
    AmbientGAN$^\dag$ was trained using ground-truth degradation information.
    The other models were trained without using this information.}
  \label{fig:examples_cifar10_all05}
\end{figure*}

\begin{figure*}[tb]
  \centering
  \includegraphics[width=0.7\textwidth]{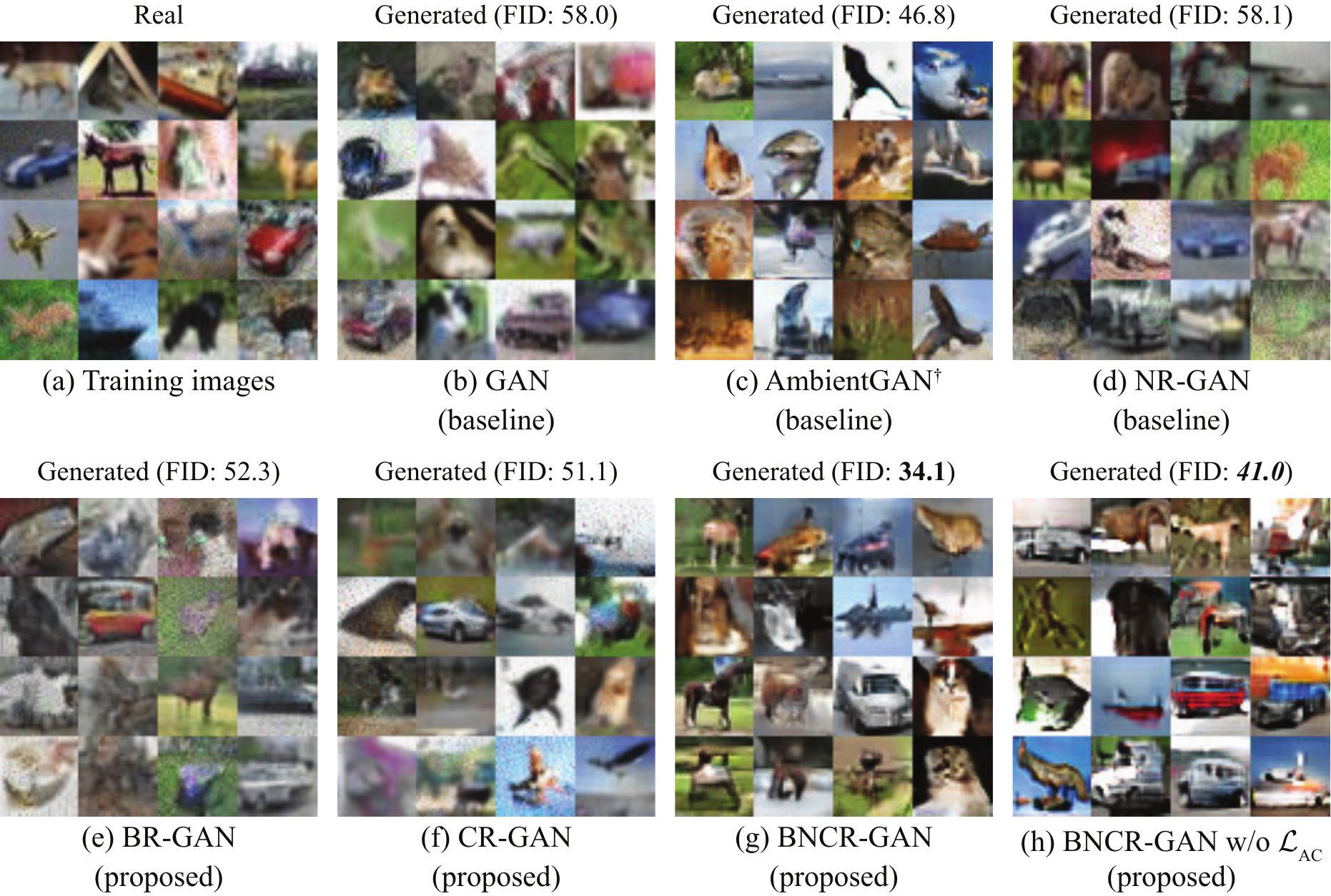}
  \caption{\textbf{Examples of blur, noise, and compression robust image generation on \textsc{CIFAR-10} with degradation setting (N).}
    AmbientGAN$^\dag$ was trained using ground-truth degradation information.
    The other models were trained without using this information.}
  \label{fig:examples_cifar10_all}
\end{figure*}

\begin{itemize}
\item Figure~\ref{fig:examples_cifar10_clean}:
  Examples of standard image generation on \textsc{CIFAR-10} with clean image setting (A).
\item Figure~\ref{fig:examples_cifar10_blur}:
  Examples of blur robust image generation on \textsc{CIFAR-10} with blur setting (D).
\item Figure~\ref{fig:examples_cifar10_noise}:
  Examples of noise robust image generation on \textsc{CIFAR-10} with noise setting (R).
\item Figure~\ref{fig:examples_cifar10_compression}:
  Examples of compression robust image generation on \textsc{CIFAR-10} with compression setting (J).
\item Figure~\ref{fig:examples_cifar10_all05}:
  Examples of blur, noise, and compression robust image generation on \textsc{CIFAR-10} with degradation setting (P).
\item Figure~\ref{fig:examples_cifar10_all}:
  Examples of blur, noise, and compression robust image generation on \textsc{CIFAR-10} with degradation setting (N).
\end{itemize}

\subsection{Examples of generated images on FFHQ}
\label{subsec:examples_ffhq}

\begin{figure*}[htbp]
  \centering  
  \includegraphics[height=0.8\textheight]{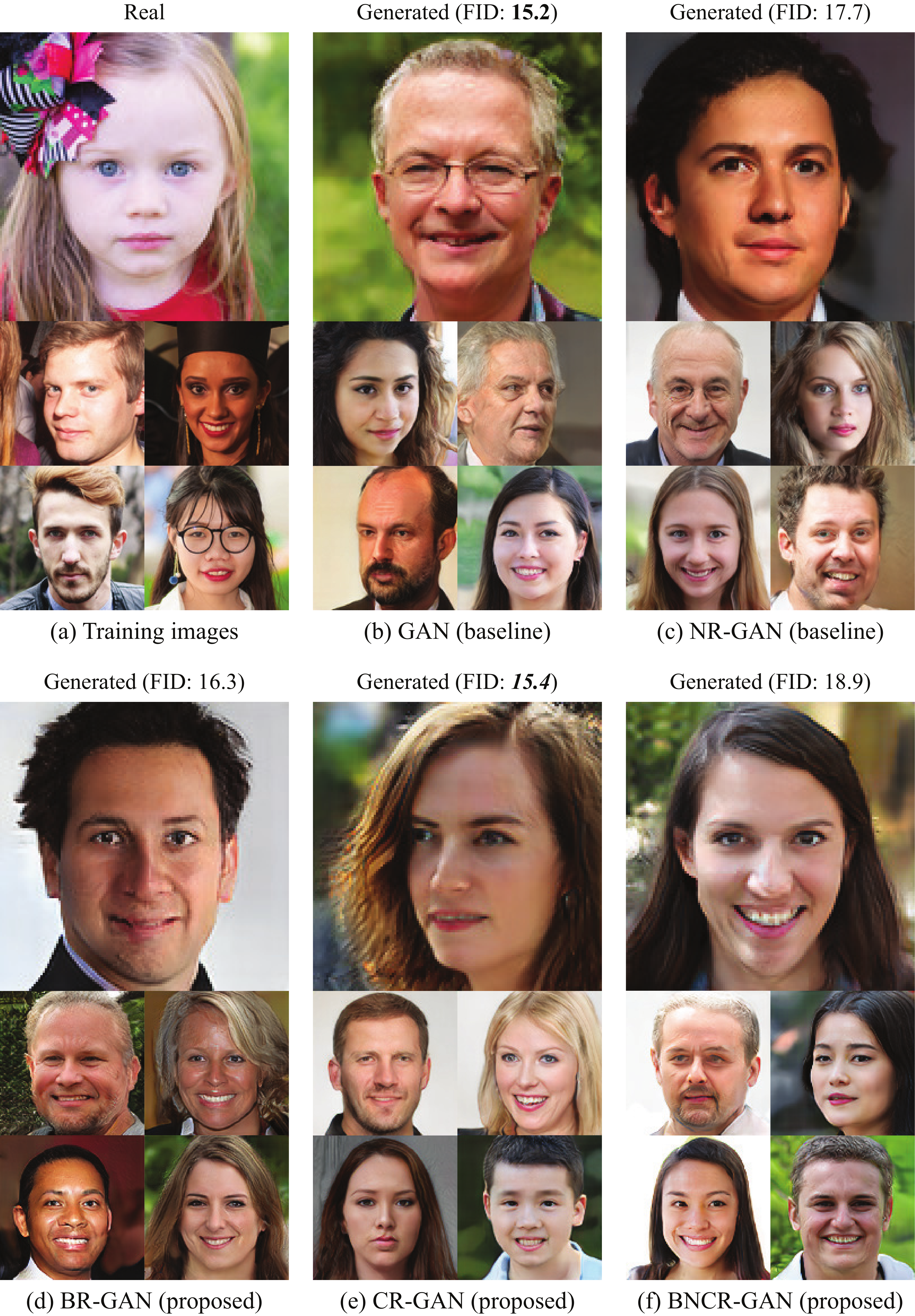}
  \caption{\textbf{Examples of standard image generation on \textsc{FFHQ} with clean image setting (A).}}
  \label{fig:examples_ffhq_clean}
\end{figure*}

\begin{figure*}[htbp]
  \centering  
  \includegraphics[height=0.8\textheight]{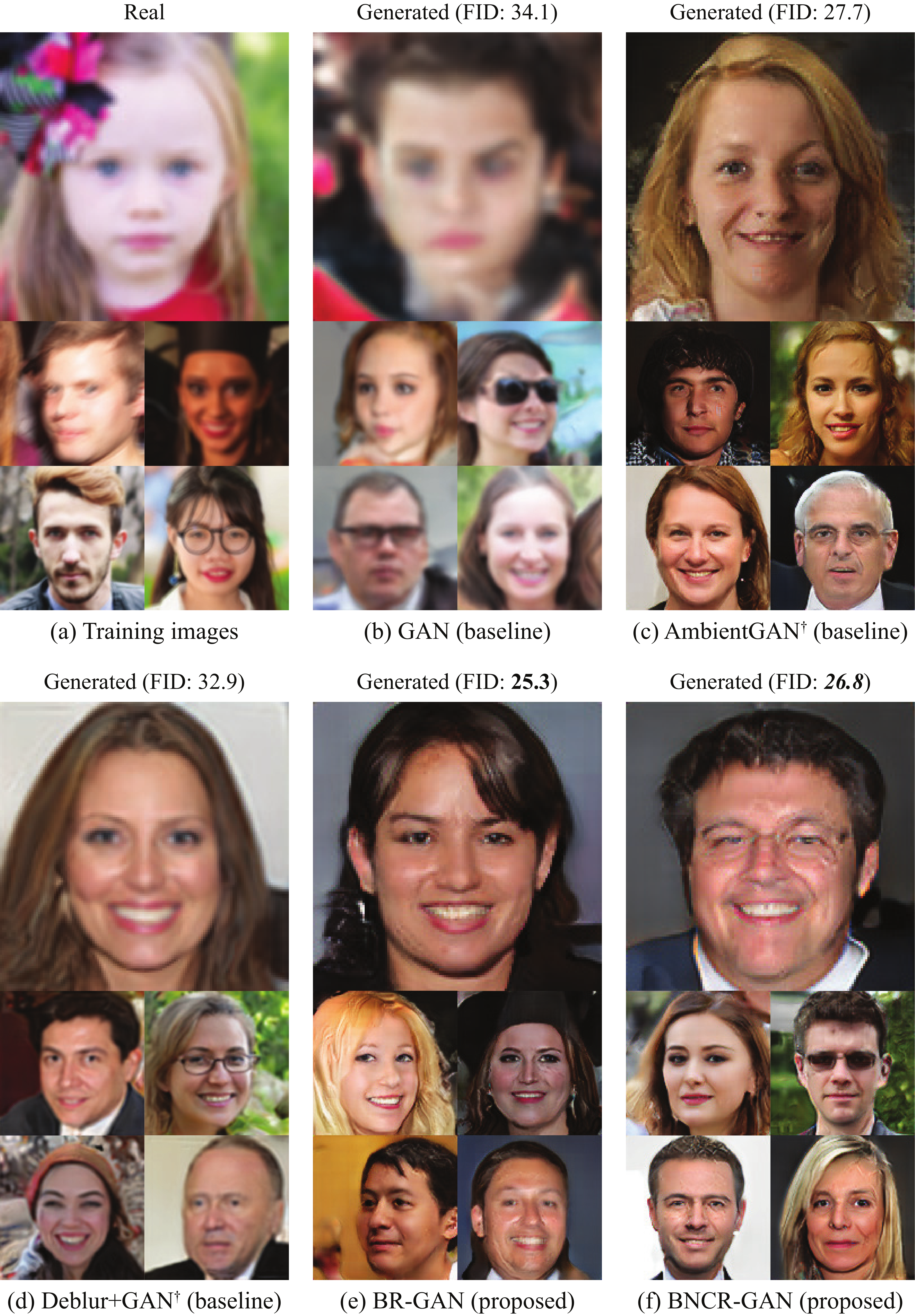}
  \caption{\textbf{Examples of blur robust image generation on \textsc{FFHQ} with blur setting (D).}
    This is an extended version of Figure~\ref{fig:examples_ffhq_blur_compression_select}.
    AmbientGAN$^\dag$ and Deblur+GAN$^\dag$ were trained using ground-truth blur information.
    The other models were trained without using this information.}
  \label{fig:examples_ffhq_blur}
\end{figure*}

\begin{figure*}[htbp]
  \centering  
  \includegraphics[height=0.8\textheight]{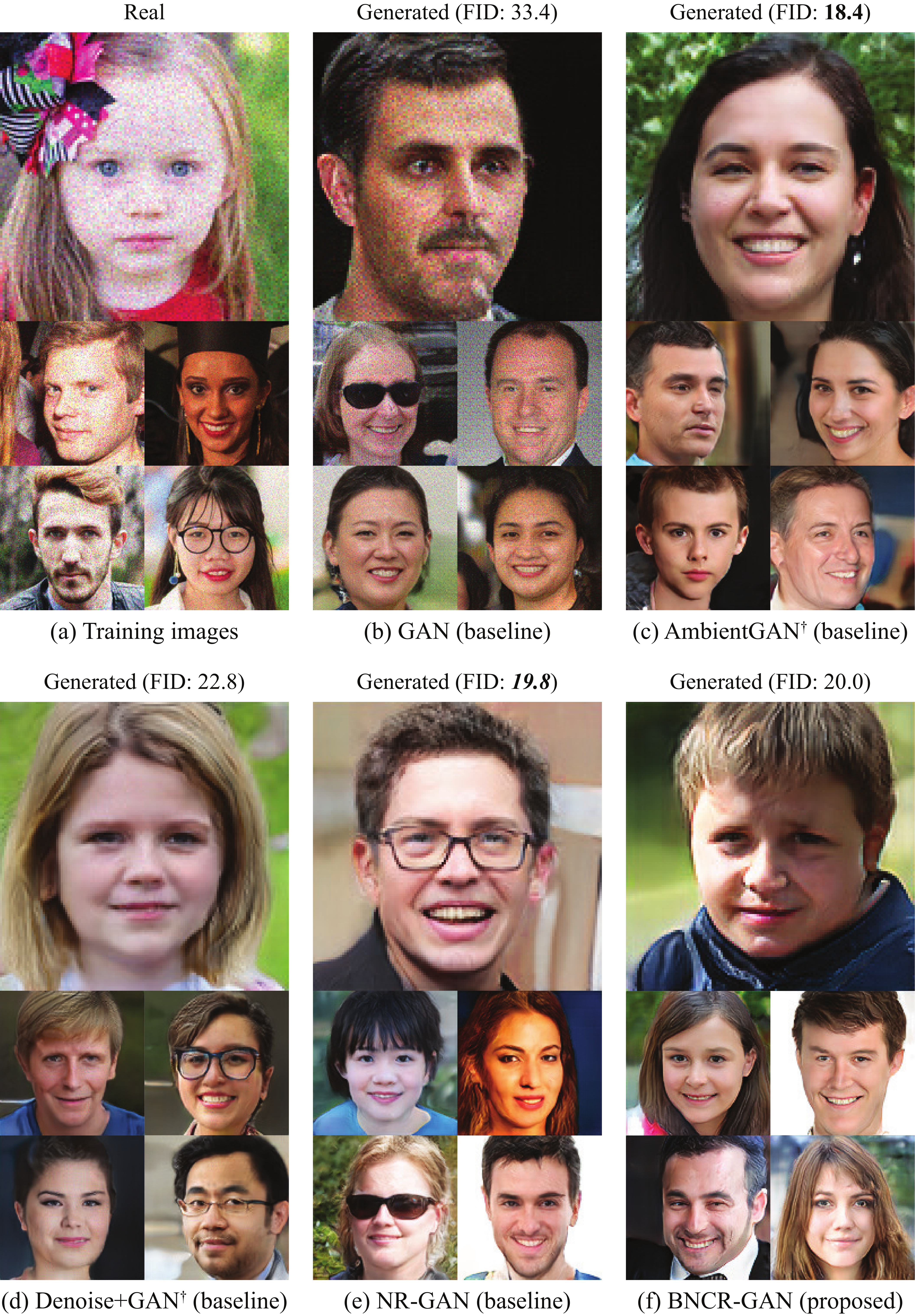}
  \caption{\textbf{Examples of noise robust image generation on \textsc{FFHQ} with noise setting (R).}
    AmbientGAN$^\dag$ and Denoise+GAN$^\dag$ were trained using ground-truth noise information.
    The other models were trained without using this information.}
  \label{fig:examples_ffhq_noise}
\end{figure*}

\begin{figure*}[htbp]
  \centering  
  \includegraphics[height=0.8\textheight]{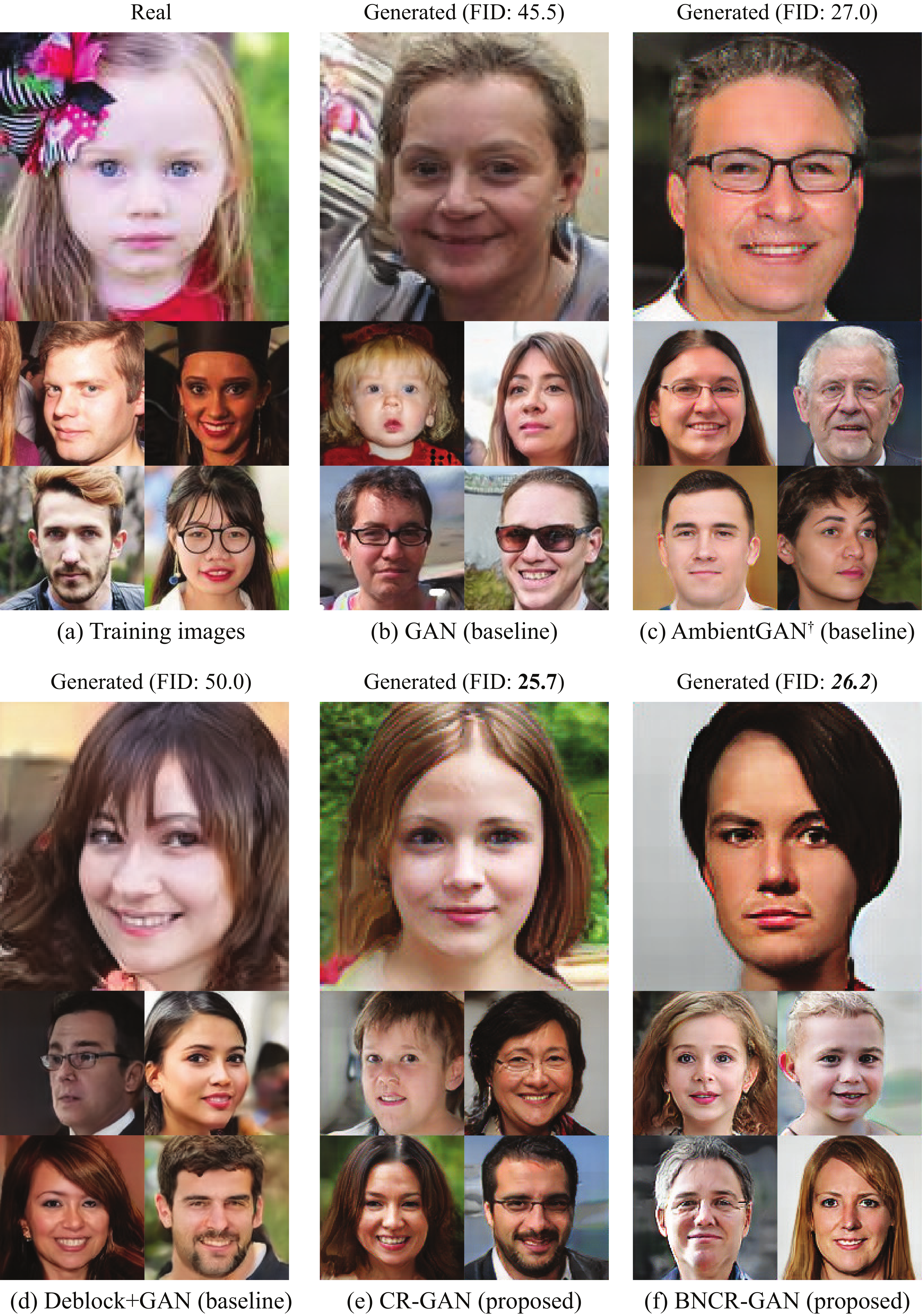}
  \caption{\textbf{Examples of compression robust image generation on \textsc{FFHQ} with compression setting (J).}
    This is an extended version of Figure~\ref{fig:examples_ffhq_blur_compression_select}.
    AmbientGAN$^\dag$ was trained using ground-truth compression information.
    The other models were trained without using this information.
    Compression artifacts tend to appear around the edges in (a) and (b).}
  \label{fig:examples_ffhq_compression}
\end{figure*}

\begin{figure*}[htbp]
  \centering
  \includegraphics[height=0.39\textheight]{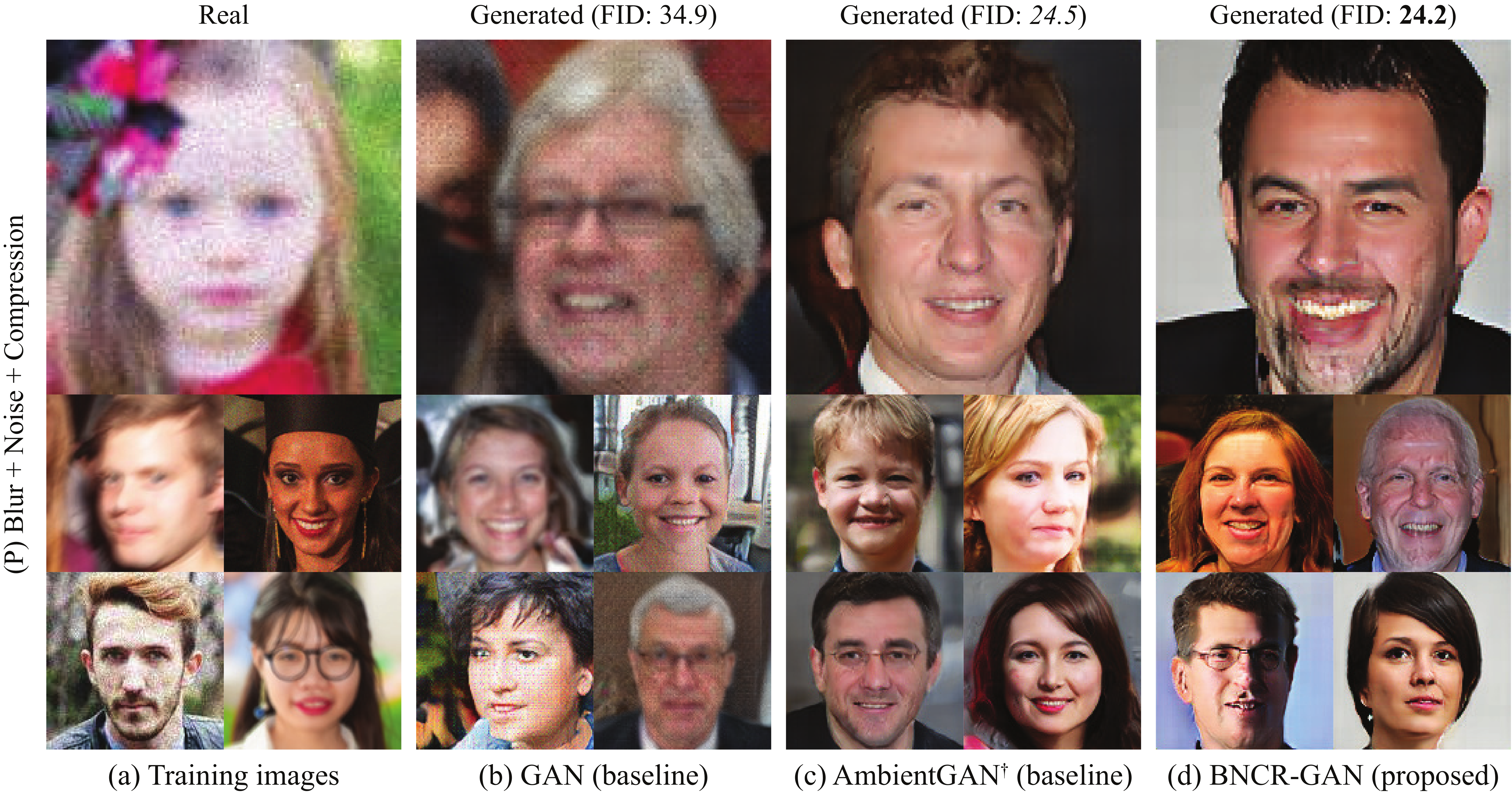}
  \caption{\textbf{Examples of blur, noise, and compression robust image generation on \textsc{FFHQ} with degradation setting (P).}
    This is an extended version of Figure~\ref{fig:examples_ffhq_all05_select}.
    AmbientGAN$^\dag$ was trained using ground-truth degradation information.
    The other models were trained without using this information.}
  \label{fig:examples_ffhq_all05}
\end{figure*}

\begin{figure*}[htbp]
  \centering
  \includegraphics[height=0.39\textheight]{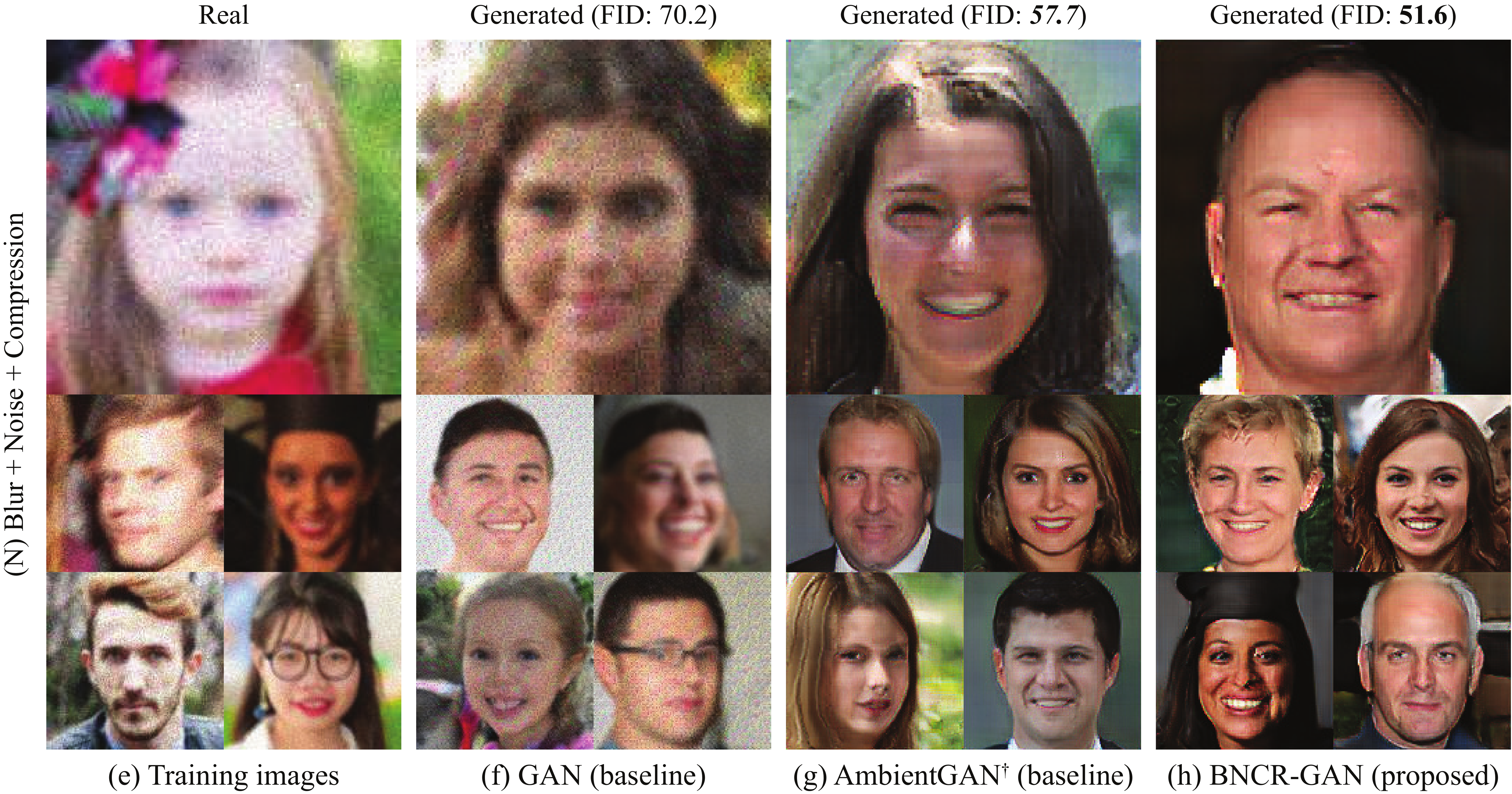}
  \caption{\textbf{Examples of blur, noise, and compression robust image generation on \textsc{FFHQ} with degradation setting (N).}
    AmbientGAN$^\dag$ was trained using ground-truth degradation information.
    The other models were trained without using this information.}
  \label{fig:examples_ffhq_all}
\end{figure*}

\begin{figure*}[htbp]
  \centering
  \includegraphics[width=1\textwidth]{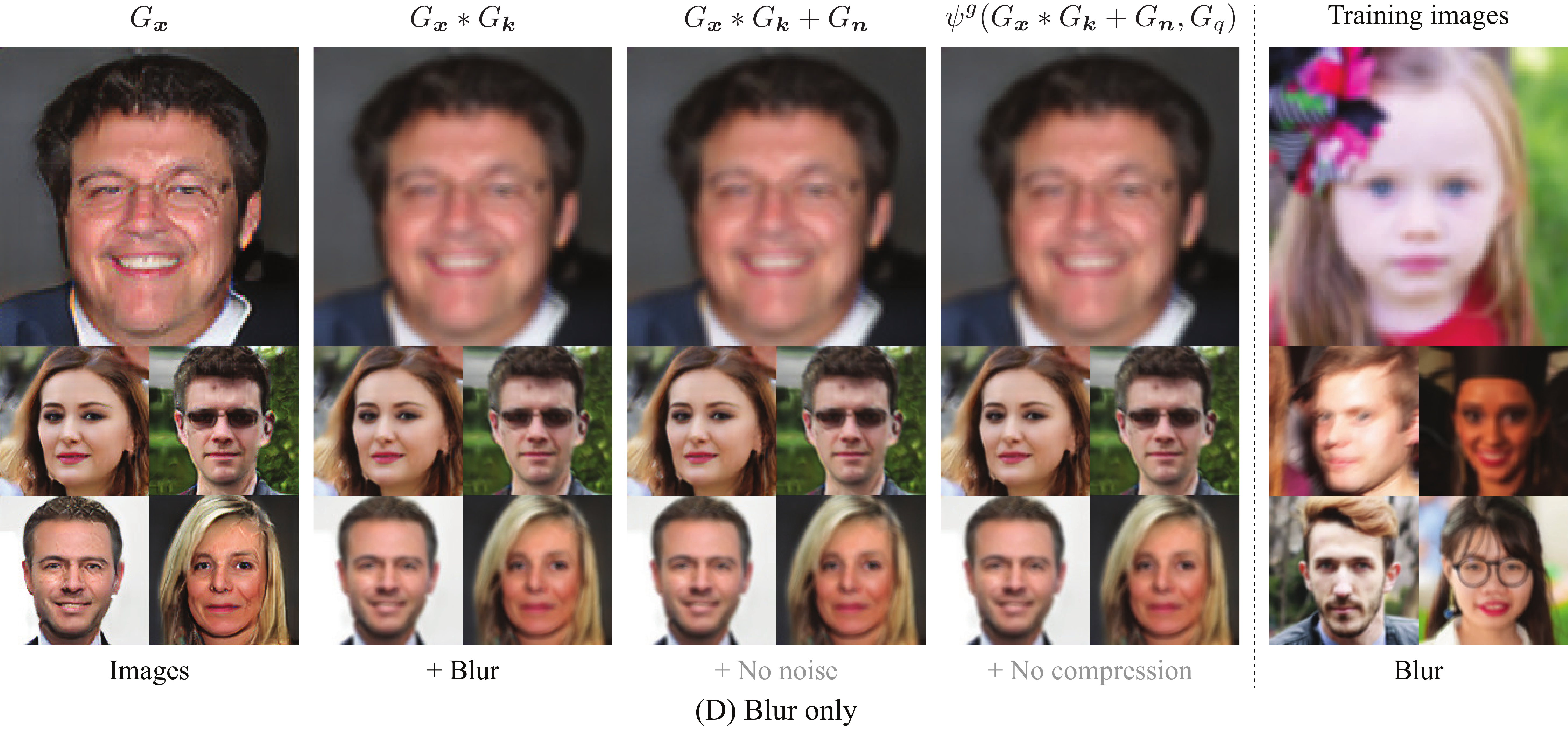}
  \caption{\textbf{Transition of generated images when incorporating BNCR-GAN generators in turn on \textsc{FFHQ} with blur setting (D).}
    On the left-hand side, we show the images generated from $G_{\bm{x}}$ only. From the neighbor to the right, we added $G_{\bm{k}}$, $G_{\bm{n}}$, and $G_{q}$ in turn.
    Here, we show the examples for setting (D), in which the training images include \textit{blur only}.
    Generators $G_{\bm{x}}$, $G_{\bm{k}}$, $G_{\bm{n}}$, and $G_{q}$ learn the image, blur kernel, noise, and quality factor, respectively, in a data-driven manner according to the degradation setting.
    Specifically, in this setting, $G_{\bm{k}}$ learns \textit{blur}, whereas $G_{\bm{n}}$ and $G_{q}$ learn \textit{no noise} and \textit{no compression}, respectively (with no large changes observed).}
  \label{fig:examples_ffhq_trans_blur}
\end{figure*}

\begin{figure*}[htbp]
  \centering
  \includegraphics[width=1\textwidth]{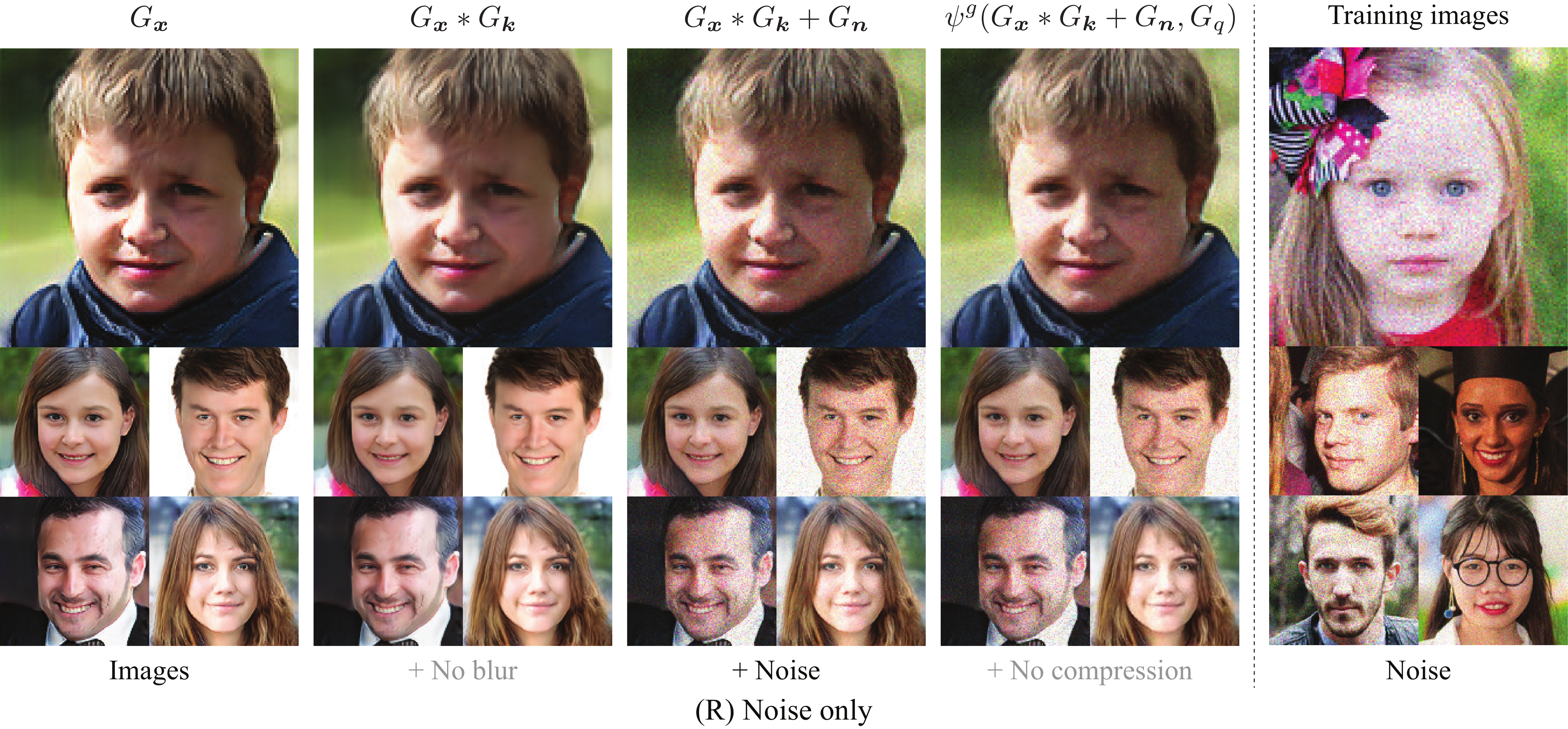}
  \caption{\textbf{Transition of generated images when incorporating BNCR-GAN generators in turn on \textsc{FFHQ} with noise setting (R).}
    On the left-most side, we show the images generated from $G_{\bm{x}}$ only. From the neighbor to the right, we added $G_{\bm{k}}$, $G_{\bm{n}}$, and $G_{q}$ in turn.
    Here, we show the examples for setting (R), in which the training images contain \textit{noise only}.
    Generators $G_{\bm{x}}$, $G_{\bm{k}}$, $G_{\bm{n}}$, and $G_{q}$ learn the image, blur kernel, noise, and quality factor, respectively, in a data-driven manner according to the degradation setting.
    Specifically, in this setting, $G_{\bm{n}}$ learns \textit{noise}, whereas $G_{\bm{k}}$ and $G_{q}$ learn \textit{no blur} and \textit{no compression}, respectively (with no large changes observed).}
  \label{fig:examples_ffhq_trans_noise}
\end{figure*}

\begin{figure*}[htbp]
  \centering  
  \includegraphics[width=1\textwidth]{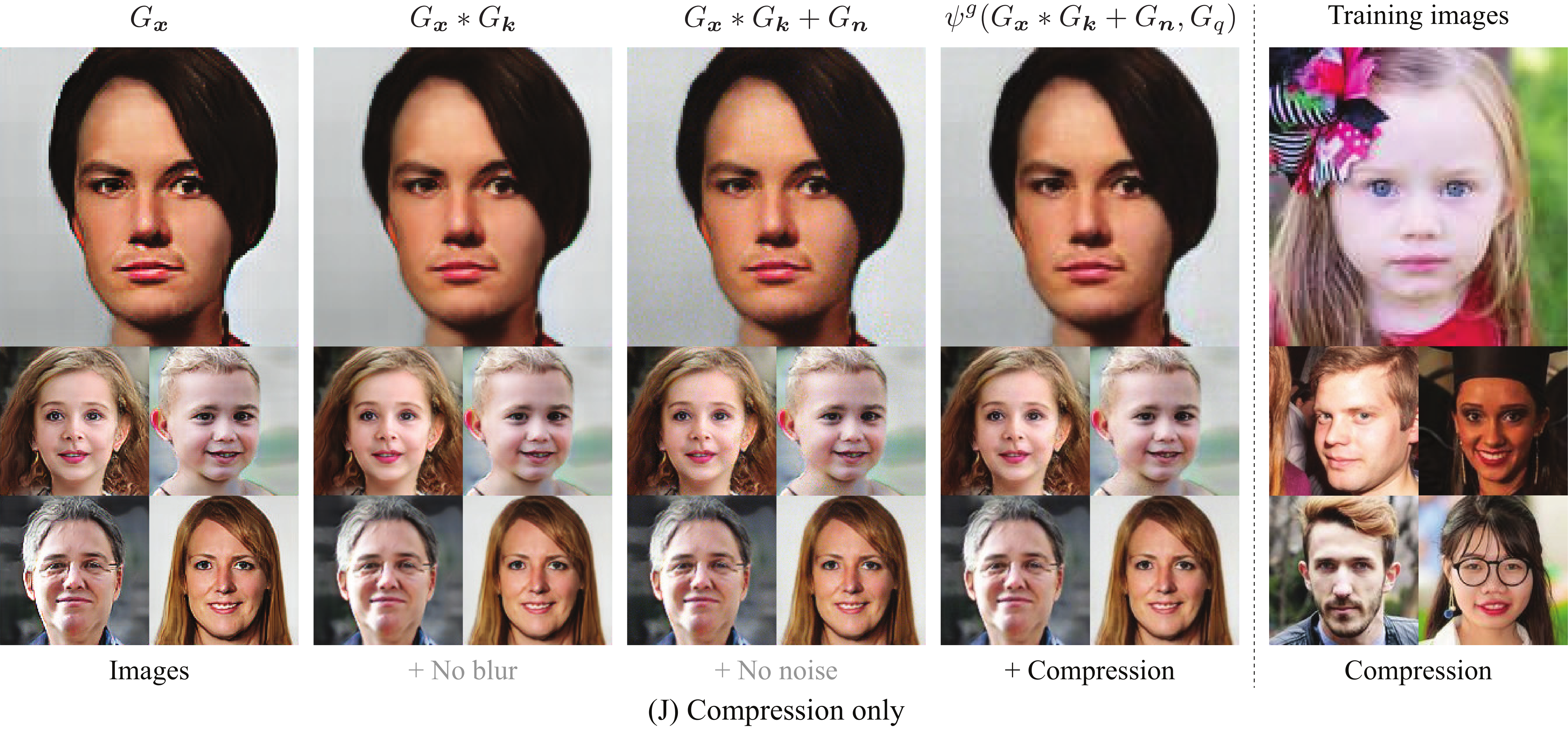}
  \caption{\textbf{Transition of generated images when incorporating BNCR-GAN generators in turn on \textsc{FFHQ} with compression setting (J).}
    On the extreme left, we show the images generated from $G_{\bm{x}}$ only.
    From the neighbor to the right, we added $G_{\bm{k}}$, $G_{\bm{n}}$, and $G_{q}$ in turn.
    Here, we show the examples for setting (J), in which the training images include \textit{compression only}.
    Generators $G_{\bm{x}}$, $G_{\bm{k}}$, $G_{\bm{n}}$, and $G_{q}$ learn the image, blur kernel, noise, and quality factor, respectively, in a data-driven manner according to the degradation setting.
    Specifically, in this setting, $G_{q}$ learns \textit{compression} (with compression artifacts observed around the edges), whereas $G_{\bm{k}}$ and $G_{\bm{n}}$ learn \textit{no blur} and \textit{no noise}, respectively (with no large changes observed).}
  \label{fig:examples_ffhq_trans_compression}
\end{figure*}

\begin{figure*}[htbp]
  \centering  
  \includegraphics[width=1\textwidth]{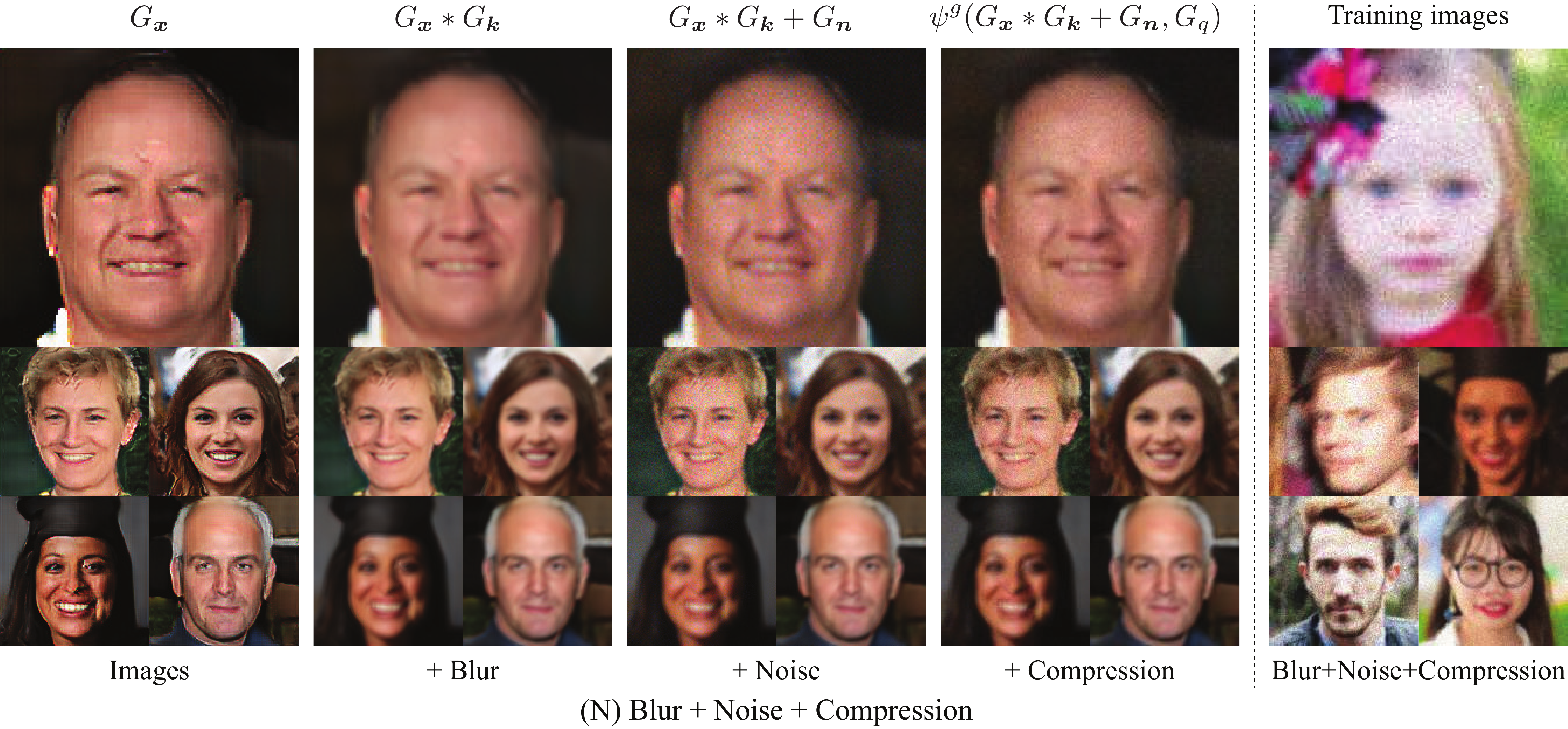}
  \caption{\textbf{Transition of generated images when incorporating BNCR-GAN generators in turn on \textsc{FFHQ} with degradation setting (N).}
    On the extreme left, we show the images generated from $G_{\bm{x}}$ only.
    From the neighbor to the right, we added $G_{\bm{k}}$, $G_{\bm{n}}$, and $G_{q}$ in turn.
    Here, we show the examples for setting (N), in which the training images contain \textit{blur, noise, and compression}.
    Generators $G_{\bm{x}}$, $G_{\bm{k}}$, $G_{\bm{n}}$, and $G_{q}$ learn the image, blur kernel, noise, and quality factor, respectively, in a data-driven manner according to the degradation setting.
    Specifically, in this setting, $G_{\bm{k}}$, $G_{\bm{n}}$, and $G_{q}$ learn \textit{blur}, \textit{noise}, and \textit{compression}, respectively.}
  \label{fig:examples_ffhq_trans_all}
\end{figure*}

\begin{figure*}[htbp]
  \centering  
  \includegraphics[width=1\textwidth]{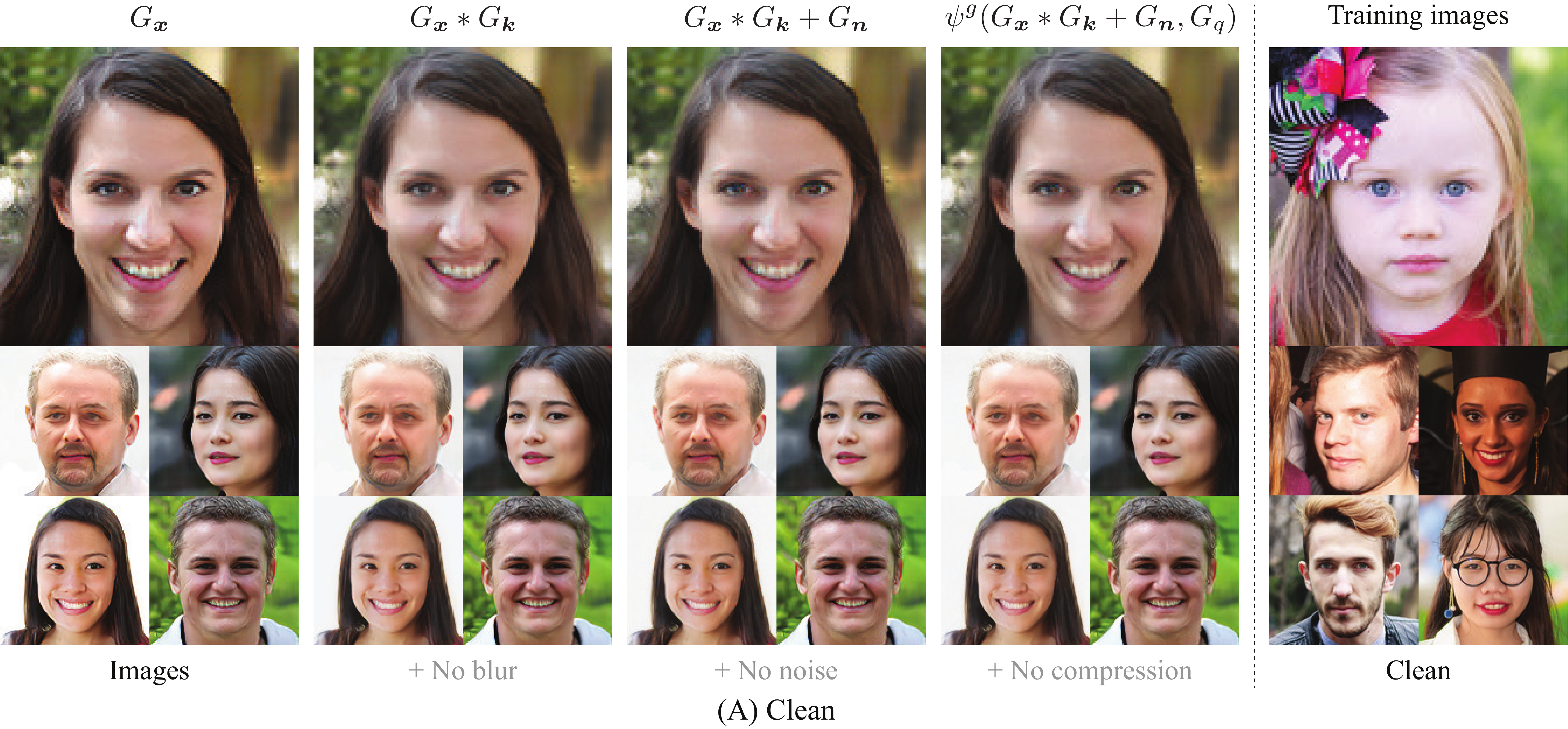}
  \caption{\textbf{Transition of generated images when incorporating BNCR-GAN generators in turn on \textsc{FFHQ} with clean image setting (A).}
    On the left-hand side, we show the images generated from $G_{\bm{x}}$ only.
    From the neighbor to the right, we added $G_{\bm{k}}$, $G_{\bm{n}}$, and $G_{q}$ in turn.
    Here, we show the examples for setting (A), in which the training images do \textit{not contain any blur, noise, or compression}.
    Generators $G_{\bm{x}}$, $G_{\bm{k}}$, $G_{\bm{n}}$, and $G_{q}$ learn the image, blur kernel, noise, and quality factor, respectively, in a data-driven manner according to the degradation setting.
    Specifically, in this setting, $G_{\bm{k}}$, $G_{\bm{n}}$, and $G_{q}$ learn \textit{no blur}, \textit{no noise}, and \textit{no compression}, respectively (with no large changes observed).}
  \label{fig:examples_ffhq_trans_clean}
\end{figure*}

\begin{figure*}[htbp]
  \centering
  \includegraphics[height=0.5\textheight]{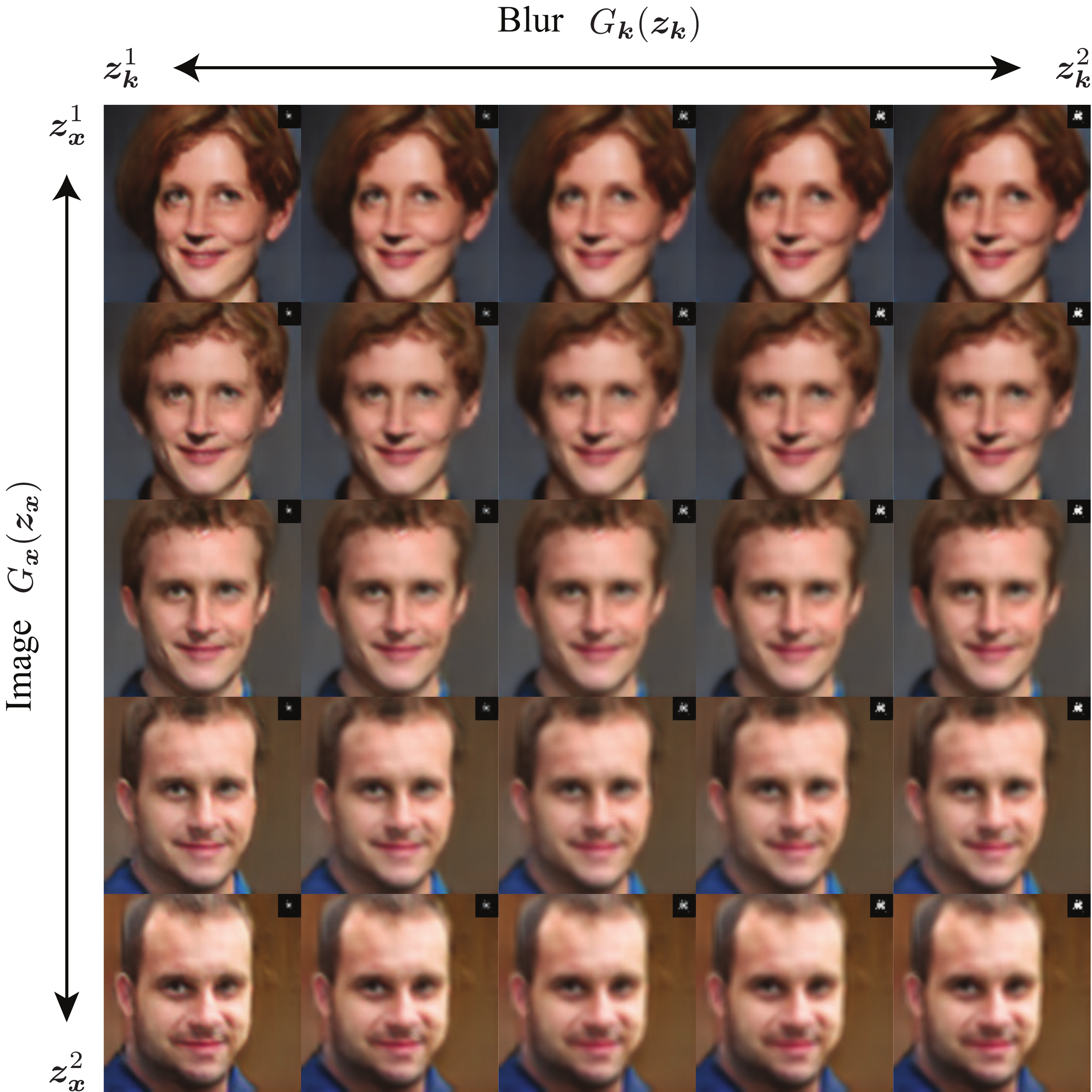}
  \caption{\textbf{Linear interpolation in latent spaces of the BNCR-GAN image and blur-kernel generators on \textsc{FFHQ} with degradation setting (N).}
    The corresponding generated kernel is visualized in the top-right corner of each image.
    Linear interpolation in the vertical and horizontal directions was conducted in the latent spaces of the image generator $G_{\bm{x}}$ and blur-kernel generator $G_{\bm{k}}$, respectively.
    As shown here, we can conduct linear interpolation independently between an image and blur kernel because we model them separately.
    This is the strength of individual modeling.
    Figures~\ref{fig:examples_ffhq_noise_interpolation} and \ref{fig:examples_ffhq_compression_interpolation} show the linear interpolation in the latent spaces of the noise generator $G_{\bm{n}}$ and quality-factor generator $G_q$, respectively.
    These generators were learned simultaneously with the generators used in this figure (i.e., $G_{\bm{x}}$ and $G_{\bm{k}}$).}
  \label{fig:examples_ffhq_blur_interpolation}
\end{figure*}

\begin{figure*}[htbp]
  \centering
  \includegraphics[height=0.5\textheight]{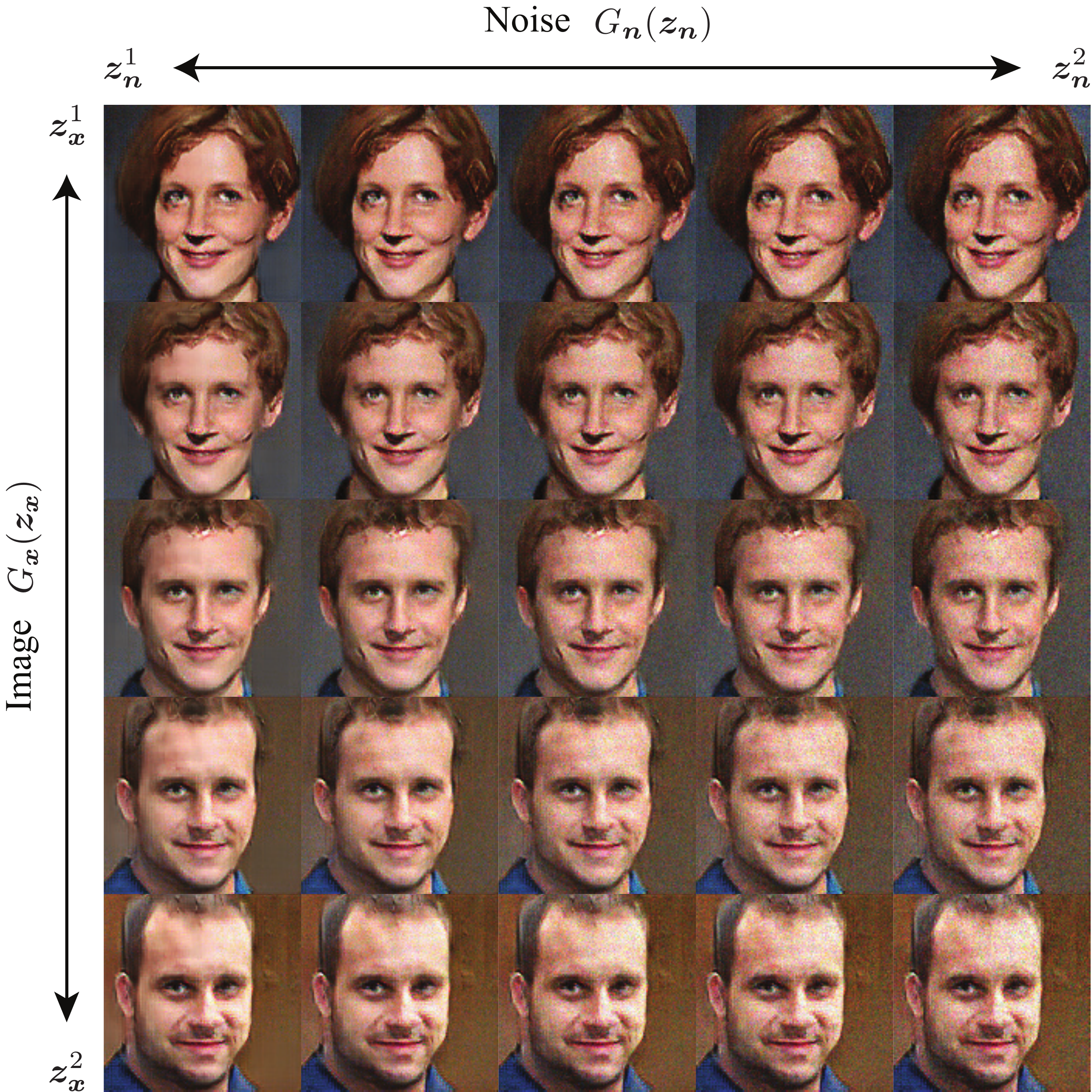}
  \caption{\textbf{Linear interpolation in latent spaces of the BNCR-GAN image and noise generators on \textsc{FFHQ} with degradation setting (N).}
    Linear interpolation in the vertical and horizontal directions was conducted in the latent spaces of the image generator $G_{\bm{x}}$ and noise generator $G_{\bm{n}}$, respectively.
    As shown here, we can conduct linear interpolation independently between an image and noise because we model them separately.
    Figures~\ref{fig:examples_ffhq_blur_interpolation} and \ref{fig:examples_ffhq_compression_interpolation} show the linear interpolation in the latent spaces of the blur-kernel generator $G_{\bm{k}}$ and quality-factor generator $G_q$, respectively.
    These generators were learned simultaneously with the generators used in this figure (i.e., $G_{\bm{x}}$ and $G_{\bm{n}}$).}
  \label{fig:examples_ffhq_noise_interpolation}
\end{figure*}

\begin{figure*}[htbp]
  \centering
  \includegraphics[height=0.5\textheight]{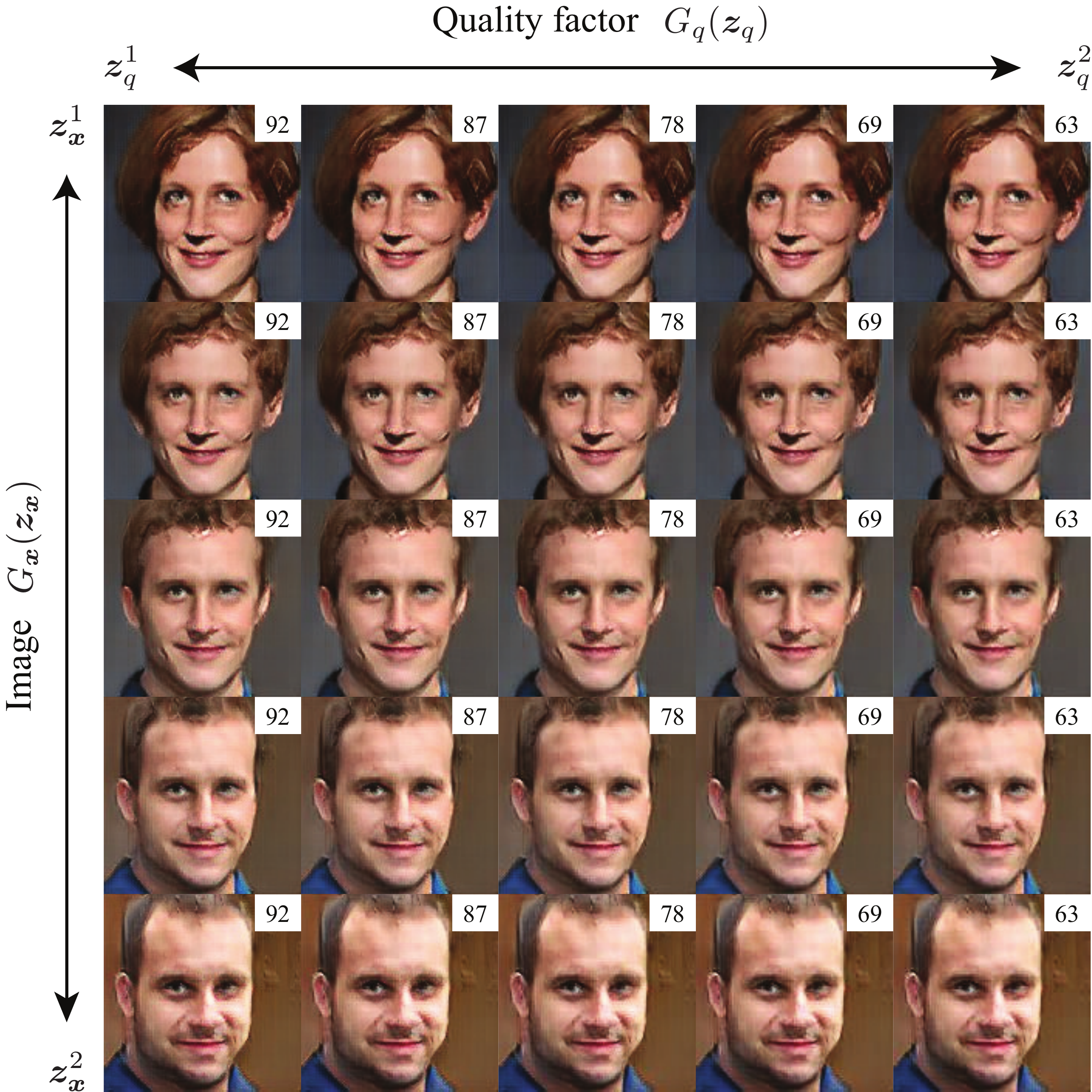}
  \caption{\textbf{Linear interpolation in latent spaces of the BNCR-GAN image and quality-factor generators on \textsc{FFHQ} with degradation setting (N).}
    The corresponding quality factor is shown in the top-right corner of each image.
    Linear interpolation in the vertical and horizontal directions was conducted in the latent spaces of the image generator $G_{\bm{x}}$ and quality-factor generator $G_q$, respectively.
    As shown here, we can conduct linear interpolation independently between an image and the quality factor because we model them separately.
    Figures~\ref{fig:examples_ffhq_blur_interpolation} and \ref{fig:examples_ffhq_noise_interpolation} show the linear interpolation in the latent spaces of the blur-kernel generator $G_{\bm{k}}$ and noise generator $G_{\bm{n}}$, respectively.
    These generators were learned simultaneously with the generators used in this figure (i.e., $G_{\bm{x}}$ and $G_q$).}
  \label{fig:examples_ffhq_compression_interpolation}
\end{figure*}

\begin{itemize}
\item Figure~\ref{fig:examples_ffhq_clean}:
  Examples of standard image generation on \textsc{FFHQ} with clean image setting (A).
\item Figure~\ref{fig:examples_ffhq_blur}:
  Examples of blur robust image generation on \textsc{FFHQ} with blur setting (D).
\item Figure~\ref{fig:examples_ffhq_noise}:
  Examples of noise robust image generation on \textsc{FFHQ} with noise setting (R).
\item Figure~\ref{fig:examples_ffhq_compression}:
  Examples of compression robust image generation on \textsc{FFHQ} with compression setting (J).
\item Figure~\ref{fig:examples_ffhq_all05}:
  Examples of blur, noise, and compression robust image generation on \textsc{FFHQ} with degradation setting (P).
\item Figure~\ref{fig:examples_ffhq_all}:
  Examples of blur, noise, and compression robust image generation on \textsc{FFHQ} with degradation setting (N).
\item Figure~\ref{fig:examples_ffhq_trans_blur}:
  Transition of generated images when incorporating BNCR-GAN generators in turn on \textsc{FFHQ} with blur setting (D).
\item Figure~\ref{fig:examples_ffhq_trans_noise}:
  Transition of generated images when incorporating BNCR-GAN generators in turn on \textsc{FFHQ} with noise setting (R).
\item Figure~\ref{fig:examples_ffhq_trans_compression}:
  Transition of generated images when incorporating BNCR-GAN generators in turn on \textsc{FFHQ} with compression setting (J).
\item Figure~\ref{fig:examples_ffhq_trans_all}:
  Transition of generated images when incorporating BNCR-GAN generators in turn on \textsc{FFHQ} with degradation setting (N).
\item Figure~\ref{fig:examples_ffhq_trans_clean}:
  Transition of generated images when incorporating BNCR-GAN generators in turn on \textsc{FFHQ} with clean image setting (A).
\item Figure~\ref{fig:examples_ffhq_blur_interpolation}:
  Linear interpolation in latent spaces of the BNCR-GAN image and blur-kernel generators on \textsc{FFHQ} with degradation setting (N).
\item Figure~\ref{fig:examples_ffhq_noise_interpolation}:
  Linear interpolation in latent spaces of the BNCR-GAN image and noise generators on \textsc{FFHQ} with degradation setting (N).
\item Figure~\ref{fig:examples_ffhq_compression_interpolation}:
  Linear interpolation in latent spaces of the BNCR-GAN image and quality-factor generators on \textsc{FFHQ} with degradation setting (N).
\end{itemize}

\subsection{Examples of restored images}
\label{subsec:examples_restoration}

\begin{figure*}[htbp]
  \centering
  \includegraphics[width=1\textwidth]{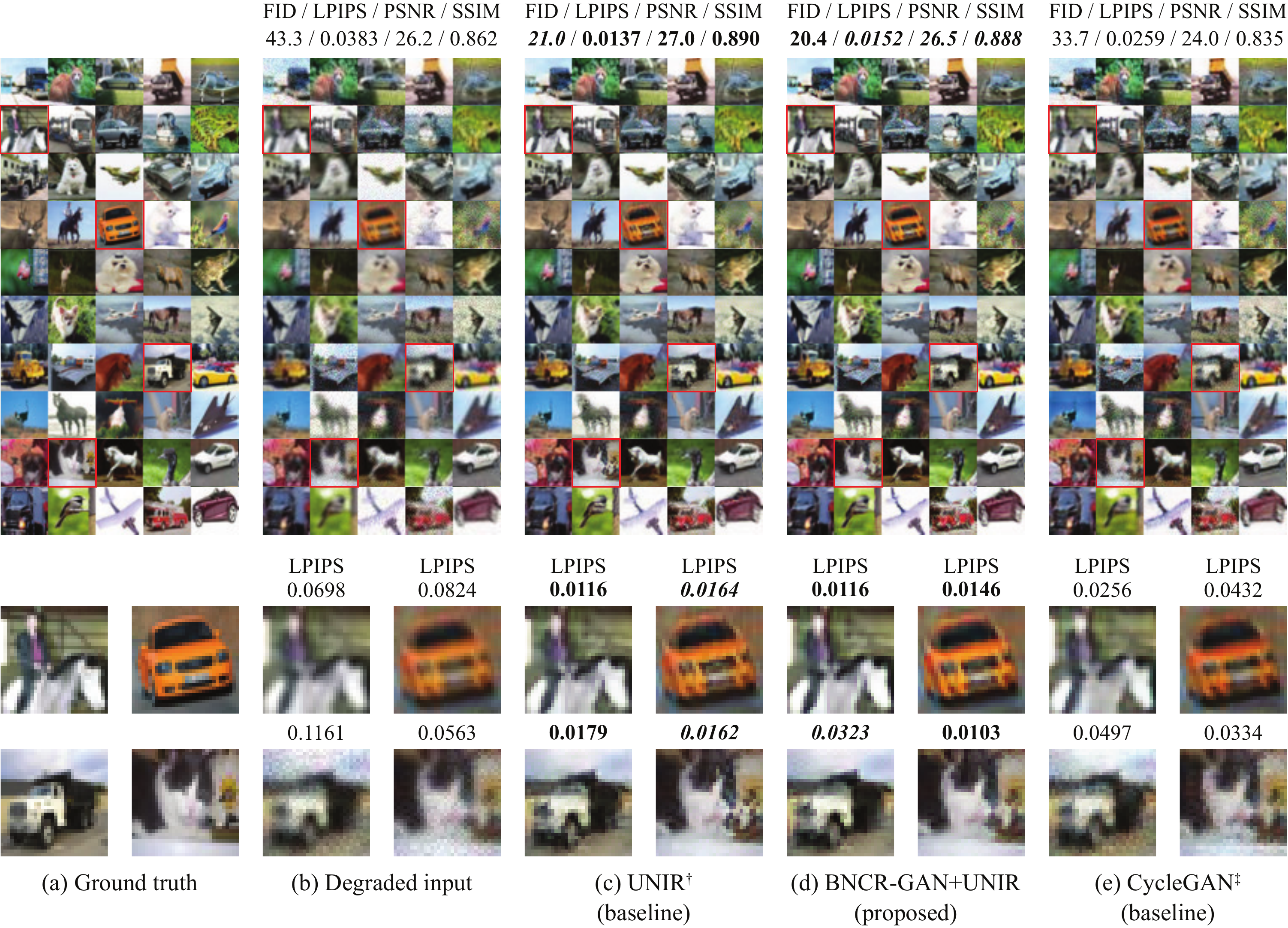}
  \caption{\textbf{Examples of restored images}.
    These are the results for the setting where the training and test images were degraded in setting (Q) and setting (N), respectively.
    The numbers on the top indicate the FID$\downarrow$, LPIPS$\downarrow$, PSNR$\uparrow$, and SSIM$\uparrow$ values that were calculated for all test images.
    These values were similar to those listed in Table~\ref{table:eval_restoration_ex}.
    The number above the bottom individual image represents the LPIPS$\downarrow$ between the image (in either (b), (c), (d), or (e)) and the corresponding ground-truth image (in (a)).
    UNIR$^\dag$ and CycleGAN$^\ddag$ were trained with additional supervision (i.e., the predefined degradation model and set-level supervision, respectively), while BNCR-GAN+UNIR was trained without them.}
  \label{fig:examples_restoration}
\end{figure*}

\begin{itemize}
\item Figure~\ref{fig:examples_restoration}:
  Examples of restored images.
\end{itemize}

\clearpage
\clearpage
\section{Implementation details}
\label{sec:details}

\smallskip\noindent\textbf{Notation.}
We use the following notation in our description of the network architectures.
\begin{itemize}
  \setlength{\parskip}{1pt}
  \setlength{\itemsep}{1pt}
\item Linear:
  Linear layer
\item Conv:
  Convolutional layer
\item ReLU:
  Rectified linear unit~\cite{VNairICML2010}
\item ResBlock:
  Residual block~\cite{KHeCVPR2016}
\item Norm:
  Normalization by sum
\end{itemize}

We use the following notation in our description of the training settings.
We used an Adam optimizer~\cite{DPKingmaICLR2015} for all training.
\begin{itemize}
  \setlength{\parskip}{1pt}
  \setlength{\itemsep}{1pt}
\item $\alpha$:
  Learning rate of Adam
\item $\beta_{1}$:
  First-order momentum parameter of Adam
\item $\beta_{2}$:
  Second-order momentum parameter of Adam
\end{itemize}

\subsection{Details on CIFAR-10
  (Sections~\ref{subsec:experimental_settings}--\ref{subsec:eval_cifar10_all})}
\label{subsec:details_cifar10}

\noindent\textbf{Network architectures.}
Table~\ref{table:net_cifar10} lists the generator and discriminator network architectures.
Following the study of NR-GAN~\cite{TKanekoCVPR2020}, we implemented $G_{\bm{x}}$ and $D_{\bm{y}}$ using ResNet architectures~\cite{KHeCVPR2016}.
In particular, we multiplied the output of ResBlock by $0.1$ and did not use any batch normalization~\cite{SIoffeICML2015} following a study on real gradient penalty regularization ($R_1$ regularization)~\cite{LMeschederICML2018}, which we used as the GAN regularization.

We implemented $G_{\bm{k}}$ and $G_{q}$ using two-hidden-layer multilayer perceptrons (MLPs).
In $G_{\bm{k}}$, the size of the blur kernel was set to $9 \times 9$.
When calculating $\hat{\bm{k}}^g$, we applied sigmoid and then normalized based on the sum (denoted by \textit{Norm} in Table~\ref{table:net_cifar10}) because a typical blur kernel is non-negative with a sum of $1$.
When calculating $q$ in $G_{q}$, we applied sigmoid and then multiplied by 100 to make the generated quality factor be within the range of $[0, 100]$.
In $G_{\bm{k}}$ and $G_{q}$, masking architectures were used, in which the final outputs were calculated using masks $\bm{m}_{\bm{k}}$ and $m_{q}$, respectively.

We used almost the same network architecture for $G_{\bm{x}}$ and $G_{\bm{n}}$.
As an exception, in $G_{\bm{n}}$, the output channels were doubled, among which the first half was used to express the standard deviation of the read noise $\bm{\sigma}_{\text{read}}$, and the remaining half was used to represent the standard deviation of the shot noise $\bm{\sigma}_{\text{shot}}$.
By using $\bm{\sigma}_{\text{read}}$ and $\bm{\sigma}_{\text{shot}}$, the read and shot noise $\bm{n}^g$ is calculated using $\bm{n}^g = \bm{\sigma}_{\text{read}} \cdot \bm{\epsilon}_{\text{read}} + \bm{\sigma}_{\text{shot}} \cdot \bm{\epsilon}_{\text{shot}} \cdot \sqrt{\bm{x}^g}$, where $\bm{\epsilon}_{\text{read}} \sim \mathcal{N}(\bm{0}, \bm{I})$ and $\bm{\epsilon}_{\text{shot}} \sim \mathcal{N}(\bm{0}, \bm{I})$.
Here, we used a reparameterization trick~\cite{DKingmaICLR2014}.

\smallskip\noindent\textbf{Training settings.}
Similar to the study on NR-GAN~\cite{TKanekoCVPR2020}, as a GAN objective function, we used a non-saturating GAN loss~\cite{IGoodfellowNIPS2014} with $R_1$ regularization~\cite{LMeschederICML2018}.
We set the weight parameter for $R_1$ regularization as $10$.
Inspired by the findings in~\cite{TKanekoCVPR2020}, we imposed a diversity-sensitive regularization~\cite{DYangICLR2019,QMaoCVPR2019} on $G_{\bm{k}}$, $G_{\bm{n}}$, and $G_{q}$ with weight parameters of $0.0004$, $0.02$, and $0.00001$, respectively.
Additionally, for BNCR-GAN, we added $\mathcal{L}_{\text{AC}}$ with a weight parameter of $0.1$, a scale parameter $\mu_{\bm{k}}$ of $1$, and a scale parameter $\mu_q$ of $10$.

We trained the networks for $300k$ iterations using the Adam optimizer~\cite{DPKingmaICLR2015} with $\alpha = 0.0002$, $\beta_1 = 0$, $\beta_2 = 0.99$, and a batch size of $64$.
We alternatively updated the generator and discriminator.
Similar to previous studies~\cite{TKarrasICLR2018,LMeschederICML2018,TKanekoCVPR2020}, we applied an exponential moving average with a decay of $0.999$ over the weights when creating the final generator.

\begin{table}[tb]
  \centering
  \setlength{\tabcolsep}{0pt}
  \scriptsize{
    \begin{tabularx}{\columnwidth}{CC}
      \addlinespace[-\aboverulesep]
      \cmidrule[\heavyrulewidth](r){1-1}
      \cmidrule[\heavyrulewidth]{2-2}
      \textsc{Image generator} $G_{\bm{x}}(\bm{z}_{\bm{x}})$
      & \textsc{Discriminator} $D_{\bm{y}}(\bm{y})$
      \\
      \cmidrule[\lightrulewidth](r){1-1}
      \cmidrule[\lightrulewidth]{2-2}
      $\bm{z}_{\bm{x}} \in \mathbb{R}^{128}$
      & $\bm{y} \in \mathbb{R}^{32 \times 32 \times 3}$
      \\
      \cmidrule(r){1-1}
      \cmidrule{2-2}
      Linear $\rightarrow$ $4 \times 4 \times 128$
      & ResBlock down $128$
      \\
      \cmidrule(r){1-1}
      \cmidrule{2-2}
      ResBlock up $128$
      & ResBlock down $128$
      \\
      \cmidrule(r){1-1}
      \cmidrule{2-2}
      ResBlock up $128$
      & ResBlock $128$
      \\
      \cmidrule(r){1-1}
      \cmidrule{2-2}
      ResBlock up $128$
      & ResBlock $128$
      \\
      \cmidrule(r){1-1}
      \cmidrule{2-2}
      ReLU, $3 \times 3$ Conv $3$
      & ReLU, Global average pooling
      \\
      \cmidrule(r){1-1}
      \cmidrule{2-2}
      Tanh $\rightarrow$ $\bm{x}^g$
      & Linear $\rightarrow$ $1$
      \\
      \cmidrule[\heavyrulewidth](r){1-1}
      \cmidrule[\heavyrulewidth]{2-2}
    \end{tabularx}
    \begin{tabularx}{\columnwidth}{CC}
      \cmidrule[\heavyrulewidth](r){1-1}
      \cmidrule[\heavyrulewidth]{2-2}
      \textsc{Blur-kernel generator} $G_{\bm{k}}(\bm{z}_{\bm{k}})$
      & \textsc{Quality-factor generator} $G_{q}(\bm{z}_{q})$
      \\
      \cmidrule[\lightrulewidth](r){1-1}
      \cmidrule[\lightrulewidth]{2-2}
      $\bm{z}_{\bm{k}} \in \mathbb{R}^{128}$
      & $\bm{z}_{q} \in \mathbb{R}^{128}$
      \\
      \cmidrule(r){1-1}
      \cmidrule{2-2}
      Linear $\rightarrow$ $128$, ReLU
      & Linear $\rightarrow$ $128$, ReLU
      \\
      \cmidrule(r){1-1}
      \cmidrule{2-2}
      Linear $\rightarrow$ $128$, ReLU
      & Linear $\rightarrow$ $128$, ReLU
      \\
      \cmidrule(r){1-1}
      \cmidrule{2-2}
      Linear $\rightarrow$ $9 \times 9$, Sigmoid, Norm $\rightarrow$ $\hat{\bm{k}}^g$
      & Linear $\rightarrow$ $1$, Sigmoid, $\times 100$ $\rightarrow$ $q^g$
      \\
      Linear $\rightarrow$ $9 \times 9$, Sigmoid $\rightarrow$ $\bm{m}_{\bm{k}}$
      & Linear $\rightarrow$ $1$, Sigmoid $\rightarrow$ $m_q$
      \\
      \cmidrule(r){1-1}
      \cmidrule{2-2}
      $\bm{k}^g = \bm{m}_{\bm{k}} \cdot \hat{\bm{k}}^g + (1 - \bm{m}_{\bm{k}}) \cdot \bm{k}_I$
      & $\bm{y}^g = m_q \psi^g (\bm{x}^g, q^g) + (1 - m_q) \bm{x}^g$
      \\
      \cmidrule[\heavyrulewidth](r){1-1}
      \cmidrule[\heavyrulewidth]{2-2}
    \end{tabularx}
  }
  \caption{\textbf{Generator and discriminator architectures for \textsc{CIFAR-10}.}}
  \label{table:net_cifar10}
  \vspace{-4mm}
\end{table}

\smallskip\noindent\textbf{Metrics.}
As discussed in Section~\ref{subsec:experimental_settings}, we used FID~\cite{MHeuselNIPS2017} as an evaluation metric because its validity has been demonstrated in large-scale studies on GANs~\cite{MLucicNeurIPS2018,KKurachICML2019}.
Furthermore, its sensitivity to image degradation has also been demonstrated~\cite{MHeuselNIPS2017,TKanekoCVPR2020}.
FID measures the 2-Wasserstein distance between a real distribution $p^r$ and a generative distribution $p^g$ by
\begin{flalign}
  \label{eqn:fid}
  d^2(p^r, p^g) & = \| \bm{m}^r - \bm{m}^g \|_2^2
  \nonumber \\
  & + \text{Tr} (\bm{C}^r + \bm{C}^g - 2 (\bm{C}^r \bm{C}^g)^{\frac{1}{2}}),
\end{flalign}
where $\{ \bm{m}^r, \bm{C}^r \}$ and $\{ \bm{m}^g, \bm{C}^g \}$ indicate the mean and covariance of the final feature vectors of the Inception model~\cite{CSzegedyCVPR2016} computed over real and generated samples, respectively.
More precisely, we calculated FID using $10k$ real test samples and $10k$ generated samples, following the suggestion from prior large-scale studies on GANs~\cite{MLucicNeurIPS2018,KKurachICML2019}.

\begin{table}[tb]
  \centering
  \setlength{\tabcolsep}{0pt}
  \scriptsize{    
    \begin{tabularx}{\columnwidth}{CC}
      \addlinespace[-\aboverulesep]
      \cmidrule[\heavyrulewidth](r){1-1}
      \cmidrule[\heavyrulewidth]{2-2}
      \textsc{Image generator} $G_{\bm{x}}(\bm{z}_{\bm{x}})$
      & \textsc{Discriminator} $D_{\bm{y}}(\bm{y})$
      \\
      \cmidrule[\lightrulewidth](r){1-1}
      \cmidrule[\lightrulewidth]{2-2}
      $\bm{z}_{\bm{x}} \in \mathbb{R}^{256}$
      & $\bm{y} \in \mathbb{R}^{128 \times 128 \times 3}$
      \\
      \cmidrule(r){1-1}
      \cmidrule{2-2}
      Linear $\rightarrow$ $4 \times 4 \times 1024$
      & ResBlock down $64$
      \\
      \cmidrule(r){1-1}
      \cmidrule{2-2}
      ResBlock up $1024$
      & ResBlock down $128$
      \\
      \cmidrule(r){1-1}
      \cmidrule{2-2}
      ResBlock up $512$
      & ResBlock down $256$
      \\
      \cmidrule(r){1-1}
      \cmidrule{2-2}
      ResBlock up $256$
      & ResBlock down $512$
      \\
      \cmidrule(r){1-1}
      \cmidrule{2-2}
      ResBlock up $128$
      & ResBlock down $1024$
      \\
      \cmidrule(r){1-1}
      \cmidrule{2-2}
      ResBlock up $64$
      & ResBlock $1024$
      \\
      \cmidrule(r){1-1}
      \cmidrule{2-2}
      ReLU, $3 \times 3$ Conv $3$
      & ReLU, Global average pooling
      \\
      \cmidrule(r){1-1}
      \cmidrule{2-2}
      Tanh $\rightarrow$ $\bm{x}^g$
      & Linear $\rightarrow$ $1$
      \\
      \cmidrule[\heavyrulewidth](r){1-1}
      \cmidrule[\heavyrulewidth]{2-2}
    \end{tabularx}
    \begin{tabularx}{\columnwidth}{CC}
      \cmidrule[\heavyrulewidth](r){1-1}
      \cmidrule[\heavyrulewidth]{2-2}
      \textsc{Blur-kernel generator} $G_{\bm{k}}(\bm{z}_{\bm{k}})$
      & \textsc{Quality-factor generator} $G_{q}(\bm{z}_{q})$
      \\
      \cmidrule[\lightrulewidth](r){1-1}
      \cmidrule[\lightrulewidth]{2-2}
      $\bm{z}_{\bm{k}} \in \mathbb{R}^{256}$
      & $\bm{z}_{q} \in \mathbb{R}^{256}$
      \\
      \cmidrule(r){1-1}
      \cmidrule{2-2}
      Linear $\rightarrow$ $128$, ReLU
      & Linear $\rightarrow$ $128$, ReLU
      \\
      \cmidrule(r){1-1}
      \cmidrule{2-2}
      Linear $\rightarrow$ $128$, ReLU
      & Linear $\rightarrow$ $128$, ReLU
      \\
      \cmidrule(r){1-1}
      \cmidrule{2-2}
      Linear\! $\rightarrow$\! $15 \times 15$, Sigmoid, Norm\! $\rightarrow$\! $\hat{\bm{k}}^g$
      & Linear $\rightarrow$ $1$, Sigmoid, $\times 100$ $\rightarrow$ $q^g$
      \\
      Linear $\rightarrow$ $15 \times 15$, Sigmoid $\rightarrow$ $\bm{m}_{\bm{k}}$
      & Linear $\rightarrow$ $1$, Sigmoid $\rightarrow$ $m_q$
      \\
      \cmidrule(r){1-1}
      \cmidrule{2-2}
      $\bm{k}^g = \bm{m}_{\bm{k}} \cdot \hat{\bm{k}}^g + (1 - \bm{m}_{\bm{k}}) \cdot \bm{k}_I$
      & $\bm{y}^g = m_q \psi^g (\bm{x}^g, q^g) + (1 - m_q) \bm{x}^g$
      \\
      \cmidrule[\heavyrulewidth](r){1-1}
      \cmidrule[\heavyrulewidth]{2-2}
    \end{tabularx}
  }
  \caption{\textbf{Generator and discriminator architectures for \textsc{FFHQ}.}}
  \label{table:net_ffhq}
  \vspace{-4mm}
\end{table}

\subsection{Details on FFHQ (Section~\ref{subsec:eval_ffhq})}
\label{subsec:details_ffhq}

\noindent\textbf{Network architectures.}
Table~\ref{table:net_ffhq} lists the generator and discriminator network architectures.
We used similar network architectures as those used on \textsc{CIFAR-10} (Appendix~\ref{subsec:details_cifar10}) except for the change according to the change in image size.
Specifically, we implemented $G_{\bm{x}}$ and $D_{\bm{y}}$ using ResNet architectures~\cite{KHeCVPR2016}, in which we multiplied the output of ResBlock by $0.1$ and did not apply any batch normalization~\cite{SIoffeICML2015}.

We implemented $G_{\bm{k}}$ and $G_{q}$ using two-hidden-layer MLPs.
In $G_{\bm{k}}$, the size of the blur kernel was set to $15 \times 15$.
When calculating $\hat{\bm{k}}^g$, we applied sigmoid and then normalized based on the sum (denoted by \textit{Norm} in Table~\ref{table:net_ffhq}) to obtain a non-negative kernel with a sum of $1$.
When calculating $q$ in $G_{q}$, we applied sigmoid and then multiplied by 100 to make the generated quality factor within the range of $[0, 100]$.
In $G_{\bm{k}}$ and $G_{q}$, masking architectures were used, in which the final outputs were computed using masks $\bm{m}_{\bm{k}}$ and $m_{q}$, respectively.

We implemented $G_{\bm{n}}$ using a network architecture similar to that of $G_{\bm{x}}$, except that the output channels were doubled.
As described in Appendix~\ref{subsec:details_cifar10}, the read and shot noise $\bm{n}^g$ is calculated as follows: $\bm{n}^g = \bm{\sigma}_{\text{read}} \cdot \bm{\epsilon}_{\text{read}} + \bm{\sigma}_{\text{shot}} \cdot \bm{\epsilon}_{\text{shot}} \cdot \sqrt{\bm{x}^g}$, where $\bm{\sigma}_{\text{read}}$ indicates the first half of the output channels of $G_{\bm{n}}$, $\bm{\sigma}_{\text{read}}$ indicates the latter half of the output channels of $G_{\bm{n}}$, $\bm{\epsilon}_{\text{read}} \sim \mathcal{N}(\bm{0}, \bm{I})$, and $\bm{\epsilon}_{\text{shot}} \sim \mathcal{N}(\bm{0}, \bm{I})$.
Here, a reparameterization trick~\cite{DKingmaICLR2014} was used.

\smallskip\noindent\textbf{Training settings.}
Similar to the settings of \textsc{CIFAR-10} (Appendix~\ref{subsec:details_cifar10}), as a GAN objective function, we used a non-saturating GAN loss~\cite{IGoodfellowNIPS2014} with $R_1$ regularization~\cite{LMeschederICML2018} where the weight parameter was set to $10$.
We imposed a diversity-sensitive regularization~\cite{DYangICLR2019,QMaoCVPR2019} on $G_{\bm{k}}$, $G_{\bm{n}}$, and $G_{q}$ with weight parameters of $0.0004$, $0.2$, and $0.00001$, respectively.
Additionally, for BNCR-GAN, we added $\mathcal{L}_{\text{AC}}$ with a weight parameter of $0.1$, a scale parameter $\mu_{\bm{k}}$ of $1$, and a scale parameter $\mu_q$ of $10$.

All networks except for those used in setting (N) were trained for $200k$ iterations using the Adam optimizer with $\alpha = 0.0001$, $\beta_1 = 0$, $\beta_2 = 0.99$, and a batch size of $64$.
In setting (N), which was the most degraded setting, we trained the networks for $300k$ iterations because the convergence was relatively slow.
We alternatively updated the generator and discriminator.
Similar to previous studies~\cite{TKarrasICLR2018,LMeschederICML2018,TKanekoCVPR2020}, to construct the final generator, we applied an exponential moving average with a decay of $0.999$ over the weights.

\smallskip\noindent\textbf{Metrics.}
For the same reason as that mentioned in Appendix~\ref{subsec:details_cifar10}, we used FID as an evaluation metric.
We calculated FID using $10k$ real test samples and $10k$ generated samples.
This protocol has been recommended in previous large-scale studies on GANs~\cite{MLucicNeurIPS2018,KKurachICML2019}.

\subsection{Details on image restoration (Section~\ref{subsec:restoration})}
\label{subsec:details_restoration}

\begin{table}[tb]
  \centering
  \setlength{\tabcolsep}{0pt}
  \scriptsize{
    \begin{tabularx}{0.5\columnwidth}{C}
      \toprule
      \textsc{Image translator} $T_{\bm{y}}(\bm{y})$
      \\ \cmidrule[\lightrulewidth]{1-1}
      $\bm{y}$
      \\ \midrule
      $3 \times 3$ Conv 128
      \\ \midrule
      ResBlock 128
      \\ \midrule
      ResBlock 128
      \\ \midrule
      ResBlock 128
      \\ \midrule
      ReLU, $3 \times 3$ Conv 3
      \\ \midrule
      Tanh $\rightarrow$ $\bm{x}$
      \\ \bottomrule
    \end{tabularx}
  }
  \vspace{1mm}
  \caption{\textbf{Image translator architecture used for image restoration.}}
  \label{table:net_translator}
  \vspace{-4mm}
\end{table}

\noindent\textbf{Network architectures.}
Table~\ref{table:net_translator} provides the image translator network architecture used for image restoration.
The image translator $T_{\bm{y}}$ converts the degraded image $\bm{y}$ into a corresponding clean image $\bm{x}$.
Based on the findings of a study on UNIR~\cite{APajotICLR2019}, which was used as a baseline in this study, we did not add downsampling and upsampling layers to the image translator.
Regarding the discriminator, we used the same network that was used for training BNCR-GAN (Table~\ref{table:net_cifar10}).
To diminish the effect caused by the network architecture differences, the same network architectures were used when implementing image translators and discriminators in UNIR, BNCR-GAN+UNIR, and CycleGAN.

\smallskip\noindent\textbf{Training settings.}
Following a previous study~\cite{APajotICLR2019}, when training \textit{UNIR}, we used adversarial and reconstruction losses.
The adversarial loss is defined as
\begin{flalign}
  \label{eqn:unir_adv}
  & \mathcal{L}_{\text{adv}} = \mathbb{E}_{\bm{y}^r \sim p^r(\bm{y})} [\log D_{\bm{y}}(\bm{y}^r)]
  \nonumber \\
  & + \mathbb{E}_{\bm{y}^r \sim p^r(\bm{y}), \bm{k}^r \sim p^r(\bm{k}), \bm{n}^r \sim p^r(\bm{n}), q^r \sim p^r(q)} [\log (1 - D_{\bm{y}}( \bm{y}^g ))],
\end{flalign}
where $\bm{y}^g = \psi^g((T_{\bm{y}}(\bm{y}^r) * \bm{k}^r + \bm{n}^r), q^r)$, and $D_{\bm{y}}$ is a discriminator that distinguishes a real \textit{degraded} image $\bm{y}^r$ from a \textit{degraded} translated image $\bm{y}^g$.
Similar to AmbientGAN~\cite{ABoraICLR2018}, a blur kernel, noise, and quality factor in the second term were sampled from the predefined blur-kernel distribution $\bm{k}^r \sim p^r(\bm{k})$, noise distribution $\bm{n}^r \sim p^r(\bm{n})$, and quality-factor distribution $q^r \sim p^r(q)$, respectively.
Intuitively, by training $T_{\bm{y}}$ and $D_{\bm{y}}$ adversarially using this loss, $p^g(\bm{y})$ is encouraged to approximate $p^r(\bm{y})$.
Furthermore, here, the same degradation model is used to produce $\bm{y}^g$ and $\bm{y}^r$; therefore, the corresponding before-degradation image distributions, that is, $p^g(\bm{x})$ and $p^r(\bm{x})$, are also encouraged to be coincident.

The reconstruction loss $\mathcal{L}_{\text{rec}}$ is defined as
\begin{flalign}
  \label{eqn:unir_rec}
  & \mathcal{L}_{\text{rec}} = \mathbb{E}_{\bm{y}^r \sim p^r(\bm{y}), \bm{k}^r \sim p^r(\bm{k}), \bm{n}^r \sim p^r(\bm{n}), q^r \sim p^r(q)}
  \nonumber \\
  & \:\:\:\:\:\:\:\:\:\:\:\:\:\: \| \bm{y}^g - \psi^g((T_{\bm{y}}(\bm{y}^g) * \bm{k}^r + \bm{n}^r), q^r) \|_2^2.
\end{flalign}
Similar to the cycle-consistency loss used in CycleGAN~\cite{JYZhuICCV2017}, this loss is useful for finding the optimal pseudo-pair between the degraded image (i.e., $\bm{y}^g$) and the restored image (i.e., $T_{\bm{y}}(\bm{y}^g)$) through the cycle conversion process.

In the initial experiments, we found that an identity mapping loss, which strengthens the identity between conversion, is useful for content preservation.
Therefore, we used the identity mapping loss in the experiments.
The identity mapping loss is defined as
\begin{flalign}
  \label{eqn:unir_id}
  \mathcal{L}_{\text{id}} = \mathbb{E}_{\bm{y}^r \sim p^r(\bm{y})} \| \bm{y}^r - T_{\bm{y}}(\bm{y}^r) \|_2^2.
\end{flalign}

The full objective of UNIR is written as
\begin{flalign}
  \label{eqn:unir}
  \mathcal{L}_{\text{UNIR}} = \mathcal{L}_{\text{adv}} + \lambda_{\text{rec}} \mathcal{L}_{\text{rec}} + \lambda_{\text{id}} \mathcal{L}_{\text{id}},
\end{flalign}
where $T_{\bm{y}}$ was optimized by minimizing $\mathcal{L}_{\text{UNIR}}$, whereas $D_{\bm{y}}$ was optimized by maximizing $\mathcal{L}_{\text{UNIR}}$.
Here, $\lambda_{\text{rec}}$ and $\lambda_{\text{id}}$ are weight parameters, and are set to $1$ and $0.1$, respectively, in the experiments.

A limitation of UNIR is that it assumes that degradation simulation models (that is, $\bm{k}^r \sim p^r(\bm{k})$, $\bm{n}^r \sim p^r(\bm{n})$, and $q^r \sim p^r(q)$, which are used in Equations~\ref{eqn:unir_adv} and \ref{eqn:unir_rec}), are predefined.
To overcome this limitation, in \textit{BNCR-GAN+UNIR}, the degradation simulation models were replaced with a kernel, noise, and quality-factor generators (i.e., $G_{\bm{k}}$, $G_{\bm{n}}$, and $G_q$, respectively), which were learned by BNCR-GAN in an unsupervised manner.
Except for this replacement, similar objective functions were used when training BNCR-GAN+UNIR.

When training \textit{CycleGAN}~\cite{JYZhuICCV2017}, we first divided the training images into clean images $\hat{\bm x}^r$ and degraded images $\hat{\bm{y}}^r$ under the assumption that set-level supervision (i.e., labels indicating whether images are degraded) is provided.
Note that this assumption was not required in UNIR and BNCR-GAN+UNIR.
We then trained two image translators, $T_{\hat{\bm{y}}}$, which converts a degraded image $\hat{\bm{y}}$ into the corresponding clean image $\hat{\bm{x}}$; and $T_{\hat{\bm{x}}}$, which converts a clean image $\hat{\bm{x}}$ into the corresponding degraded image $\hat{\bm{y}}$, with two discriminators, a clean image discriminator $D_{\hat{\bm{x}}}$ and a degraded image discriminator $D_{\hat{\bm{y}}}$, using adversarial and cycle-consistency losses.
See CycleGAN paper~\cite{JYZhuICCV2017} for their detailed definitions.
In the initial experiments, we found that the identity mapping loss (Equation~\ref{eqn:unir_id}) was also useful for CycleGAN.
Therefore, it was used simultaneously.

All networks were trained for $300k$ iterations using the Adam optimizer with $\alpha = 0.0001$, $\beta_1 = 0$, $\beta_2 = 0.99$, and a batch size of $64$.
As a GAN objective function, we used a non-saturating GAN loss~\cite{IGoodfellowNIPS2014} with $R_1$ regularization~~\cite{LMeschederICML2018}.
We alternatively updated the image translator and discriminator.
To obtain stable results, we applied an exponential moving average~\cite{TKarrasICLR2018} with a decay of $0.999$ over the weights when creating the final image translator.

\smallskip\noindent\textbf{Metrics.}
As discussed in Section~\ref{subsec:restoration} and Appendix~\ref{subsec:analyses_image_restoration}, we used FID and LPIPS as the main evaluation metrics in Section~\ref{subsec:restoration} and examined the utility of PSNR and SSIM in Appendix~\ref{subsec:analyses_image_restoration}.
In the evaluation, we applied the image translator (which was learned by either UNIR, BNCR-GAN+UNIR, or CycleGAN) to $10k$ real test images that were \textit{degraded} in setting (N).
Note that these test images were excluded from the training, and the remaining $50k$ images were used for training.
$10k$ real \textit{clean} (i.e., before-degradation) test images were used as the ground truth.

\end{document}